\documentclass[a4paper,times,fleqn]{cas-dc}
\usepackage[numbers]{natbib}

\usepackage{graphicx}\graphicspath{{images/}}
\usepackage{enumitem}
\usepackage{multicol}
\usepackage{multirow}
\usepackage{longtable}
\usepackage{lscape}
\usepackage{float}

\begin{document}
\let\WriteBookmarks\relax
\def\floatpagepagefraction{1}
\def\textpagefraction{.001}

 \shorttitle{Ovarian Cancer Data Analysis using DL}

\shortauthors{M Hira et~al.}

\title [mode = title]{Ovarian Cancer Data Analysis using Deep Learning: A Systematic Review from the Perspectives of Key Features of Data Analysis and AI Assurance}                      



%
\author[1]{Muta Tah Hira} [orcid=0000-0001-7656-7618]

\ead{m.hira@tees.ac.uk}

\author[2]{Mohammad A. Razzaque}[orcid=0000-0002-5572-057X]

\ead{m.razzaque@tees.ac.uk}




\author[1]{Mosharraf Sarker}[orcid=0000-0003-4698-2161]

\ead{m.sarker@tees.ac.uk}

    
\affiliation[1]{organization={School of Health and Life Sciences, Teesside University}, country={UK}}
\affiliation[2]{organization={School of Computing, Engineering and Digital Technologies, Teesside University}, country={UK}}

\cortext[cor1]{Corresponding author: Mohammad A. Razzaque}

\fntext[fn1]{First and second authors contributed equally.}


\begin{abstract}
Background and objectives: Technological advancement and the adoption of digital technologies in cancer care and research have generated big data. These diverse and multi-modal data contain high-density valuable information in various cancer subdomains, including early detection and accurate diagnosis. By extracting this information, Machine or Deep Learning (ML/DL)-based autonomous data analysis tools can assist clinicians and cancer researchers in discovering patterns and relationships from complex data sets. Many DL-based analyses on ovarian cancer (OC) data have recently been published. These analyses are highly diverse in various aspects of cancer (e.g., subdomain(s) and cancer type they address) and data analysis features (e.g., data modality, analysis method). However, a comprehensive understanding of these analyses in terms of these features and AI assurance (AIA) is currently lacking. This systematic review aims to fill this gap by examining the existing literature and identifying important aspects of OC data analysis using DL, explicitly focusing on the key features and AI assurance perspectives.

Methods: The PRISMA framework was used to conduct comprehensive searches in three journal databases. Only studies published between 2015 and 2023 in peer-reviewed journals were included in the analysis.

Results: In the review, a total of 96 DL-driven analyses were examined. The findings reveal several important insights regarding DL-driven ovarian cancer data analysis: \\
- Most studies 71\% (68 out of 96) focused on detection and diagnosis, while no study addressed the prediction and prevention of OC.\\
- The analyses were predominantly based on samples from a non-diverse population (75\% (72/96 studies)), limited to a geographic location or country.\\
- Only a small proportion of studies (only 33\% (32/96)) performed integrated analyses, most of which used homogeneous data (clinical or omics).\\
- Notably, a mere 8.3\% (8/96) of the studies validated their models using external and diverse data sets, highlighting the need for enhanced model validation, and \\
- The inclusion of AIA in cancer data analysis is in a very early stage; only 2.1\% (2/96) explicitly addressed AIA through explainability.

Conclusions: This systematic review highlights the critical areas requiring attention in DL-based cancer data analyses, especially OC data analysis. Future research should address the identified gaps, including exploring diverse and heterogeneous data-driven integrated analyses, validating models using external data sets from different demographic populations, and emphasising AI assurance through all aspects, including explainability and safety.

\end{abstract}




\maketitle

\section{Introduction}
\label{sec1}


Ovarian cancer (OC) is a widespread and fatal gynaecological cancer with a high death rate in developed countries. It causes 5\% of all cancer deaths in women in the UK~\cite{ovarian-uk} and USA~\cite{OvCAStats2018}. OC is often diagnosed at an advanced stage due to its asymptomatic early stage and non-specific symptoms in the late stage~\cite{doubeni2016diagnosis}. Furthermore, the location and position of the ovaries make it difficult to detect until the ovarian mass is large or metastasis occurs. Early detection and diagnosis of OC are difficult due to its non-specific symptoms, and multiple doctor visits and tests may be required. In this context, advanced technologies-driven research into the underlying mechanisms of cancer can lead to better diagnostic tools,  treatments and prognosis~\cite{doubeni2016diagnosis, Crucial_role_multiomic2018}. 


The advancement of technology (e.g., high-throughput technologies, ICT) has led to a vast amount of data being generated in cancer research and care. This big data mainly includes clinical data such as various modalities of image data (e.g., CT, MRI, histopathological), molecular or multi-omics~\cite{ OmiVAE2019a,chen2022deep, saida2022diagnosing, Multi-omics-2021}, and cancer management related data (e.g., personal management data~\cite{Empowerment18}). These data, especially histopathological images and molecular data contain high-density information, such as predictive and prognostic molecular biomarkers, which are very useful in early detection and accurate cancer diagnosis. Extracting this information requires autonomous data analysis tools such as AI/ML/DL as they are impossible to extract by human experts (e.g., radiologists and pathologists)~\cite{DL-histology21}. As a result, ML and DL approaches~\cite{DL-histology21, OmiVAE2019a, Multi-omics-2021} have become valuable tools in cancer research and care. These approaches allow researchers to analyse large and complex data sets, leading to discoveries and insights that can improve diagnosis, treatment, and prognosis. Over the past decade, more than 50 thousand research works (as searched in Google and other databases) have been published on ML for cancer, and many of them are in OC~\cite{Multi-omics-2021, OV-detection-DL2019, zhao2020cup,wang2022weakly,han2022cell,kim2021prognostic,zhu2021sio}. Many of these works are based on DL as the ANN (Artificial Neural Network) of DL allows models to handle increasing amounts and diversity of data efficiently~\cite{lecun2015deep}. This efficiency makes DL particularly effective in tackling complex computational issues, such as large-scale image classification and integrated multi-omics analysis of cancer~\cite{lecun2015deep}. 

Although DL-based data analyses, such as cancer diagnosis and treatment selection, offer significant benefits, they also introduce risks that must be addressed. The autonomous, complex, and scalable nature of DL/ML systems presents risks that extend beyond those posed by typical software. These characteristics fundamentally challenge our current methods for evaluating and mitigating the risks associated with digital technologies~\cite{mokander2021ethics}, including DL-based cancer analyses. To ensure effective and wide-scale adoption of AI systems in any domain, including cancer~\cite{hasani2022trustworthy}, they must ensure that they function as planned. To offer such assurances, it is necessary to quantify capabilities and risks across various dimensions, including data quality, algorithm performance, statistical considerations, trustworthiness, security, and explainability. Furthermore, assurances in many dimensions must be domain-specific (e.g., cancer) as a generic assurance solution across all subareas, and domains could be suboptimal~\cite{batarseh2021survey}. For example, a DL system segmenting kidney images as an initial task for radiologist evaluation requires a different level of trust than a DL system diagnosing cancer and initiating chemotherapy.


The articles published on any individual cancer, such as OC, are highly diverse mainly in terms of (i) type of cancer (e.g., high-grade ovarian serous carcinoma (HGOSC), low-grade serous ovarian carcinoma (LGOSC) they address, (ii) their aims or goals (e.g., prediction, detection, diagnosis), (iii) types (e.g., image, molecular) and origin/source (e.g., cell, tissue) of the data, (iv) data integration methods used for the analyses, and their ML/DL approaches (e.g., CNN, LSTM). For example, Lei et al. ~\cite{OV-detection-DL2019} used a fine-tuned GoogLeNet neural network to extract high-level features from OC ultrasound (US) images for further analysis. On the other hand, Hwangbo et al.~\cite{hwangbo2021development} developed ML and DL models to predict platinum sensitivity in patients with HGSOC. Considering the importance of OC data analysis in prediction, early detection, and accurate diagnosis and prognosis, a holistic view of these DL-based diverse analyses, including the AIA perspective of these analyses, is essential. However, no work provides a holistic view of the field. Therefore, this study provides a holistic view of the field through a systematic review of DL-based OC data analyses from the key features and AIA perspectives. This review will try to answer the following research questions: 
\begin{enumerate}
    \item What are the key features of an ML/DL-based cancer data analysis?
    \item How much research has been done on DL-based OC data analysis from the perspective of each identified key feature?
    \item How is AIA linked to ML/DL-based cancer data analysis? What is the state of the existing DL-based OC data analyses from an AIA perspective?

\end{enumerate}


The primary contributions of this survey are as follows:
\begin{itemize}
    \item Identification of the key features of cancer data analysis and an overview of them, which are cancer agnostic, can be used in any cancer-related ML/DL-based study;
    \item A comprehensive and systematic review of the existing DL-based analysis of OC data using the outlined features, including AIA and the PRISMA framework-based systematic selection approach of existing studies;
    \item An overview of the open research challenges and future directions in DL-based OC data analysis.
    
\end{itemize}

The subsequent sections of the paper are structured as follows. Section 2 presents the procedures used to investigate and select the existing OC data analyses. In Section~\ref{sec3}, we first identify the key features of cancer data analysis, present an overview of each, and then present a summary of the existing DL-based OC data analyses from the perspectives of those features and AIA. This section also provides an overview of AIA, including various aspects of it, and a mapping between the relevant key features of cancer data analysis with AIA. Next, we outline open research challenges, suggesting future research directions, in Section~\ref{sec4}. Finally, Section~\ref{sec5} summarises the paper with our future work.

\begin{table*}[!hbt]
\small
\caption{Boolean Search Strings used for the Journal Databases.}
\label{tab:1}
\begin{tabular}{p{0.07\textwidth}p{0.86\textwidth}}
\hline
Database & Boolean search strings \\ \hline
PubMed & ("Deep Learning"[MeSH Terms] OR "Deep Learning"[Title/Abstract] OR "Deep Neural Network"[Title/Abstract] OR "Convolutional Neural Network"[Title/Abstract] OR "CNN"[Title/Abstract] OR "Autoencoder"[Title/Abstract]) AND (("ovarian neoplasms"[MeSH Terms] OR "ovarian cancer"[Title/Abstract] OR "epithelial ovarian cancer"[Title/Abstract] OR "serous ovarian cancer"[Title/Abstract] OR "high-grade serous carcinoma"[Title/Abstract] OR "low-grade serous carcinoma"[Title/Abstract] OR "endometrioid carcinoma"[Title/Abstract] OR "mucinous ovarian cancer"[Title/Abstract] OR "clear cell carcinoma"[Title/Abstract] OR "ovarian granulosa cell tumour"[Title/Abstract] OR "granulosa cell tumour of ovary"[Title/Abstract] OR "germ cell tumour"[Title/Abstract] OR ("teratoma"[MeSH Terms] OR "teratoma"[All Fields] OR "teratomas"[All Fields]) OR "embryonal carcinoma"[Title/Abstract] OR "choriocarcinoma"[Title/Abstract] OR "polyembryomas"[Title/Abstract] OR "gonadoblastoma"[Title/Abstract]))
\\ \hline

Web of Science & "((TS=( ovarian cancer OR epithelial ovarian cancer OR serous ovarian cancer OR high-grade serous carcinoma OR low-grade serous carcinoma OR endometrioid carcinoma OR mucinous ovarian cancer OR ovarian granulosa cell tumour OR granulosa cell tumour of
ovary OR ovarian theca cell tumour OR germ cell tumour OR teratoma OR embryonal carcinoma OR choriocarcinoma OR polyembryomas OR gonadoblastoma) AND TS = (Deep learning OR deep neural network OR Convolutional Neural Network OR CNN OR Autoencoder) NOT TS= (Comment OR editorial OR letter OR case reports)) AND PY=(2015-2023) and Meeting Abstract (Exclude – Document Types) and Editorial Material (Exclude – Document Types) and Proceeding Paper or Review Article (Exclude – Document Types)"
 \\ \hline
Scopus & ("Deep Learning" OR "Deep Neural Network" OR "Convolutional Neural Network" OR "CNN" OR "Autoencoder") AND ("ovarian cancer" OR "epithelial ovarian cancer" OR "serous ovarian cancer" OR "high-grade serous carcinoma" OR "low-grade serous carcinoma" OR "mucinous ovarian cancer" OR "ovarian granulosa cell tumour" OR "granulosa cell tumour of ovary" ) AND PUBYEAR > 2014 AND PUBYEAR < 2024 AND ( LIMIT-TO (DOCTYPE, "ar" ) ) AND ( LIMIT-TO (LANGUAGE, "English"))  \\ \hline

\end{tabular}
\end{table*}

\begin{figure*}[hbt!]
\centering
\includegraphics[width=15cm]{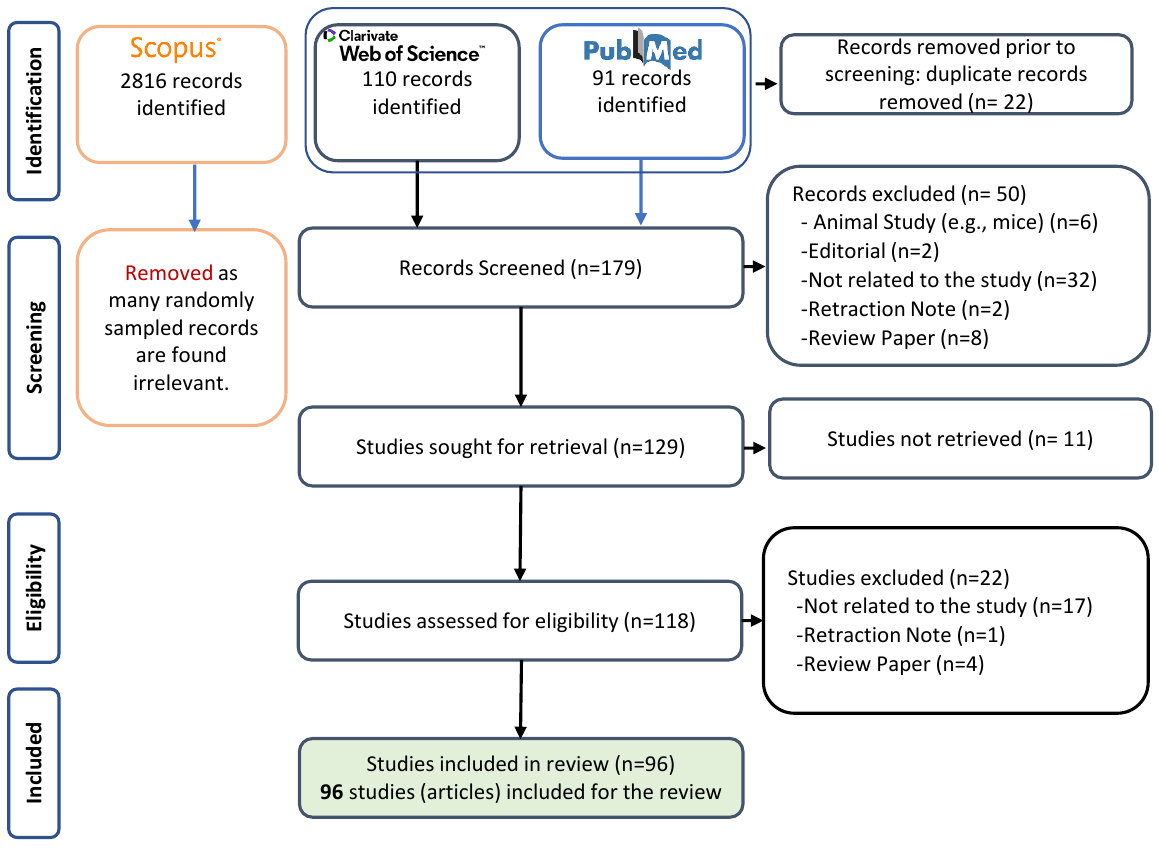}
\caption{PRISMA-based Existing Work Selection Strategies }
\label{fig1a}
\end{figure*}

\section{Existing Work Selection Strategies}

We used the recent PRISMA guideline~\cite{Pagen71} to filter our search results systematically. We identified 96 qualified scientific articles for the review study (Fig.~\ref{fig1a}). The selection started with 110 and 91 articles from the search results in PubMed and Web of Science databases using the Boolean search strings shown in Table~\ref{tab:1}. We also searched the Scopus database and found 2816 entries but excluded them as random sampling showed that most were irrelevant. Our study included all publications up to May 20, 2023, and 22 duplicate entries in two databases were removed (Fig~\ref{fig1a}). Subsequently, we selected the titles and abstracts of the publications, removing conference papers, editorials, perspectives, non-journal articles, other review and survey studies, and irrelevant studies that were not DL, OC or human-related. We were left with 118 scientific studies to be assessed for inclusion in this review study. Finally, after reviewing the list following the exclusion criteria, we found 96 journal articles to review in this study.

\begin{table*}[]
\caption{List of Acronyms with their Definitions}
\label{tab:acronym}
\resizebox{\textwidth}{!}{%
\begin{tabular}{llllll}
\hline
Acronym & Definition                          & Acronym & Definition                     & Acronym & Definition                                \\ \hline
A       & Accuracy                            & DQI     & Data Quality Improvement       & NBNBY   & Network-based Non-Bayesian                \\ \hline
AI    & Artificial Intelligence        & DQIC  & Improve Data Quality and Classification            & NF-BY   & Network-free Bayesian                  \\ \hline
AIA   & AI Assurance                   & ECODI & Environmental, Clinical and Omics Data Integration & NFNBY   & Network Free Non Bayesian              \\ \hline
BCP     & Bayesian Change Point               & EOC     & Epithelial Ovarian Cancer      & ODI     & Omics Data Integration                    \\ \hline
BY      & Bayesian                            & EV      & External Validation            & O-RADS  & Ovarian Adnexal Reporting and Data System \\ \hline
C       & Classification/Clinical                     & FCN     & Fully Convolutional Networks   & OS      & Overall Survival                          \\ \hline
CA125   & Cancer Antigen 125                  & FDI     & Feature Level Data Integration & P       & Prognosis                                 \\ \hline
CBC   & Changes in Body Composition    & GAN   & Generative Adversarial Network                     & PP      & Prediction and Prevention              \\ \hline
CCC   & Clear Cell Carcinoma           & GAT   & Graph Attention Network                            & RBM     & Restricted Boltzmann Machine           \\ \hline
CCLE  & Cancer Cell Line Encyclopaedia & HE4   & Human Epididymis Protein 4                         & RL      & Reinforcement Learning                 \\ \hline
CDG   & cancer data generation         & HGSOC & High-grade Serous Carcinoma                        & RNN     & Recurrent Neural Network               \\ \hline
CDI     & Clinical Data Integration           & HP      & Histopathology                 & S       & Sensitivity                               \\ \hline
Cl      & Clustering                          & Hybrid1 & ML + DL                        & SA      & Survival Analysis                         \\ \hline
CNVs    & Copy Number Variation               & Hybrid2 & DL + Statistical               & SCD     & Single-centred                            \\ \hline
CODI    & Clinical and Omics Data Integration & IM      & Image                          & SL      & Supervised Learning                       \\ \hline
CRE/M & Cancer Risk Estimation/Mapping & IM1   & Radiological Image                                 & SSL     & Semi-supervised Learning               \\ \hline
CSD     & Cancer Sub Domain                   & IM2     & Histopathological Image        & SSL     & Semi-Supervised Learning                  \\ \hline
CST     & Cancer Sub Type                     & IM3     & Other Images                   & SVM     & Support Vector Machine                    \\ \hline
DA      & Data Analysis                       & IV      & Internal validation            & TCGA    & The Cancer Genome Atlas                   \\ \hline
DAM     & Data Analysis Method                & LGSOC   & Low-grade Serous Carcinoma     & TL      & Transfer Learning                         \\ \hline
DBM     & Deep Boltzmann machine              & LSTM    & Long short-term memory         & TM      & Treatment and Management                  \\ \hline
DD      & Detection and Diagnosis             & MCD     & Multi-centred                  & TMP     & TM and Prognosis                          \\ \hline
DDP     & DD and Prognosis                    & MIA3G   & Multivariate Index Assay 3G    & TN      & Text or Numeric                           \\ \hline
DDTM    & DD and Treatment and Management     & ML      & Machine Learning               & UDA     & Use of Data Analysis                      \\ \hline
DeLI  & Decision Level Integration     & MM    & Multi-Modal data                                   & UKCTOCS & UK Collaborative Trial of OC Screening \\ \hline
DL      & Deep Learning                       & MMD-VAE & Maximum Mean Discrepancy VAE   & UL      & Unsupervised Learning                     \\ \hline
DLI     & Data Level Integration              & NB      & Network-based                  & VAE     & Variational Autoencoder                   \\ \hline
DNAs    & DNA Methylation                     & NB-BY   & Network-based Bayesian         & VM      & Validation Method                         \\ \hline
\end{tabular}%
}
\end{table*}

\section{Result and Discussion}\label{sec3}

\begin{figure*}[hbt!]
\centering
\includegraphics[width=17cm]{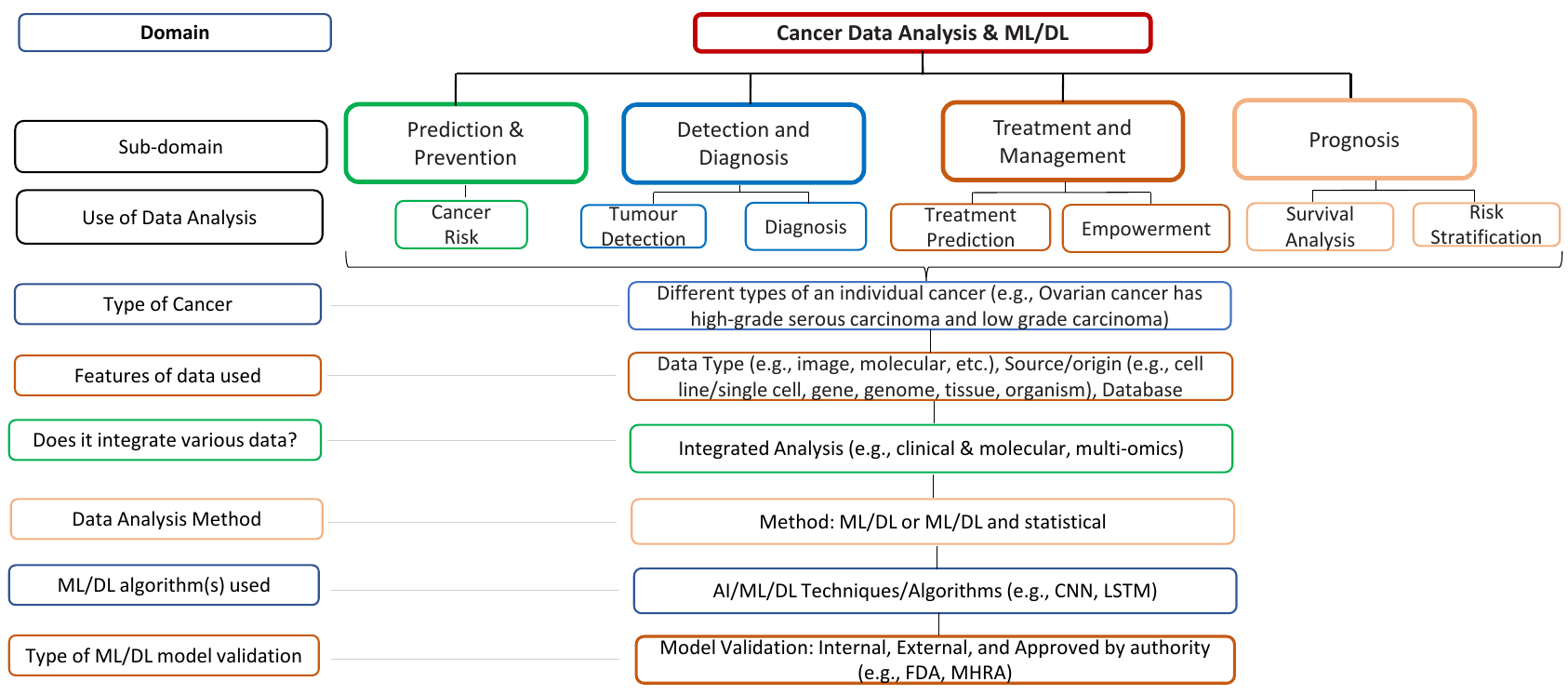}
\caption{Key features of Existing ML/DL-driven Data Analysis of Cancer}
\label{fig1}
\end{figure*}

Cancer is a very important and active (research and development-wise) application domain of DL-based solutions. Many research works have been published in various sub-domains (e.g., prediction, diagnosis, treatment) of individual cancer (e.g., Ovarian, Lung Cancer) and pancancer~\cite{ML-vs-Stats-Methods20, OmiVAE2019a}. Existing ML/DL-based data analyses in cancer exhibit diversity in key features (e.g., analysis method, data type, ML/DL algorithms). Therefore, it is crucial to identify and review these key features to gain a comprehensive understanding of the field and contribute to future research, particularly from the perspective of individual cancers such as OC. To make the features set generic (not specific to DL), we have identified and summarised them in Figure~\ref{fig1} based on the existing ML/DL-based data analyses~\cite{ML-in-Cancer2006, ML-in-Cancer2015, ML-in-Cancer2018, Explainable-ML-in-Cancer2021}. In the following subsections, we briefly discuss them individually with their status from existing DL-based OC data analyses.

\subsection{\textbf{Sub-domains of a Cancer}}

Generally, any study based on ML/DL-based cancer data analysis can be categorised into one of the four subdomains: (i) prediction and prevention, (ii) detection and diagnosis, (iii) treatment and management, and (iv) prognosis. In the following, we briefly discuss them individually, including the cancer issues they address.

\subsubsection{\textit{Prediction and Prevention (P\&P)}} 

Prevention and early intervention are the most effective approaches to avoid psychological, physical, and economic suffering from cancer. However, such proactive intervention needs the ability to accurately anticipate an individual's susceptibility or risk of cancer. ML/DL-based data analysis~\cite{CancerRiskPrediction2018, CancerRiskPrediction2020a, CancerRiskPrediction2021} is useful for predicting the probability of developing cancer.


\subsubsection{\textit{Detection and Diagnosis (DD)}}
This is the most active sub-domain of cancer~\cite{ML-in-Cancer2006, ML-in-Cancer2015, ML-in-Cancer2020, Explainable-ML-in-Cancer2021} where ML or DL are used to screen, detect and diagnose the disease. Work in this subdomain mainly uses ML / DL to analyse data from one of the three stages or types of diagnosis: (i) lab-test based~\cite{Labtest-Based-2021}, (ii) imaging-test based (radiomics-driven diagnosis)~\cite{Imagingtest-Based-2020a, Imagingtest-Based-2020b}, and (iii) biopsy-based (e.g., histopathological image analysis-based diagnosis)~\cite{ML-histology18, DL-histology21}, with an aim to assist clinicians in informed-decision making.

\subsubsection{\textit{Treatment and Management (TM)}}
Resistance to therapy is a leading cause of cancer treatment failure, resulting in many cancer-related deaths. The existing treatment procedure is based primarily on cancer subtypes and genetic modifications. However, the existence of a genetic modification does not necessarily indicate the therapeutic response, and the response may differ depending on the cancer subtype. As a result, an integrated study is needed to identify optimal treatment plans or therapies for a patient. ML/DL-based predictive models can use integrated data to predict treatment response (e.g.,~\cite{hwangbo2021development, wang2022weakly}) for individuals or groups and identify a personalised and effective treatment. Importantly, these models~\cite {sun2020dtf, Drug-design-2021} aid researchers and the industry in developing and designing cancer drugs to promote precision oncology (PO). Additionally, utilising PO in conjunction with ML/DL-based startification~\cite{ML-in-Cancer2020} will empower oncologists and ML/DL-based data analysis (e.g., behaviour, lifestyle, social media data~\cite{Patient-Empowerment2018}) can aid patients in managing their cancer.

\subsubsection{\textit{Prognosis (P)}}
Prognosis in cancer is another active sub-domain of ML and DL~\cite{ML-in-Cancer2006, ML-in-Cancer2015, Explainable-ML-in-Cancer2021}. In this sub-domain, ML or DL are used for (i) survival analysis~(e.g.,\cite{avesani2022ct, Multi-omics-2021}), (ii) cancer recurrence prediction (e.g.,~\cite{Recurrence-prediction-2020}), and (iii) residual disease prediction (e.g.,~\cite{ Residual-Disease2021}). Survival analysis (SA) includes survival risk startification~\cite{Risk-startification2020} and the prognosis, such as disease-specific or overall survival (OS) after diagnosis or therapy. On the other hand, cancer recurrence prediction estimates the likelihood of redeveloping, and residual disease prediction estimates cancer cells that remain after surgery.

\begin{figure}[hbt!]
\centering
\includegraphics[width=8.45cm]{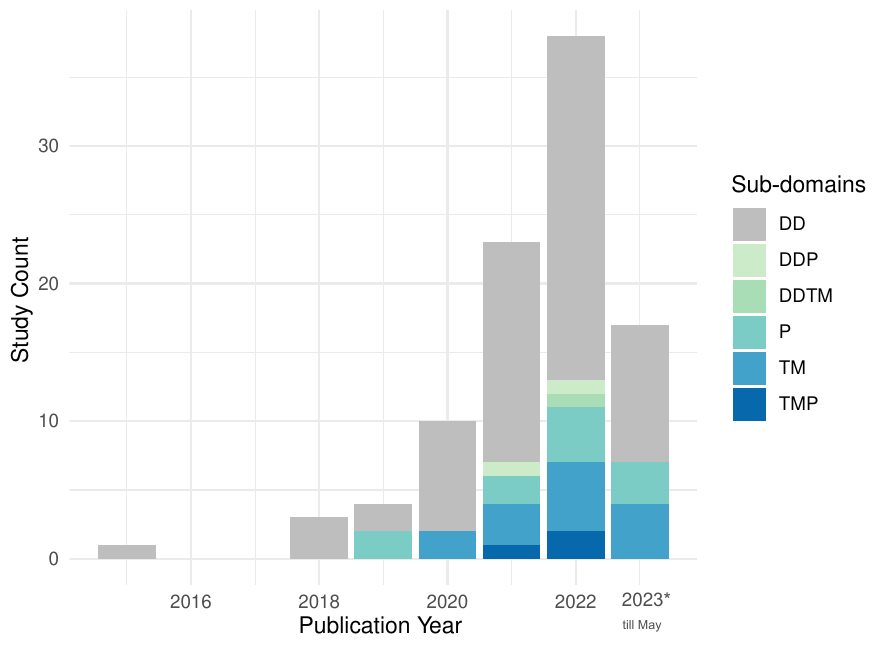}
\caption{Yearly Published Studies}
\label{fig1b}
\end{figure}

\begin{figure}[hbt!]
\centering
\includegraphics[width=8cm]{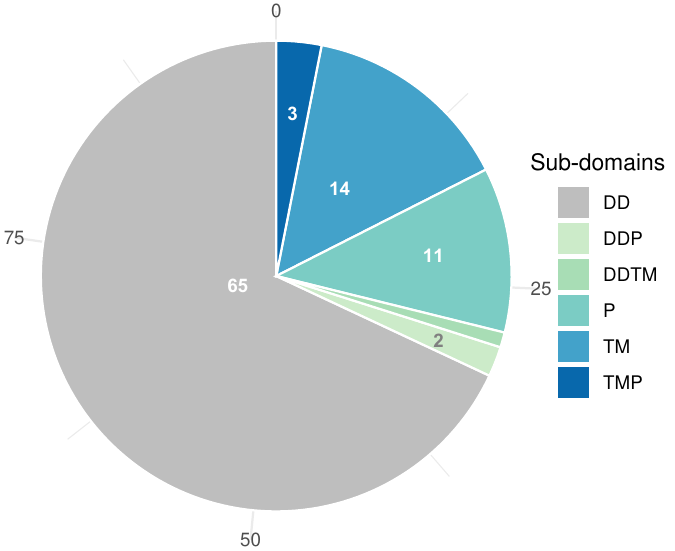}
\caption{Sub-domain wise Distribution of Existing Studies}
\label{fig2b}
\end{figure}

\subsection{Subdomains of Existing OC Data Analyses}

Figure~\ref{fig1b} and~\ref{fig2b} summarise subdomains of the existing OC data analyses and yearly published studies in each domain. Also, Table~\ref{tab:DD}, \ref{tab:TM}, and \ref{tab:P} respectively summarise the existing studies on DD, TM, and P in terms of the identified key features of DA and AIA. Figure \ref{fig1b} illustrates that most studies (95 out of 96) were published between 2018 and 2023, with a gradual increase in the published studies in the respective subdomains. In Figure~\ref{fig2b}, analysis distribution shows that 67.7\% (65 out of 96) focused on DD, 11.45\% (11 out of 96) on P, 14.6\% (14 out of 96) on TM, while the remaining 6 out of 96 covered mixed subdomains (e.g., DDP-DD and P, DDTM- DD and TM). In particular, the PP subdomain lacked work based on DL, except for a few studies based on ML~\cite{CancerRiskPrediction2018, kim2014empirical}. The scarcity of publications in this subdomain and the lower number of studies in the P and TM subdomains, in contrast to DD, can be attributed to the inaccessibility of relevant data sets. 

\begin{table*}[]
\caption{DL-based Data Analysis for Ovarian Cancer Detection and Diagnosis}
\label{tab:DD}
\resizebox{\textwidth}{!}{%
\begin{tabular}{llllllllllllll}
\hline
 &
   &
   &
  \multicolumn{3}{l}{Data Features} &
  \multicolumn{3}{l}{Integrated Analysis} &
  \multicolumn{3}{l}{Anlaysis Methods \& DL details} &
   &
   \\ \hline
Analysis &
  CST &
  Use-of-DA &
  Source &
  Format &
  Repository &
  Data &
  Level &
  Method &
  AM &
  LT &
  Algorithms &
  VM &
  AIA \\ \hline
  \cite{chen2022deep}                 & OC    & C     & C    & IM1 & SCD  & CDI  & FDeLI & NF-NBY1 & DL      & SL      & CNN          & IV       &      \\ \hline
  \cite{saida2022diagnosing}          & EOC   & C     & C    & IM1 & SCD  &      &       &         & DL      & SL      & CNN          & IV       &      \\ \hline
  \cite{Multi-omics-2021} & HGSOC & DQICCl & CO & TN & MCD1 & CODI & FLI & NF-NBY2 & Hybrid2 & UL\&SL & VAE\&DNN & IV \\ \hline
  \cite{avesani2022ct}                  & HGSOC & SA  & C   & IM1  & MCD1 & CDI   & FLI     & NF-NBY1 & Hybrid2 & SL           & CNN      & EV1  \\ \hline
\cite{hu2023deep}                   & EOC   & DQI   & C    & IM1 & MCD1 &      &       &         & DL      & SL      & CNN      & IV       &      \\ \hline
\cite{wang2023deep}                 & EOC   & DQIC  & C    & IM1 & SCD  &      &       &         & Hybrid2 & SL      & CNN          & IV       &      \\ \hline
\cite{gangadhar2023deep}            & OC    & C     & C    & IM3 & SCD  &      &       &         & DL      & SL      & CNN          & IV       &      \\ \hline
\cite{wang2023attention}            & OC    & C     & C    & IM1 & MCD1 &      &       &         & DL      & SL      & CNN          & IV       &      \\ \hline
\cite{gajjela2023leveraging}        & OC    & C     & C    & IM3 & SCD  &      &       &         & Hybrid1 & SL      & CNN          & IV       &      \\ \hline
\cite{wei2023associating}           & EOC   & C     & C    & MM  & MCD2 & CDI  & DeLI  & NF-NBY3 & Hybrid2 & SL      & CNN          & EV2      &      \\ \hline
\cite{boyanapalli2023ovarian}       & OC    & C     & C    & IM1 & MCD1 &      &       &         & Hybrid2 & SL      & CNN          & IV       &      \\ \hline
\cite{wang2022automated}            & OC    & C     & C    & IM2 & SCD  &      &       &         & DL      & SL      & CNN          & IV       &      \\ \hline
\cite{kodipalli2023computational}   & OC    & C     & C    & IM1 & SCD  &      &       &         & Hybrid1 & UL\&SL  & CNN          & IV       &      \\ \hline
\cite{wang2023dmff}                 & OC    & DQIC  & C    & IM1 & SCD  &      &       &         & DL      & SL      & CNN          & IV       &      \\ \hline
\cite{ho2023deep}                   & HGSOC & DQIC  & C    & IM2 & SCD  &      &       &         & DL      & SL      & DNN         & IV       &      \\ \hline
\cite{hema2022region}               & OC    & DQIC  & C    & IM3 & MCD1 &      &       &         & DL      & SL      & CNN          & IV       &      \\ \hline
\cite{hsu2022automatic}             & OC    & DQIC  & C    & IM1 & SCD  &      &       &         & Hybrid2 & SL\&TL  & CNN          & IV       &      \\ \hline
\cite{bahado2022precision}          & OC    & C     & O    & TN  & SCD  &      &       &         & Hybrid1 & SL      & DNN          & IV       &      \\ \hline
\cite{wang2022automatic}            & OC    & C     & C    & IM1 & SCD  &      &       &         & DL      & SL      & CNN          & IV       &      \\ \hline
\cite{nero2022deep}                 & EOC   & DQIC  & C    & IM2 & SCD  &      &       &         & DL      & SL      & CNN\&Tr      & IV       &      \\ \hline
\cite{jung2022ovarian}              & OC    & C     & C    & IM1 & SCD  &      &       &         & DL      & UL\&SL  & AE\&CNN      & IV       &      \\ \hline
\cite{mayer2022learn}               & HGSOC & C     & C    & IM2 & MCD2 &      &       &         & Hybrid2 & SL      & CNN          & EV2      &      \\ \hline
\cite{farahani2022deep}             & OC    & DQIC  & C    & IM2 & MCD2 &      &       &         & DL      & SL\&TL  & CNN          & EV2      &      \\ \hline
\cite{wu2022investigation}          & OC    & C     & O    & MM  & MCD1 &      &       &         & DL      & SL      & CNN          & IV       &      \\ \hline
\cite{white2022deep}                & OCO   & DQIC  & C    & MM  & MCD2 &      &       &         & DL      & UL\&SL  & AE\&CNN      & EV2      & XAI1 \\ \hline
\cite{reilly2022analytical}         & OC    & C     & C    & TN  & SCD  & CDI  & FLI   & NF-NBY1 & DL      & SL      & MIA3G        & IV       &      \\ \hline
\cite{sun2022xgbg}                  & OC    & DQIC  & O    & TN  & MCD2 & ODI  & FLI   & NB-NBY  & DL      & SL      & GCN          & IV       &      \\ \hline

\cite{gao2022deep}                  & OC    & C     & C    & IM1 & MCD1 &      &       &         & DL      & SL      & CNN          & EV1      &      \\ \hline

\cite{sengupta2022deep}             & OC    & C     & C    & MM  & SCD  & CDI  & DeLI  & NF-BY   & DL      & SL      & CNN          & IV       &      \\ \hline
\cite{jian2022mri}                  & EOC   & C     & C    & IM1 & MCD1 &      &       &         & Hybrid2 & SL      & CNN          & IV       &      \\ \hline
\cite{jeong2022deepcia}             & OCO   & DQI   & O    & TN  & MCD1 &      &       &         & DL      & SL      & CNN          & IV       &      \\ \hline
\cite{jeya2023factorization}        & OC    & DQIC  & C    & IM2 & MCD2 &      &       &         & DL      & SL      & CNN\&MBiLSTM & IV       &      \\ \hline
\cite{ramasamy2023hybridized}       & OC    & DQIC  & C    & IM2 & MCD1 &      &       &         & DL      & SL      & CNN          & IV       &      \\ \hline
\cite{kodipalli2022inception}       & OC    & C     & C    & IM1 & MCD1 &      &       &         & DL      & SL      & CNN          & IV       &      \\ \hline
\cite{jiang2022computational}       & HGSOC & DQIC  & C    & IM2 & SCD  &      &       &         & Hybrid2 & SL      & CNN          & IV       &      \\ \hline
\cite{ahn2022transcriptome}         & OCO   & CCl   & O    & TN  & MCD1 &      &       &         & DL      & SL      & DNN          & IV       &      \\ \hline
\cite{petrovsky2021managing}        & OCO   & C     & O    & MM  & SCD  & ODI  & FLI   & NF-NBY1 & DL      & SL      & CNN          & IV       &      \\ \hline
\cite{qazi2021silico}               & OC    &       & O    & TN  & MCD2 &      &       &         & DL      & SL      & DNN          & IV       &      \\ \hline
\cite{jian2021multiple}             & EOC   & C     & C    & IM1 & SCD  &      &       &         & DL      & SL      & CNN          & IV       &      \\ \hline
\cite{ye2021ovarian}                & OC    & C     & O    & TN  & MCD2 & ODI  & FLI   & NB-BY   & DL      & UL\&SL  & GAT\&DNN     & IV       &      \\ \hline
\cite{meng2021computationally}      & OC    & DQIC  & C    & IM2 & MCD1 &      &       &         & DL      & SSL\&SL & GAN\&CNN     & IV       &      \\ \hline
\cite{Pastuszak_2021}               & OC    & C     & CO   & IM3 & MCD2 &      &       &         & DL      & SL      & DNN          & IV       &      \\ \hline
\cite{xie2021diagnosis}             & OCO   & C     & C    & IM1 & SCD  & CDI  & FLI   & NF-NBY1 & DL      & SL      & CNN          & IV       &      \\ \hline
\cite{gonzalez2021characterization} & OC    & DQIC  & O    & MM  & MCD1 &      &       &         & DL      & SL      & CNN          & IV       &      \\ \hline
\cite{shin2021style}                & OC    & DQIC  & C    & IM2 & MCD2 &      &       &         & DL      & UL\&SL  & GAN\&CNN     & EV2      &      \\ \hline
\cite{christiansen2021ultrasound}   & OC    & C     & C    & IM1 & MCD1 &      &       &         & DL      & SL\&TL  & CNN          & IV       &      \\ \hline
\cite{wang2021evaluation}           & OC    & C     & C    & MM  & SCD  & CDI  & DeLI  & NF-NBY2 & DL      & SL      & DNN          & IV       &      \\ \hline
\cite{ghoniem2021multi}             & OC    & C     & CO   & MM  & MCD1 & CODI & DeLI  & NF-NBY1 & DL      & SL      & CNN\&LSTM    & IV       &      \\ \hline
\cite{Gupta_2021}                   & OCO   & C     & CO   & TN  & MCD2 &      &       &         & Hybrid1 & SL      & DNN          & IV       &      \\ \hline
\cite{Kopylov_2021}                 & OCO   & DQIC  & O    & TN  & SCD  &      &       &         & DL      & UL\&SL  & CNN          & IV       &      \\ \hline
\cite{chen2021prediction} &
  OC &
  DQIC &
  CO &
  TN &
  SCD &
  CODI &
  FLI &
  NF-NBY1 &
  Hybrid1 &
  UL\&SL &
  GCN &
  IV &
   \\ \hline
\cite{mohammed2021stacking}         & OCO   & DQIC  & O    & TN  & MCD1 &      &       &         & Hybrid2 & UL\&SL  & CNN          & IV       &      \\ \hline
\cite{guo2020deep}                  & OC    & CCl   & O    & TN  & MCD2 & ODI  & FLI   & NF-NBY1 & Hybrid2 & UL\&SL  & AE           & EV2      &      \\ \hline
\cite{cancers12092373}               & EOC   & DQIC  & C    & TN  & SCD  & CDI  & FLI   & NF-NBY1 & DL      & SL      & CNN          & IV       &      \\ \hline
\cite{Basharat_2020}                & OC    & DQIC  & O    & IM3 & MCD2 &      &       &         & DL      & SL      & CNN          & IV       &      \\ \hline
\cite{urase2020simulation}          & OC    & DQIC  & C    & IM2 & MCD1 &      &       &         & Hybrid2 & SL      & CNN          & IV       &      \\ \hline
\cite{kilicarslan2020diagnosis}     & OCO   & DQIC  & O    & TN  & MCD1 &      &       &         & Hybrid1 & UL\&SL  & AE\& CNN      & IV       &      \\ \hline
\cite{zhao2020cup}                     & OCO   & DQIC & O   & TN   & MCD2 &       &         &         & DL      & SL           & CNN      & EV2  \\ \hline
\cite{mallavarapu2020pathway} &
 
  OCO &
  Cl &
  O &
  TN &
  MCD1 &
  ODI &
  FLI &
  NB-NBY &
  Hybrid1 &
  UL\& SL &
  RBM\& DBN & IV \\ \hline
\cite{levine2020synthesis}             & OCO   & CDG  & C   & IM2  & MCD2 &       &         &         & DL      & UL\&SL       & GAN\&CNN & IV   \\ \hline
\cite{klein2019maldi}                   & EOC   & C    & C   & IM3  & SCD  &       &         &         & Hybrid1 & SL           & CNN      & IV   \\ \hline
\cite{alshibli2019shallow}              & OCO   & C    & O   & TN   & MCD2 &       &         &         & DL      & SL           & CNN      & IV   \\ \hline
\cite{vazquez2018quantitative} &

  OC &
  C &
  C &
  TN &
  MCD1 &
  CDI &
  FLI &
  NF-NBY2 &
  Hybrid2 &
  SL &
  RNN &
  IV \\ \hline
\cite{du2018classification}           & OCO   & DQIC & C   & IM2  & MCD1 &       &         &         & DL      & SL\&TL       & CNN      & IV   \\ \hline
\cite{wu2018automatic}               & OC    & DQIC & C   & IM2  & SCD  &       &         &         & DL      & SL           & CNN      & IV   \\ \hline
\cite{Liang2015}                        & OCO   & SA  & CO  & TN   & MCD2 & CDI   & FLI     & NF-NBY2 & DL      & UL\&SL       & DBN      & IV   \\ \hline
\cite{irajizad2022blood}               & OC    & C    & O   & TN   & MCD1 &       &         &         & Hybrid2 & SL           & DNN      & EV1  \\ \hline

\cite{li2022deep} & OC    & DQIC & C   & IM2  & SCD  &       &         &         & DL      & SL           & CNN      & IV   \\ \hline
\end{tabular}%
}
\end{table*}

\begin{table*}[]
\caption{DL-based Data Analysis for Ovarian Cancer TM}
\label{tab:TM}
\resizebox{\textwidth}{!}{%
\begin{tabular}{llllllllllllll}
\hline
         &     &           & \multicolumn{3}{l}{Data Features} & \multicolumn{3}{l}{Integrated Analysis} & \multicolumn{3}{l}{Anlaysis Methods \& DL details} &    &     \\ \hline
Analysis & CST & Use-of-DA & Source   & Format   & Repository  & Data       & Level       & Method       & AM           & LT           & Algorithms           & VM & AIA \\ \hline
\cite{wang2022weakly}          & EOC   & C    & C  & MM  & MCD1 &      &     &         & DL      & SSL\&SL & CNN       & IV  &      \\ \hline
\cite{han2022cell}             & OC    & DQIC & C  & IM2 & MCD2 & CDI  & FLI & NF-NBY2 & DL      & SSL\&SL & CNN       & IV  &      \\ \hline
\cite{hwangbo2021development}  & HGSOC & C    & C  & TN  & MCD1 & CDI  & FLI & NF-NBY1 & Hybrid2 & SL     & DNN\&O     & IV  &      \\ \hline
\cite{sun2020dtf}              & OCO   & C    & C  & TN  & MCD2 &      &     &         & DL      & SL     & DNN       & IV  &      \\ \hline
\cite{reilly2023validation}    & OC    & C    & C  & IM3 & MCD1 &      &     &         & DL      & SL     & MIA3G     & IV  &      \\ \hline
\cite{wang2023ensemble}        & EOC   & DQIC & C  & MM  & MCD1 &      &     &         & Hybrid2 & SL     & CNN       & IV  &      \\ \hline
\cite{wang2023interpretable}   & EOC   & DQIC & C  & IM2 & MCD1 &      &     &         & Hybrid2 & SSL\&SL & FCN\&CNN   & IV  & XAI2 \\ \hline
\cite{nasimian2023deep}        & OC    & DQIC & O  & TN  & MCD2 &      &     &         & Hybrid1 & SL     & TabNet & IV  & XAI2 \\ \hline
\cite{laios2022stratification} & HGSOC & C    & C  & MM  & SCD  & CDI  & FLI & NF-NBY2 & Hybrid1 & SL     & DNN   & IV  &      \\ \hline
\cite{laios2022factors}        & EOC   & C    & C  & TN  & SCD  & CDI  & FLI & NF-NBY1 & Hybrid1 & SL     & DNN\&O     & IV  & XAI1 \\ \hline
\cite{wang2022a-weakly}        & HGSOC & C    & C  & IM2 & MCD1 &      &     &         & DL      & SSL\&SL & FCN       & EV1 &      \\ \hline

\cite{laury2021artificial}     & HGSOC & C    & C  & IM1 & SCD  &      &     &         & DL      & SL     & DNN       & IV  &      \\ \hline

\cite{liu2021transynergy}      & OCO   & C    & CO & TN  & MCD2 & CODI & FLI & NF-NBY1 & DL      & UL\&SL  & CNN\&Tr    & IV  & XAI2 \\ \hline
\cite{yu2020deciphering}       & OC    & C    & CO & MM  & MCD1 & CODI & FLI & NF-NBY2 & DL      & SL     & CNN       & IV  &      \\ \hline

\cite{lei2022deep}             & EOC   & DQIC & C  & MM  & SCD  & CDI  & FLI & NF-NBY2 & Hybrid1 & UL\&SL  & CNN   & IV  &      \\ \hline
\cite{bote2022multivariate}    & OC    & DQIC & CO & TN  & MCD1 & CODI & FLI & NF-NBY1 & DL      & UL     & AE        & IV  & XAI2 \\ \hline

\end{tabular}%
}
\end{table*}

\begin{table*}[]
\caption{DL-based Data Analysis for OC Prognosis (P)}
\label{tab:P}
\resizebox{\textwidth}{!}{%
\begin{tabular}{@{}llllllllllllll@{}}
\toprule
 &  &  & \multicolumn{3}{l}{Data Features} & \multicolumn{3}{l}{Integrated Analysis} & \multicolumn{3}{l}{Anlaysis Methods \& DL details} &  &  \\ \midrule
Analysis                     & CST   & Use-of-DA & Source & Format & Repository & Data & Level & Method  & AM      & LT     & Algorithms & VM  & AIA \\ \midrule
\cite{kim2021prognostic}       & EOC   & SA   & C  & TN  & SCD  & CDI  & FLI & NF-NBY1 & DL      & SL     & DNN       & IV  &      \\ \hline
\cite{zhu2021sio}            & HGSOC & DQISA     & CO     & MM     & MCD2       & CODI & FLI   & NF-NBY1 & DL      & UL\&SL & CNN        & IV  &     \\ \midrule
\cite{wang2023ensemble}      & OC    & SA        & C      & IM2    & MCD1       &      &       &         & DL      & SSL    & CNN        & IV  &     \\ \midrule
\cite{liu2023deep}           & HGSOC & DQISA    & C      & IM1    & SCD        & CDI  & FLI   & NF-NBY2 & Hybrid2 & SL     & CNN        & IV  &     \\ \midrule
\cite{zhang2023assessing}    & HGSOC & SA       & CO     & TN     & MCD2       &      &       &         & Hybrid1 & UL\&SL & DNN\&O     & EV2 &     \\ \midrule

\cite{zheng2022preoperative} & HGSOC & SA       & C      & MM     & SCD        & CDI  & FLI   & NF-NBY1 & Hybrid2 & SL     & RNN\&Tr    & IV  &     \\ \midrule
\cite{LIU2022117643}        & EOC   & C         & C      & IM2    & MCD1       &      &       &         & Hybrid2 & SL     & DCAS       & IV  &     \\ \midrule
\cite{liu2022platelet}       & OC    & SA       & O      & TN     & MCD1       &      &       &         & Hybrid2 & SL     & DeepCox    & IV  &     \\ \midrule
\cite{yokomizo2022o3c}       & OC    & DQIC      & C      & IM2    & SCD        &      &       &         & DL      & SL     & CNN        & IV  &     \\ \midrule

\cite{tong2021integrating}   & OCO   & SA       & CO     & TN     & MCD1       & ODI  & FLI   & NF-NBY1 & Hybrid2 & UL\&SL & AE         & IV  &     \\ \midrule
\cite{hao2019interpretable}   & OCO   & SA       & CO     & TN     & MCD2       & CODI & FLI   & NB-NBY  & Hybrid2 & SL     & DNN        & IV  &     \\ \midrule
\cite{WANG2019171}           & HGSOC & SA       & C      & MM     & MCD1       & CDI  & FLI   & NF-NBY1 & Hybrid2 & SL     & DNN        & EV1 &     \\ \bottomrule
\end{tabular}%
}
\end{table*}

\subsection{\textbf{Use of Data Analysis (UDA)}}

The Use of Data Analysis (UDA) in cancer can be grouped into one or a
combination of the four capabilities: prediction (e.g., survival prediction), classification (e.g., subtypes, therapy stratification), association/clustering (e.g., subtypes clustering, dimensionality reduction), and optimisation. However, existing cancer studies generally use DA to do the first three.

\subsection{\textbf{UDA of Existing OC Data Analyses}}
As the distribution reveals in Figure~\ref{fig3d}, 45.83\% (44/96) studies used DA to classify (C) cancer-related activities (e.g., subtype classifications~\cite{Multi-omics-2021}, treatment stratification~\cite{guo2020deep}) and 34.4\% (33 out of 96) to improve data quality and classify (DQIC) (e.g., dimensionality reduction~\cite{Multi-omics-2021}, data compression~\cite{alshibli2019shallow}, style change~\cite{shin2021style}). Furthermore, 10.4\% (10/96) studies used DA to predict or perform SA (for example, SA~\cite{hao2019interpretable}), while the remaining 9 out of 96 used mainly for DQI, including one for cancer data generation (CDG~\cite{levine2020synthesis}).

\begin{figure}[hbt!]
\centering
\includegraphics[width=8.4cm]{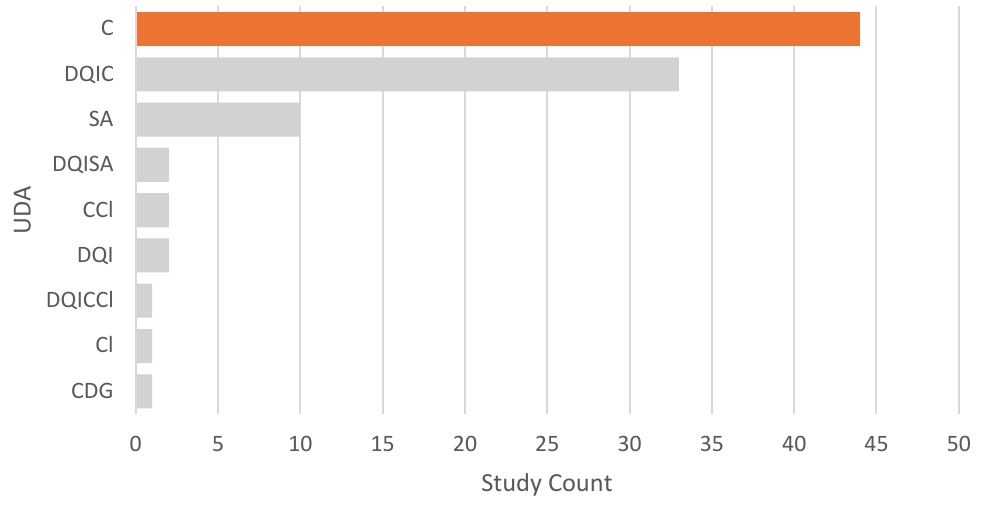}
\caption{UDA of Existing OC Data Analyses.}
\label{fig3d}
\end{figure}

\subsection{\textbf{Cancer Types}}

Generally, cancer experiences inter-tumour and intra-tumour heterogeneity~\cite{Tumour-heterogeneity10, Tumour-heterogeneity12, Tumour-heterogeneity13, Tumour-heterogeneity18}. Due to this heterogeneity, for any cancer (e.g., ovarian cancer), various tumour cells could exhibit a variety of morphological and phenotypic characteristics, including cellular morphology, gene expression, proliferation, and metastatic possibility~\cite{Tumour-heterogeneity10}. For example, ovarian carcinomas show various neoplasms with different risk factors, precursor lesions, aetiology, spread patterns, molecular profiles, clinical procedure, response to treatment, and prognosis~\cite{OV-types2021}. In this context, cancer typing or subtyping using pathological and molecular features of cancer is essential for individualised clinical decision-making~\cite{Therapy-prediction2021, OV-types2021}. Based on histopathology, immunoprofile, and molecular analysis, the WHO identified at least five major types of ovarian carcinomas: high-grade serous carcinoma (HGSC/HGSOC), endometrioid carcinoma (EC), clear cell carcinoma (CCC), low-grade serous carcinoma (LGSC/LGSOC), and mucinous carcinoma (MC)~\cite{OV-types2021}. Current ML/DL-based OC data analysis addresses one or more of these five types.


\subsection{Cancer Types of Existing OC Data Analyses}
\begin{figure}[hbt!]
\centering
\includegraphics[width=6.0cm]{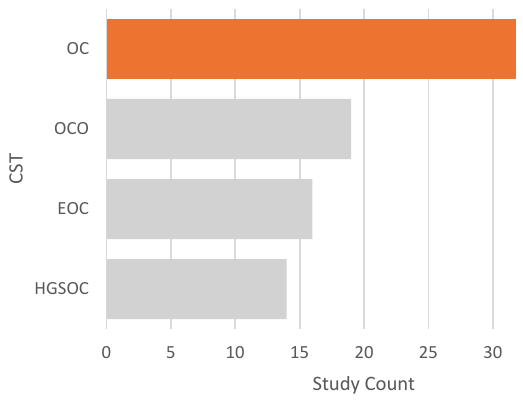}
\caption{Ovarian Cancer Subtype of Existing Studies.}
\label{fig3e}
\end{figure}

Existing OC DA are on only OC or OC with one or more other cancers (OCO). As illustrated in Figure~\ref{fig3e}, 80.2\% (77/96) studies are on OC, and the remaining 19. 8\% (19/96) studies are on OCO. However, most studies (47/77) on OC did not explicitly mention the type of ovarian carcinomas. On the contrary, only 16. 67\% (16/96) mentioned EOC as their type, and a further 14.6\% (14/96) studies mentioned HGSOC (a subtype of EOC), which signifies it as the most dangerous subtype of ovarian carcinoma.  

\subsection{Data Features (DF)}

\begin{figure*}[hbt!]
\centering
\includegraphics[width=17cm]{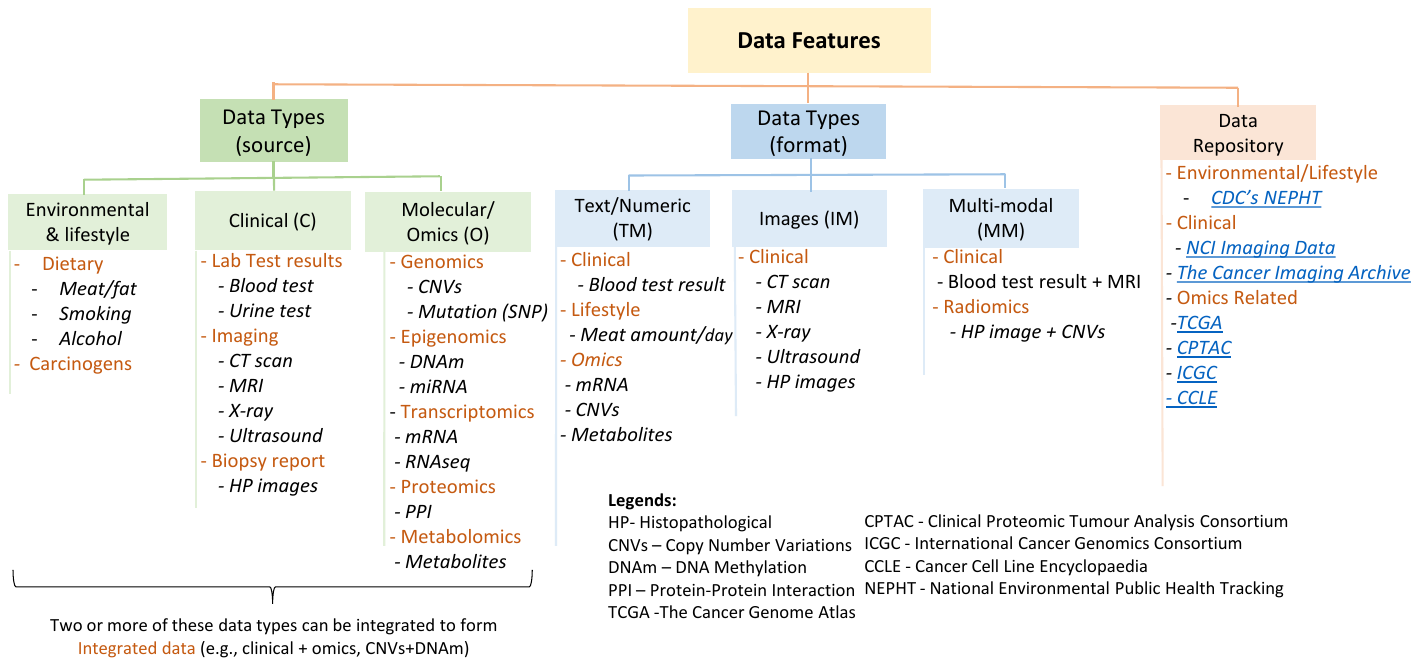}
\caption{Key features of Data used in Existing Data Analysis of Cancer}
\label{fig3}
\end{figure*}

Data are at the heart of any data analysis. The results of a data analysis depend on the used dataset's features and quality~\cite{GiGo-HealthResearch2016}. The key features of data used in ML/DL-driven cancer data analysis can be discussed from three perspectives (presented in Figure~\ref{fig3}) (i) data types (source-based), (ii) data types (data format based), and (iii) data repositories. Alongside these key features, data could be structured (e.g., lab tests, mRNA expression) and unstructured (e.g., clinical notes, reports)~\cite{zhang2020combining}. Generally, data preparation techniques take care of unstructured data and convert it into analysable forms. 

As shown in Figure~\ref{fig3}, source-based data can be (i) environmental and lifestyle data~\cite{CancerRiskPrediction2018, CancerRiskPrediction2021}, (ii) clinical data~\cite{Labtest-Based-2021,vazquez2018quantitative,DL-histology21}, and (iii) biological, especially omics data~\cite{Multi-omics-2021,OmiVAE2019a}. Similarly, cancer-related data can be one of three formats: (i) text or numeric~\cite{Labtest-Based-2021, Multi-omics-2021, Data-features2019}, (ii) image~\cite{gao2022deep, DL-histology21} and (iii) multi-modal or hybrid~\cite{ML-histology18, DL-histology21}. In cancer data analysis, most existing work relies on secondary data from repositories as primary data collection for ML/DL models is time-consuming and sometimes complex due to ethical clearance. It is recommended to use data from well-maintained and trustworthy repositories, as many exist. Figure~\ref{fig3} provides a few examples of widely used repositories and examples for each data type (both source- and format-based).

\subsection{Data Features of Existing OC Data Analyses}

\begin{figure}[hbt!]
\centering
\includegraphics[width=7.0cm]{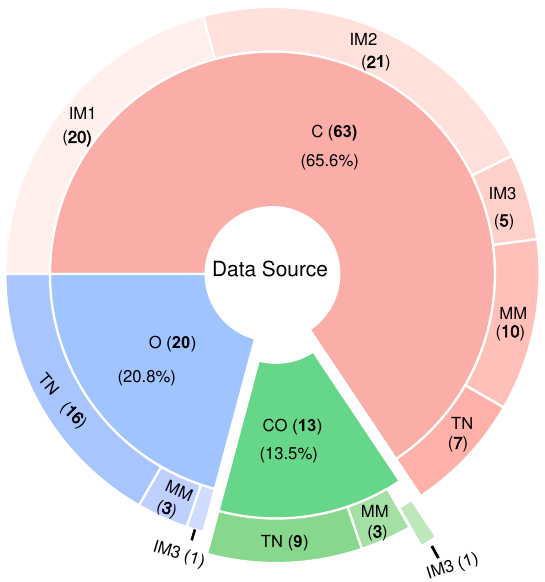}
\caption{Data-type (source) and Data-type (format) of Existing Studies.}
\label{fig3b}
\end{figure}

\textbf{Data Types}: Figure~\ref{fig3b} summarises the existing studies' data types (source) and their corresponding data formats. As depicted in the diagram, most of the studies, comprising 65.6\% (63/96), used clinical data (C) for their investigations. The omics data (O) constituted 20.8\% (20/96) of the studies, while a smaller portion, 13.5\% (13/96) of the studies, used data from clinical and omics (CO) sources in their analyses. A predominant factor contributing to the prevalence of clinical data-driven studies, accounting for 79\% (76/96) of the studies, is the availability of relevant and accessible data sets.

Figure~\ref{fig3b} illustrates that the predominant data format used in existing studies is images, accounting for 50\% or 48 out of 96 studies. In particular, 96\% (46 out of 48) of these images originate from clinical sources such as Radiological/ IM1~\cite{hu2023deep,laury2021artificial,liu2023deep}, pathomics or histopathological images/IM2~\cite{wang2022automated,wang2023interpretable, wang2023ensemble}, and other types such as mass spectrometry~\cite{reilly2023validation, Pastuszak_2021}). 
Of the remaining 50\% (48/96) of studies, 64\% (32/50) use text or numeric (TN) data, which are drawn from all three sources but are predominantly from omics. Among these, the rest, 16.67\% (16/96), integrate multimodal (MM) data, combining elements like images and text in their analyses. The prevalence of image data is mostly due to the availability of clinical imaging equipment and the advancement of DL-based image processing.

\begin{figure}[hbt!]
\centering
\includegraphics[width=7.0cm]{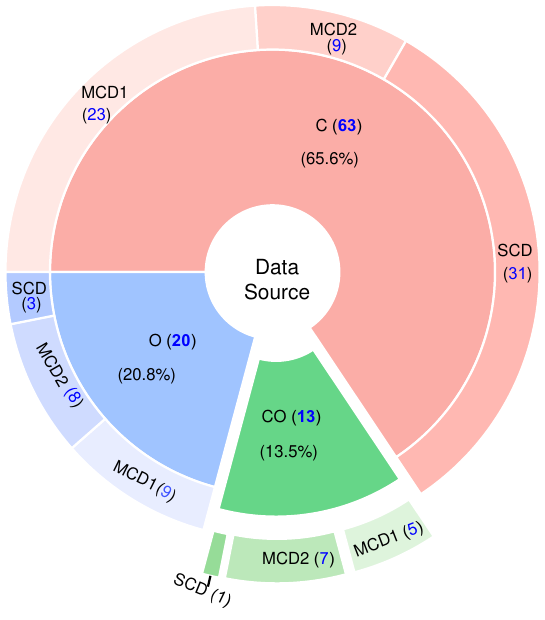}
\caption{Data-type (source) and Repository of Existing Studies}
\label{fig3c}
\end{figure}

\begin{figure}[hbt!]
\centering
\includegraphics[width=7.2cm]{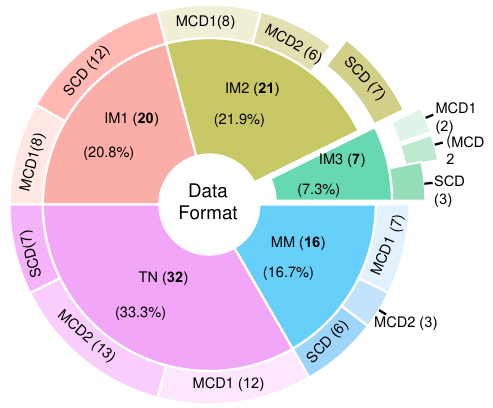}
\caption{Data-type (format) and Repository of Existing Studies}
\label{fig3d}
\end{figure}

\textbf{Repository/Databases of Existing Studies}: Generally, repositories/databases used in cancer studies can be single-centred (SCD) (e.g., one clinic or hospital) or multi-centred (MCD), data coming from multiple clinics or hospitals. In this study, we have further divided MCD into MCD1 and MCD2 to visualise the heterogeneity of the data sources or databases used. MCD1 means that the data come from multiple centres or clinics but from the same country, and in the case of MCD2, the data come from different clinics/centres and multiple countries or demography. Figures~\ref{fig3c} and~\ref{fig3d} visualise the distribution of the studies in terms of these three types of databases (SCD, MCD1, and MCD2) for the data sources and formats. As seen in Figures~\ref{fig3c} and~\ref{fig3d}, most of the studies (63.5\% or 61/96) used MCD and the remaining 36.5\% (35/96) studies used SCD. However, most of the multi-centred studies (60.65\% or 37/61) rely on data from two or more centres but are limited to a country or population, which could make the DL models or solution biased to that population, limiting the clinical deployment of the solution. In the case of clinical data (C), the study distribution of SCD and MCD is well balanced (31 vs 32 or 49.2\% vs 50.8\%), while most (88\% or 29/33) of the omics and CO data-driven analyses used MCD. On the other hand, from Figure~\ref{fig3d}, it is evident that the distribution of existing studies in terms of data formats and their associated databases or repositories exhibits heterogeneity. For instance, among the image-based studies, SCD was utilised by 45.8\% (22/48), MCD1 by 37.8\% (18/48), and the remaining 16.67\% (8/48) employed MCD2. On the contrary, in the case of studies relying on TN and MM data, the majority, with 73\% (35/48), used MCD, while the remaining 13 studies used SCD.

\begin{figure*}[!hbt]
\centering
\includegraphics[width=17.5cm]{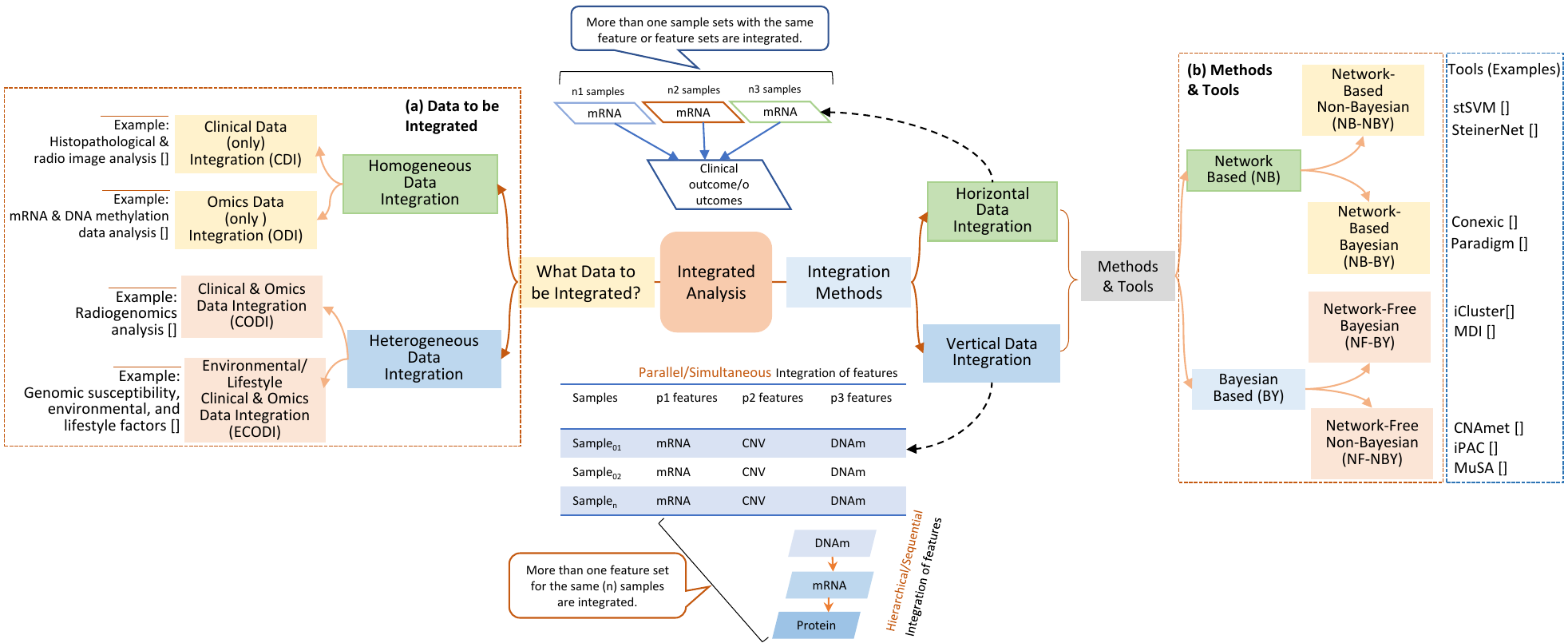}
\caption{Data Integration: (a) What data to integrate, and (b) Method \& Tools for integration }
\label{fig4}
\end{figure*}

\subsection{Integrated Analysis (IA)}
\label{subsec-4}

Cancer is characterised by inter-tumour and intra-tumour heterogeneity~\cite{Tumour-heterogeneity18}, which disrupts cellular processes at various molecular levels, such as DNA, RNA, proteins, and metabolites~\cite{hu2013multi}.
 As a result of its heterogeneity, different cancerous tumour cells may display varied morphological and phenotypic characteristics~\cite{Tumour-heterogeneity10}. Hence, one type of clinical data-based analysis (e.g., image or blood test results) may not accurately diagnose the disease. It is important to note that molecules at various levels are interconnected in modifying cellular functions~\cite{hu2013multi,cheng2017pattern}. Therefore, an analysis that only focuses on one level is not enough to fully comprehend the complex nature of cancer. To diagnose and fully understand the cellular defects that cause and contribute to cancer, an integrated analysis of data from various clinical and "omics" levels is necessary. Data integration for cancer analysis, especially the integration of omics data and corresponding analysis, is a very active research area~\cite{huang2017more, Data-integration-2017, Chakraborty2018, Data-Integration-methods2019, Data-Integration-methods2020, Musa-2021,jendoubi2021approaches}. Many research works have been published in this area. 

Currently, integrated studies typically combine data from multiple sources in two ways (Figure~\ref{fig4}): (1) by integrating similar types of clinical or 'omics' data from various studies (samples) or groups (horizontal data integration) and (2) by integrating different types of 'omics' or clinical data (features) for the same group of samples (vertical data integration).

\subsubsection{Objectives of Integrated Analysis}
Generally, integrated data analysis could be useful in one or more of the following: 
\begin{itemize}
    \item Clustering or Classification: disease subtyping and classification using clinical or multi-omics data,
    \item Prediction: prediction of biomarkers for different applications (e.g., diagnostics and prognostic genes for diseases), and
    \item Molecular mechanism: gaining knowledge on disease biology.
\end{itemize}

\subsubsection {Data to be Integrated}

Data to be integrated into a cancer analysis are heterogeneous in types (e.g., environmental, lifestyle clinical \& omics Data) and formats or modalities (e.g., text, image). Integration of these data can be categorised into two categories: (i) homogeneous data (e.g., omics or clinical only data ) integration and heterogeneous data integration (e.g., clinical and omics data). Figure~\ref{fig4} (a) presents these categories of data integration with examples.  

\subsubsection {Integration strategy or Levels}
Based on the level of data abstraction, integration levels or strategies can be categorised into 3~\cite{jendoubi2021approaches}:

\begin{itemize}
    \item Early or Data Level Integration (ELI/DLI): ELI combines or integrates two data sets by concatenating them into a single data set. ELI can provide higher accuracy at the cost of complexity.
    \item Medium- or Feature-Level Integration (FLI): The FLI method first extracts features from the data sets and then integrates them. FLI methods are less informative than ELI.
    \item Late or Decision Level Integration (DeLI): The DeLI method integrates decisions or data models into a high-level model. DeLI is simpler but less informative than ELI and FLI.
\end{itemize}

\begin{figure}[hbt!]
\centering
\includegraphics[width=6.0cm]{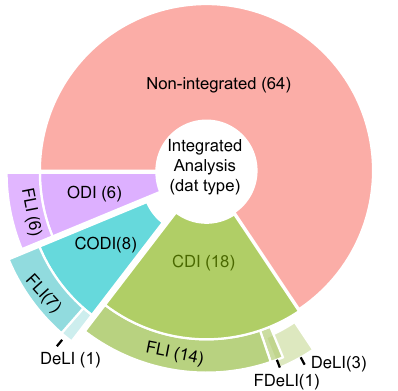}
\caption{Data-type (format) and Integration Levels}
\label{fig4b}
\end{figure}

\subsubsection{Methods and Tools}
Cancer data integration (DI) methods and tools are diverse~\cite{huang2017more, Data-integration-2017, Chakraborty2018, Data-Integration-methods2019, Data-Integration-methods2020, Musa-2021}, and most of these, especially omics data integration methods and tools use network-based (NB) (e.g., protein-protein interactions) or Bayesian (BY) approach or a combination of NB and BY~\cite{Data-integration-2017, Chakraborty2018, Data-Integration-methods2019, Data-Integration-methods2020}. Based on these two approaches, existing DI methods and tools can be categorised into one of the following. Also, Figure~\ref{fig4} (b) summarises these methods and tools with examples.

\begin{itemize}
    \item Network-based non-Bayesian (NBNBY): 
    Generally, methods in this category either utilise molecular interaction data or build networks based on correlation analysis. SteinerNet~\cite{tuncbag2012steinernet,cun2014stSVM} and stSVM~\cite{tuncbag2012steinernet,cun2014stSVM} examples of works or tools that utilise NBNBY, especially the study of a multi-weighted graph that has multi-omics knowledge.
    
    \item  Network-based Bayesian (NB-BY): Methods in this category are categorised as network-based and Bayesian, which mainly rely on Bayesian networks (BNs). BNs are probabilistic models consisting of a graph and a local parametric or non-parametric probability model. Paradigm~\cite{vaske2010paradigm} is an exemplary method for multi-omics data integration that uses patient-specific pathway activities. Conexic~\cite{akavia2010integrated} is another example of a Bayesian network-based algorithm tool.
    
    \item Network-free Bayesian (NF-BY): NF-BY-based DI methods and tools can be parametric, "strict" Bayesian, non-parametric, or distribution-free. The parametric BY DI method presumes that the prior probability distribution obeys a model reliant on one or more parameters. On the other hand, non-parametric or distribution-free DI approaches consider that the priors do not follow a given family of probability distributions depending on one or more parameters, as this family would be too big. iCluster~\cite{shen2009iCluster} is an example of a parametric BY-based DI method and MDI~\cite{kirk2012MDI} is an example of non-parametric BY-based DI method. 
    
    \item Network-free non-Bayesian (NF-NBY): NF-NBY-based DI methods and tools rely on any approach other than NB or BY (e.g., simple concatenation of data sets). Generally, these methods and tools are designed to develop specific types of omics and may use different approaches. iPAC~\cite{aure2013iPAC} and CNAmet~\cite{louhimo2011cnamet} are examples of NF-NBY-based DI tools. 

\end{itemize}

\begin{figure}[hbt!]
\centering
\includegraphics[width=8.5cm]{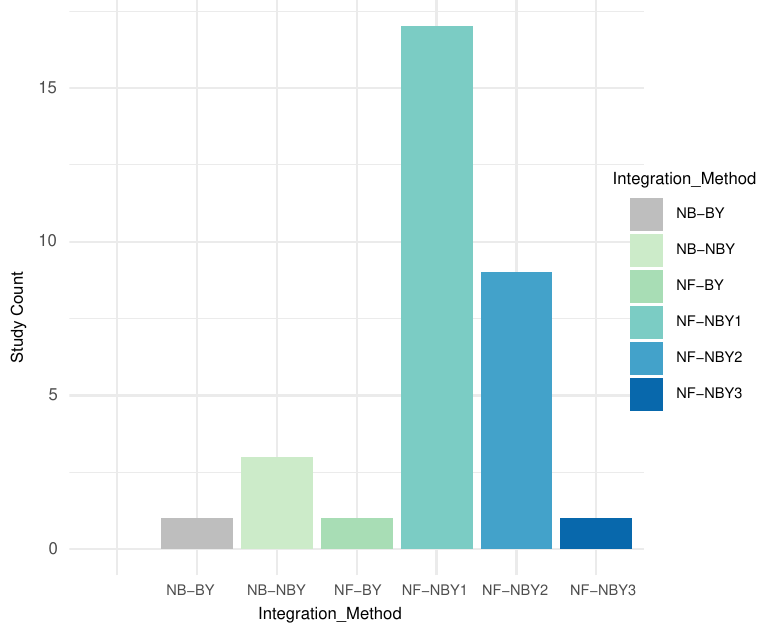}
\caption{Integration Methods of Existing Integrated Studies}
\label{fig4c}
\end{figure}

\subsection{Integrated Analysis in Existing OC Data Analyses}
Figures~\ref{fig4b} and~\ref{fig4c}, and Table~\ref{tab:IA} summarise the existing integrated analyses in terms of their key features. In Figure~\ref{fig4b} and Table~\ref{tab:IA}, it is evident that a significant number (32/96) of the studies harnessed integrated data to achieve various objectives. These objectives included disease subtype clustering/classification~\cite{Multi-omics-2021}, treatment startification~\cite{laios2022stratification}, biomarkers prediction~\cite{kim2021prognostic, bote2022multivariate}  or treatment efficacy~\cite{laios2022factors}, and gaining more information about the biology of OC~\cite{wei2023associating}. Most of these studies (24/32) integrated homogeneous data (CDI~\cite{wei2023associating} or ODI~\cite{tong2021integrating}); only eight considered heterogeneous data or CODI~\cite{liu2021transynergy}. Only two~\cite{guo2020deep, Multi-omics-2021} of these eight performed integrated tri-omics analysis without proteomics or metabolomics. Figure~\ref{fig4b} also illustrates that most (27/32 or 84.4\% of the existing studies integrated data at the feature level, as the data sets used are readily available with features (e.g., gene expression values, survived months).

On the other hand, Figure~\ref{fig4c}, confirms that most studies (84.4\% or 27/32) used NF-NBY (NF-NBY1/concatenation, NF-NBY2/correlation and NF-NBY3/ensemble). Only 3 of the remaining 5 studies (e.g. \cite{sun2022xgbg, hao2019interpretable}) used NB-NBY, and the remaining two used the Bayesian approach (NB-BY~\cite{ye2021ovarian} and NF-BY~\cite{sengupta2022deep}) approach to integrate their data. 

\begin{table*}[]
\caption{DL-based Integrated Data Analysis of OC }
\label{tab:IA}
\resizebox{\textwidth}{!}{%
\begin{tabular}{@{}lllllllllllllll@{}}
\toprule
 &
   &
   &
   &
  \multicolumn{3}{c}{Data Features} &
  \multicolumn{3}{l}{Integrated Analysis} &
  \multicolumn{3}{l}{Anlaysis Methods \& DL details} &
   &
   \\ \midrule
Analysis &
  CSD &
  CST &
  Use-of-DA &
  Source &
  Format &
  Repository &
  Data &
  Level &
  Method &
  AM &
  LT &
  Algorithms &
  VM &
  AIA \\ \midrule
\cite{wei2023associating}      & DD  & EOC   & C      & C  & MM  & MCD2 & CDI  & DeLI  & NF-NBY3 & Hybrid2 & SL     & CNN       & EV2 &      \\ \midrule
\cite{reilly2022analytical}    & DD  & OC    & C      & C  & TN  & SCD  & CDI  & FLI   & NF-NBY1 & DL      & SL     & MIA3G     & IV  &      \\ \midrule
\cite{chen2022deep}            & DD  & OC    & C      & C  & IM1 & SCD  & CDI  & FDeLI & NF-NBY1 & DL      & SL     & CNN       & IV  &      \\ \midrule
\cite{sengupta2022deep}        & DD  & OC    & C      & C  & MM  & SCD  & CDI  & DeLI  & NF-BY   & DL      & SL     & CNN       & IV  &      \\ \midrule
\cite{xie2021diagnosis}        & DD  & OCO   & C      & C  & IM1 & SCD  & CDI  & FLI   & NF-NBY1 & DL      & SL     & CNN       & IV  &      \\ \midrule
\cite{wang2021evaluation}      & DD  & OC    & C      & C  & MM  & SCD  & CDI  & DeLI  & NF-NBY2 & DL      & SL     & DNN       & IV  &      \\ \midrule
\cite{cancers12092373}          & DD  & EOC   & DQIC   & C  & TN  & SCD  & CDI  & FLI   & NF-NBY1 & DL      & SL     & CNN       & IV  &      \\ \midrule
\cite{vazquez2018quantitative} & DD  & OC    & C      & C  & TN  & MCD1 & CDI  & FLI   & NF-NBY2 & Hybrid2 & SL     & RNN       & IV  &      \\ \midrule
\cite{Liang2015}               & DD  & OCO   & SA    & CO & TN  & MCD2 & CDI  & FLI   & NF-NBY2 & DL      & UL\&SL & DBN       & IV  &      \\ \midrule
\cite{ghoniem2021multi}        & DD  & OC    & C      & CO & MM  & MCD1 & CODI & DeLI  & NF-NBY1 & DL      & SL     & CNN\&LSTM & IV  &      \\ \midrule
\cite{chen2021prediction}      & DD  & OC    & DQIC   & CO & TN  & SCD  & CODI & FLI   & NF-NBY1 & Hybrid1 & UL\&SL & GCN       & IV  &      \\ \midrule
\cite{sun2022xgbg}             & DD  & OC    & DQIC   & O  & TN  & MCD2 & ODI  & FLI   & NB-NBY  & DL      & SL     & GCN       & IV  &      \\ \midrule
\cite{petrovsky2021managing}   & DD  & OCO   & C      & O  & MM  & SCD  & ODI  & FLI   & NF-NBY1 & DL      & SL     & CNN       & IV  &      \\ \midrule
\cite{ye2021ovarian}           & DD  & OC    & C      & O  & TN  & MCD2 & ODI  & FLI   & NB-BY   & DL      & UL\&SL & GAT\&DNN  & IV  &      \\ \midrule
\cite{guo2020deep}             & DD  & OC    & CCl    & O  & TN  & MCD2 & ODI  & FLI   & NF-NBY1 & Hybrid2 & UL\&SL & AE        & EV2 &      \\ \midrule
\cite{mallavarapu2020pathway} &
  DD &
  OCO &
  Cl &
  O &
  TM &
  MCD1 &
  ODI &
  FLI &
  NB-NBY &
  Hybrid1 &
  UL\&SL &
  RBM\&DBN &
  IV &
   \\ \midrule
\cite{avesani2022ct}           & DDP & HGSOC & SA    & C  & IM1 & MCD1 & CDI  & FLI   & NF-NBY1 & Hybrid2 & SL     & CNN       & EV1 &      \\ \midrule
\cite{Multi-omics-2021} &
  DDP &
  HGSOC &
  DQICCl &
  CO &
  TM &
  MCD1 &
  CODI &
  FLI &
  NF-NBY2 &
  Hybrid2 &
  UL\&SL &
  VAE\&DNN &
  IV &
   \\ \midrule
\cite{liu2023deep}             & P   & HGSOC & DQISA & C  & IM1 & SCD  & CDI  & FLI   & NF-NBY2 & Hybrid2 & SL     & CNN       & IV  &      \\ \midrule
\cite{zheng2022preoperative}   & P   & HGSOC & SA    & C  & MM  & SCD  & CDI  & FLI   & NF-NBY1 & Hybrid2 & SL     & RNN\&Tr   & IV  &      \\ \midrule
\cite{WANG2019171}             & P   & OC    & SA    & C  & MM  & MCD1 & CDI  & FLI   & NF-NBY1 & Hybrid2 & SL     & DNN       & EV1 &      \\ \midrule
\cite{zhu2021sio}              & P   & OC    & DQISA  & CO & MM  & MCD2 & CODI & FLI   & NF-NBY1 & DL      & UL\&SL & CNN       & IV  &      \\ \midrule
\cite{hao2019interpretable}    & P   & EOC   & SA    & CO & TN  & MCD2 & CODI & FLI   & NB-NBY  & Hybrid2 & SL     & DNN       & IV  &      \\ \midrule
\cite{tong2021integrating}     & P   & HGSOC & SA    & CO & TN  & MCD1 & ODI  & FLI   & NF-NBY1 & Hybrid2 & UL\&SL & AE        & IV  &      \\ \midrule
\cite{laios2022stratification} & TM  & HGSOC & C      & C  & MM  & SCD  & CDI  & FLI   & NF-NBY2 & Hybrid1 & SL     & DNN       & IV  &      \\ \midrule
\cite{laios2022factors}        & TM  & EOC   & C      & C  & TN  & SCD  & CDI  & FLI   & NF-NBY1 & Hybrid1 & SL     & DNN\&O    & IV  & XAI1 \\ \midrule
\cite{han2022cell} & TM  & OC    & DQIC   & C  & IM2 & MCD2 & CDI  & FLI   & NF-NBY2 & DL & SSL\&SL & CNN & IV  & \\ \midrule
\cite{hwangbo2021development}  & TM  & HGSOC & C      & C  & TN  & MCD1 & CDI  & FLI   & NF-NBY1 & Hybrid2 & SL     & DNN\&O    & IV  &      \\ \midrule
\cite{liu2021transynergy}      & TM  & OCO   & C      & CO & TN  & MCD2 & CODI & FLI   & NF-NBY1 & DL      & UL\&SL & CNN\&Tr   & IV  & XAI2 \\ \midrule
\cite{yu2020deciphering}       & TM  & OC    & C      & CO & MM  & MCD1 & CODI & FLI   & NF-NBY2 & DL      & SL     & CNN       & IV  &      \\ \midrule
\cite{lei2022deep}             & TMP & EOC   & DQIC   & C  & MM  & SCD  & CDI  & FLI   & NF-NBY2 & Hybrid1 & UL\&SL & CNN       & IV  &      \\ \midrule
\cite{kim2021prognostic}       & TMP & EOC   & SA     & C  & TN  & SCD  & CDI  & FLI   & NF-NBY1 & DL      & SL     & DNN       & IV  &      \\ \midrule
\cite{bote2022multivariate}    & TMP & OC    & DQIC   & CO & TN  & MCD1 & CODI & FLI   & NF-NBY1 & DL      & UL     & AE        & IV  & XAI2 \\ \bottomrule
\end{tabular}%
}
\end{table*}

\subsection{Data Analysis Method (DAM)}
Medical data, including cancer data, can be analysed using~\cite{rajula2020Stats-ML} (i) statistical methods (e.g., linear regression, Cox regression), (ii) ML/DL based methods (e.g., OmiVAE~\cite{OmiVAE2019a}), and (iii) hybrid using ML and DL or a combination of statistical method and ML/DL (e.g. MMD-VAE4Omics ~\cite{Multi-omics-2021}). ML/DL-based analyses offer greater flexibility and scalability than traditional statistical methods, making them suitable for different tasks, including diagnosis, classification, and prognosis. As a result, they might be better suited to advanced big data industries like drug discovery, omics/multi-omics, and personalised medicine. In contrast, conventional statistical methods are more usable when the samples are significantly more than the variables count under study and prior information on the topic under investigation is significant. Although ML/DL-based cancer survival analysis has become available in recent years (e.g., SALMON~\cite{SALMON-ANN-survival2019}), combining the two methods would be preferable to one of the methods.

\begin{figure}[hbt!]
\centering
\includegraphics[width=8.0cm]{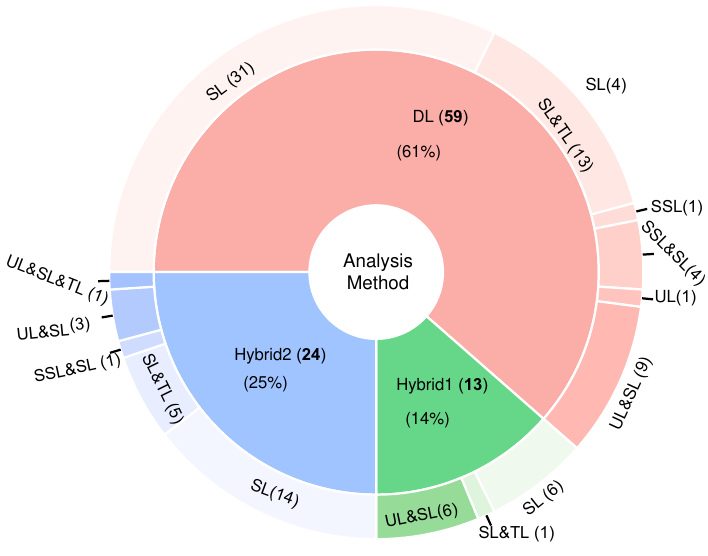}
\caption{Analysis Methods and L-T of Existing Works.}
\label{fig4d}
\end{figure}

\subsection{Analysis Methods of Existing OC Data Analyses}

We divided the data analysis methods (DAM) of the existing studies into (i) DL, (ii) Hybrid1 (DL with ML), and (iii) Hybrid2 (DL with Statistical). As visualised in Figure~\ref{fig4d}, most studies (61.45\% or 59/96) used DL-based methods to analyse their data, while the rest (38.55\% or 37/96) used one of the two hybrid approaches, of which 13.55\% (13/96) used Hybrid1 and 25\% (24/96) used Hybrid2. Figure~\ref{fig4d} also highlights that most studies (51/96) used supervised learning (SL) techniques to analyse their OC data.

\subsection{ML/DL Techniques and Algorithms (ML/DLT\&A)}

There are four ML/DL techniques used in existing cancer data analyses, namely (i) supervised learning (SL), (ii) unsupervised learning (UL), (iii) semi-supervised learning and (iv) reinforcement learning (RL)~\cite{eckardt2021reinforcement}. In SL, a \textit{labelled} training dataset is utilised to calculate or map the input data to the preferred output. On the other hand, UL techniques consider only \textit{unlabelled} training data without assuming a particular output(s) during the learning process. Instead, the learning model finds patterns or groups within the training data. Clustering, dimensionality reduction and rule-based association learning are the main unsupervised activities, whereas SL techniques mainly do classifications and regression using linear and non-linear algorithms (e.g., ANN, DL). SL techniques need massive labelled data to achieve high classification/prediction accuracy. However, the reality is that we often have only a few labelled data and a much more significant amount of unlabelled data in medical domains. Moreover, UL techniques struggle in performance, and transfer learning (TL) techniques need lesser labelled data than SL, but the number of labelled data required is still quite large~\cite{2021unsupervised,khalifa2019deep}. 

Semi-supervised learning (SSL) uses labelled and unlabelled data and is a low-cost alternative to arduous and sometimes unworkable sample labelling. Recently, many cancer-related works have used SSL for classification~\cite{SLL2016, SSL2021} and clustering~\cite{SSL2018, SSL2018a}. SL, UL, and SSL techniques are currently the most popular techniques for cancer data analysis using retrospective data. However, these techniques must better capture the dynamic circumstances in which an individual patient and clinician find themselves during oncologic treatment. Since patient and environmental variables change over time, the sequential basis of RL makes it a better fit for capturing the dynamics of oncologic therapy in real-world (prospective) scenarios. Like gameplay and autonomous driving, RL-based ML/DL techniques, which can be value-based, policy-based, and model-based, are also gaining popularity in cancer data analysis~\cite{balaprakash2019scalable,daoud2020q}.

\begin{figure}[hbt!]
\centering
\includegraphics[width=8.0cm]{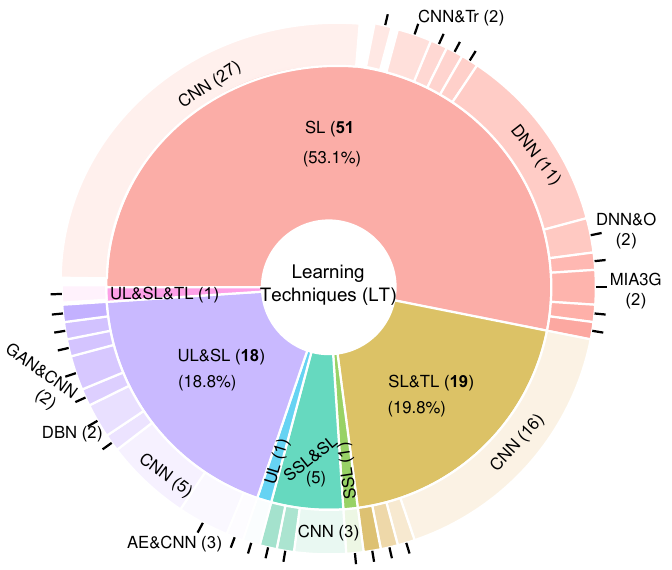}
\caption{LT and Algorithms of Existing Works.}
\label{fig4e}
\end{figure}

\begin{figure}[hbt!]
\centering
\includegraphics[width=8.0cm]{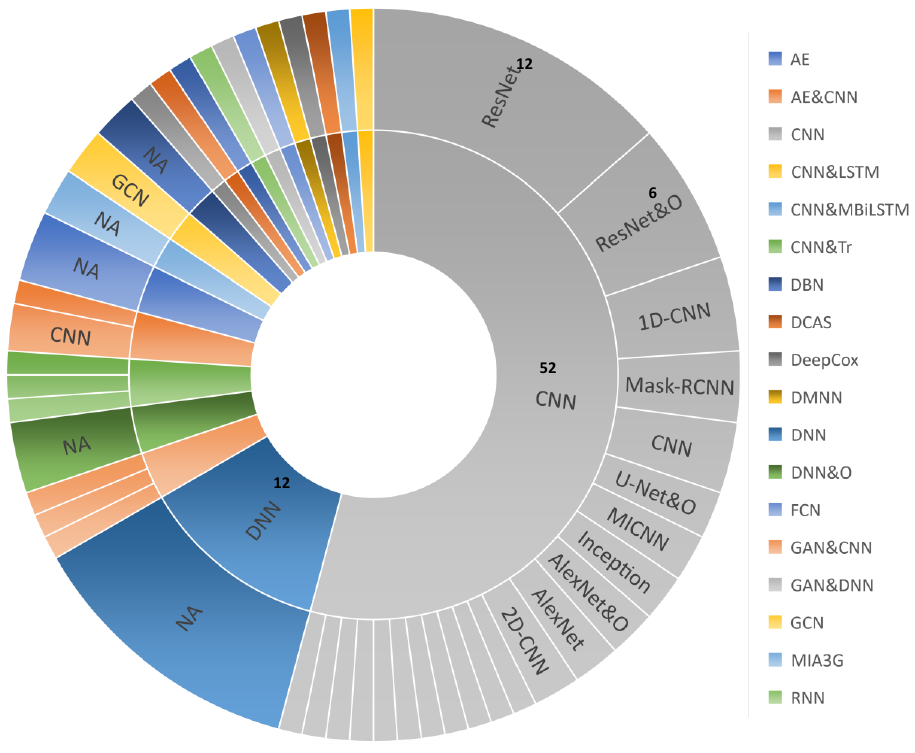}
\caption{LT and Algorithms of Existing Works.}
\label{fig4f}
\end{figure}

\subsection{ML/DLT\&A of Existing OC Data Analyses}

Figures~\ref{fig4e} and~\ref{fig4f} present the distribution of existing studies in terms of ML/DLT\&A. In Figure~\ref{fig4b}, it is evident that most studies (53.1\% or 51/96) used SL techniques, and the rest of the studies used a combination of technologies, including SL\&TL (19.8\% or 19/96), and UL\&SL (18.8\% or 18/96). The dominance of SL technique-based studies is due to the availability of labelled data (e.g., TCGA database). Similarly, the use of transfer learning (TL) is increasing due to the availability of many pre-trained and reliable models (e.g., ResNet~\cite{xie2021diagnosis}, U-Net~\cite{gonzalez2021characterization}). As seen in Figure~\ref{fig4f}, CNN is the dominant (52 out of 96) DL algorithm in the existing studies, and this is mainly due to the prevalence of image-based analysis (48 of 96 studies). ResNet is the dominant (18 out of 96 or 34.6\% studies) variant of CNN used in the existing OC data analyses. 

\begin{figure*}[hbt!]
\centering
\includegraphics[width=17.5cm]{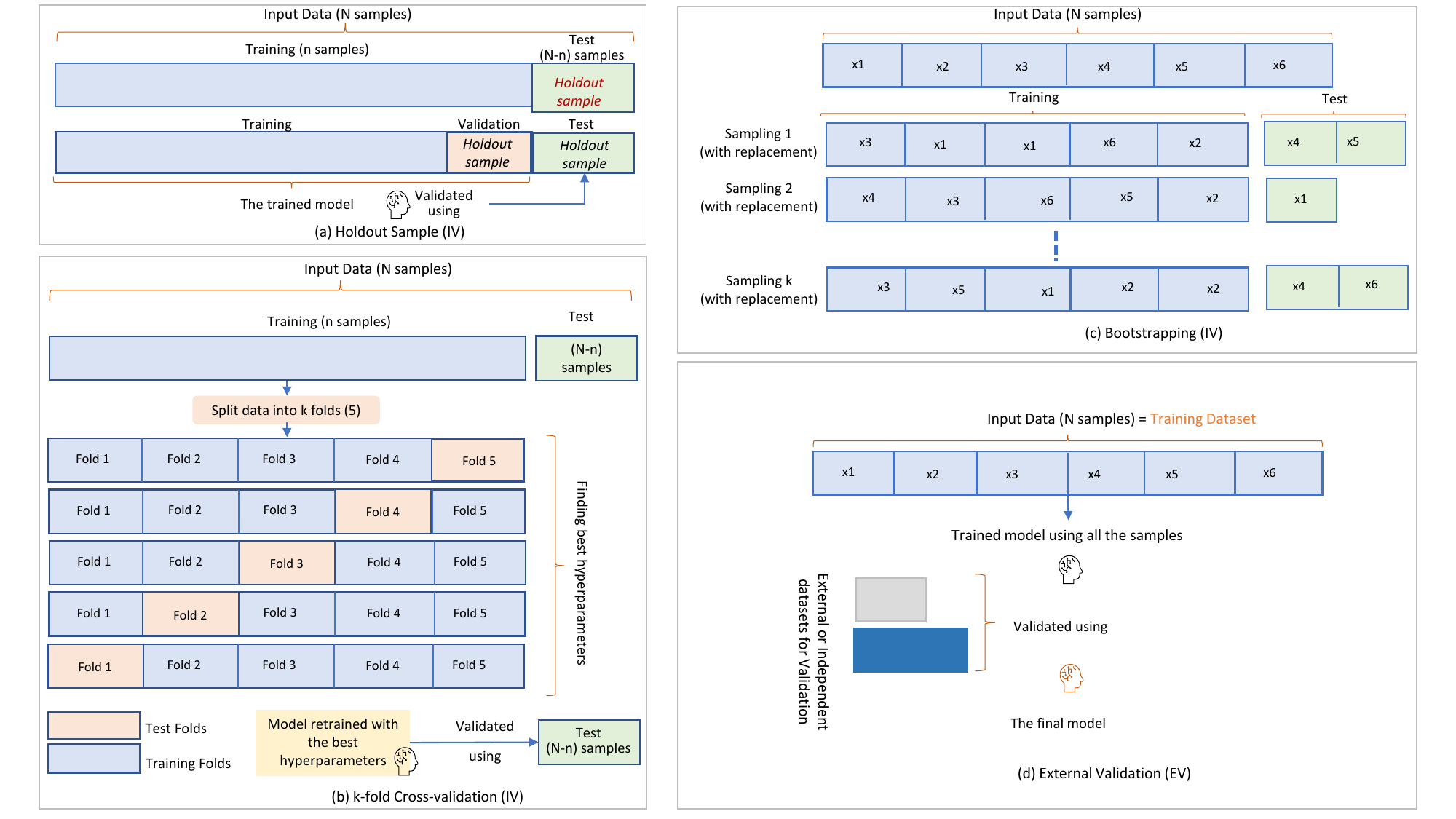}
\caption{ML/DL Model Validation processes: (a) IV using Holdout sample, (b) IV using k-fold Cross-validation, (c) IV using Bootstrapping, and (d) EV using independent and external data sets.}
\label{fig6}
\end{figure*}

\subsection{Models Validation Method (VM)}

In any data analysis, learned ML/DL models need to be evaluated through a validation process. Typically, a validation method estimates the accuracy, sensitivity and specificity of data not seen in the training set. Generally, a validation process can be categorised into "internal" and "external"~\cite{EV2020, DL-histology21}. Internal validation (IV) methods are economical and widely used validation methods (VM) that can use one of the three techniques: (i) holdout sample, (ii) cross-validation, and (iii) bootstrapping. The holdout sampling technique splits an input dataset using a ratio (e.g., 60/40, 70/30, or 80/20) into two distinct data sets labelled as a training and a test dataset (Figure~\ref{fig6} (a)). The training dataset can be split into training and validation/tuning (hyper-parameters) data sets.
In contrast, we can do the cross-validation a few different ways, but for non-time series data, one of the most popular and simple techniques is K-fold cross-validation. As shown in Figure~\ref{fig6} (b), in cross-validation of k times (generally k = 5 or 10), the input data is split into k sets of equal size (folds); one set is considered the validation set and the rest are used as the training set. The procedure is duplicated until each part/fold has been used at least once as testing/validation data. In Bootstrapping (Figure~\ref{fig6} (c)), the remaining data that were not chosen for training are utilised for testing, while the training dataset is selected at random with replacement. The model's k-fold cross-validation and the bootstrapping error rate are averaged over all the iterations. IV techniques are highly diverse and tuneable, despite being regarded as the industry standard and widely used. The analyst can choose the procedural subtype (e.g., k-fold, bootstrap) and the parameters. Most cancer data analysis uses IV techniques, but because of possibly biased training data and the complex process, they cannot ensure the quality of an ML/DL model. External validation (EV) is required for the quality of an ML/DL model, such as robustness and generalisability~\cite{siontis2015EV, EV2020a, DL-histology21}.

Unlike IV, EV uses external or independently emanated data sets to evaluate the performance of a trained model. As a result, EV is often known as independent validation. As the data set comes from a separate origin, any incorrectly chosen feature set due to quirks in the input training data (such as technical or sample bias) would probably fail. As a result, better EV performance can be considered evidence of generalisability~\cite{EV2020a, EV2020}. We can divide EV into two: EV1-if the test/validation data set(s) comes from different sources (e.g., clinics/hospitals) but from the same country or demography as the training data, and EV2-if the test/validation data set(s) come from different sources and different country or demography than the training data.

\begin{figure}[hbt!]
\centering
\includegraphics[width=6.0cm]{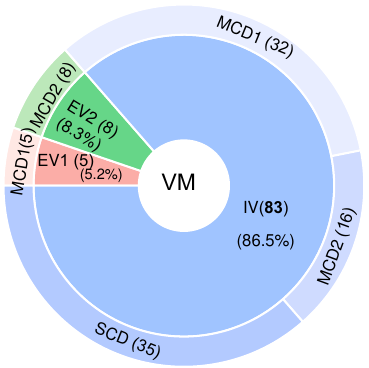}
\caption{Validation methods and Repositories of existing works.}
\label{fig:5c}
\end{figure}

\subsection{VMs of Existing OC Data Analyses}

Figure~\ref{fig:5c} illustrates the distribution of the existing OC data analyses in terms of their VM employed. The figure demonstrates that 86.5\% (83 out of 96) of the studies utilised IV to validate their DL models. Five of the remaining thirteen studies opted for EV1, while the remaining eight chose EV2. Furthermore, the figure highlights a noteworthy pattern in validation practises. All studies that underwent external validation utilised multi-centred (MCD1 or MCD2) data sets. On the contrary, many studies (48 of 83 or 50\% of the total of 96 studies) relied on multicentred data sets for internal validation. This tendency might arise from the challenge of obtaining a sufficient volume of samples for training DL algorithms, forcing researchers to employ multi-centred data sets for algorithm training rather than testing purposes.

\subsection{Overview of AI Assurance (AIA)}

AIA is a pivotal research field dedicated to safeguarding our society's fundamental pillars while developing and implementing advanced AI systems. AIA has recently gained renewed attention, with researchers, policymakers, and business leaders using the term. However, despite its increasing usage, there is still no definitive agreement on its precise definition. According to Batarseh et al.~\cite{batarseh2021survey}:\\

"A process that is applied at all stages of the AI engineering lifecycle, ensuring
that any intelligent system is producing outcomes that are valid, verified,
data-driven, trustworthy and explainable to a layman, ethical in the context of
its deployment, unbiased in its learning, and fair to its users."

Although there is no globally accepted definition of AI assurance, the above definition is generic enough to capture most aspects of assurance. It is pertinent to all AI domains and subdomains. As nations are racing to develop AI-based solutions for almost every field, including medical and driverless cars like critical fields, a race to progress in AI assurance is needed. Recently, the European Commission (EC) proposed rules and actions for excellence and trust for AI systems across the EU~\cite{AIA-EC2021}, and the UK has established a roadmap for an effective AI assurance ecosystem~\cite{AIA-UK2021}. Also, very recently~\cite{AIA-US2023} OpenAI CEO called on the US Congress to regulate AI. Individual countries or continents are making some initiatives, but a global consensus is needed for adequate AI assurance. 

Most existing AI systems have weaknesses in their ability to provide assurance. Therefore, unlike traditional software, assurance cannot be an afterthought in AI solutions. Instead, the assurance should be by design and included in the overall learning process of any intelligent agent, algorithm, or environment~\cite{AIA-Consesus2022}. In particular, all stakeholders (e.g. developers, developers, users)~\cite{AIA-UK2021} in an AI supply chain must play their roles. Being AI developers/researchers, we can play an important role (as other stakeholders can contribute to AIA if there are trustworthy AI systems) in building an effective AI assurance ecosystem by embedding assurance in the design phase of the AI system.


\begin{figure*}[hbt!]
\centering
\includegraphics[width=17.7cm]{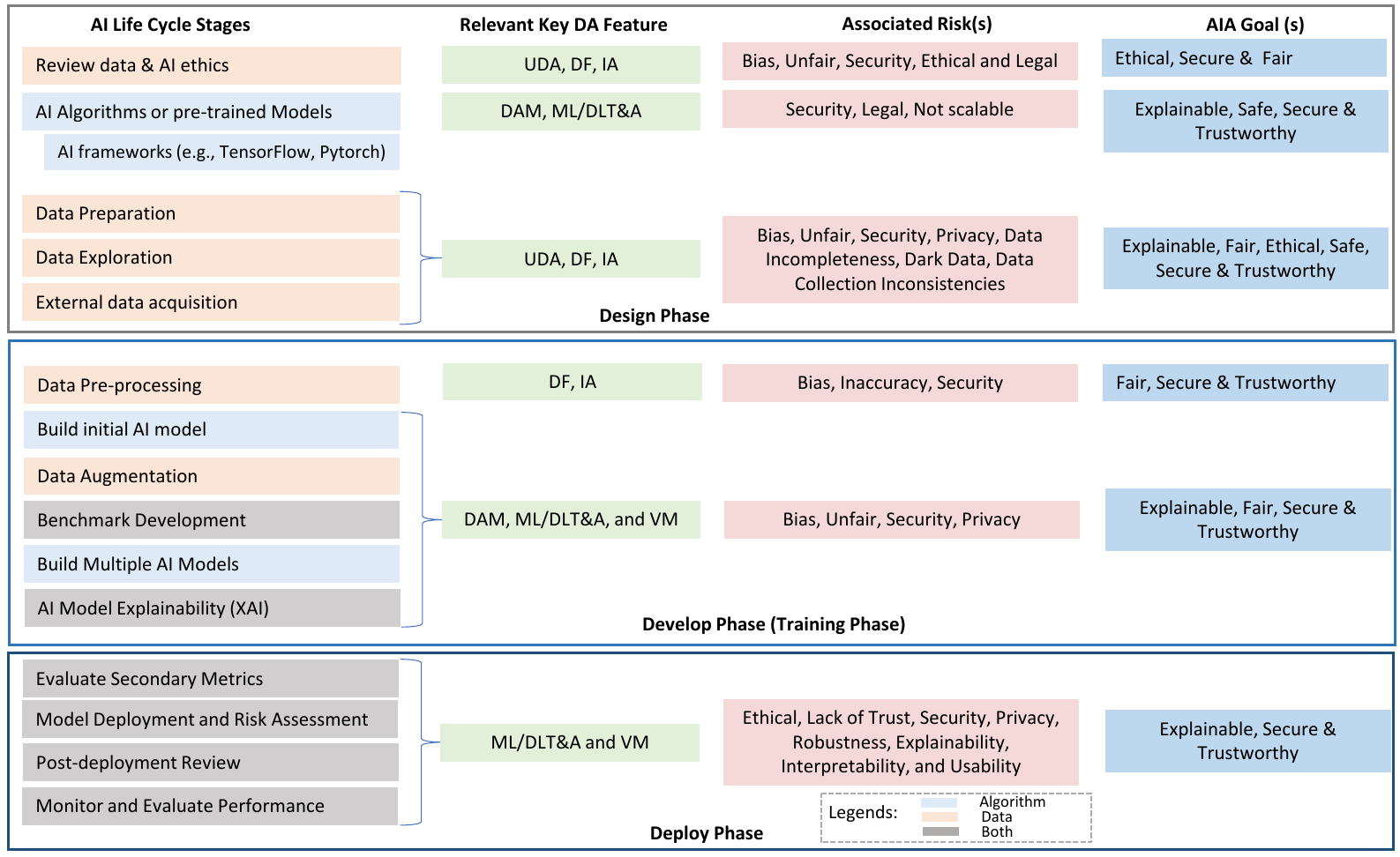}
\caption{Mapping between AI Life Cycle's Key Stages, Features of DA and Aspects of AIA.}
\label{6c}
\end{figure*}

\subsubsection{Mapping between Key Features of DA and Aspects of AIA}

Data and algorithms are the two main components of AI. Therefore, AIA can be viewed from data and algorithmic perspectives~\cite{batarseh2021survey}. Moreover, these two components go through different activities in an AI lifecycle (AIL) \cite{AI-lifecycle22}, which are closely linked to the assurance or assurance goals. For example, the data preparation stage of AIA may include a biased data set, leading to a biased and unfair ML/DL model development. Figure~\ref{6c} presents a mapping among the key AIL stages, DA (cancer) features and AIA goals, including a few potential risks associated with each stage. Batarseh et al.~\cite{batarseh2021survey} have identified 6 AI goals or 6 aspects of AIA (e.g., Ethical, Explainable, Fair, Safe, Secure and Trustworthy AI). However, not all are equally important in all AI application domains or achievable in all AI subareas (e.g., DL, ML, SL, UL). For example, in UL techniques~\cite{halliwell2020trustworthy}, knowing what to validate or assure is difficult as there is no predefined outcome like SL techniques. Different AI domains need distinct trade-offs. For example, trustworthiness is much more important in AI healthcare applications than in private sector revenue estimation. Hence, addressing AIA specific to an AI subarea and goal for an AI domain (e.g., healthcare or cancer) is better than a generic one (e.g., targeting all AI domains).

\subsection{AIA in Existing OC Data Analyses}

\begin{figure}[hbt!]
\centering
\includegraphics[width=7.0cm]{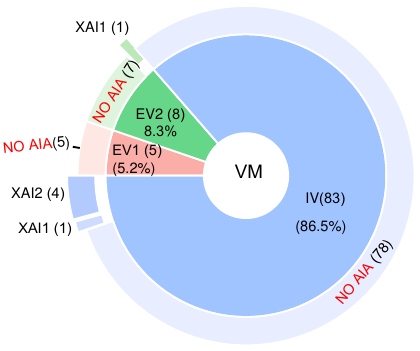}
\caption{AIA in Existing OC Data Analyses.}
\label{6d}
\end{figure}

AIA is a new but active research area. No study has explicitly addressed AIA with all six goals/aspects in the medical or healthcare domain. This is why only 6.25\% (6/96) (Figure~\ref{6d}) of the existing OC data analyses addressed one aspect (XIA) of AIA. Two studies~\cite{white2022deep, laios2022factors} directly addressed XAI (XAI1) and four studies~\cite{wang2023interpretable, nasimian2023deep, liu2021transynergy, bote2022multivariate} indirectly addressed XAI (XAI2) by interpretability. Interestingly, 5/6 studies are from the T\&M subdomain and only 1 from the D\&D subdomain. 

\section{Open Research Challenges and Research Directions}\label{sec4}

Despite the advances in data analysis using DL in OC, some unresolved research challenges remain to be addressed. For example, further investigation is necessary for predicting and preventing OC, data diversity and DL models' assurance, integrated clinical-multi-omics analysis, and external validation methods. Notably, most (86.5\%, Figure~\ref{fig:5c}) current studies validated their models internally. Also, 84.4\% of the existing integrated studies used NF-NBY-based integration methods. Moreover, only 6.25\% (6/96, Figure~\ref{6d}) of the studies partially and indirectly addressed AIA via XIA or interpretability. According to this survey, significant progress has been made in overcoming many obstacles to DL-based data analysis in cancer, including OC. However, there are still some remaining challenges that need to be addressed.

\subsection{Cancer Subdomains}
Unfortunately, risk prediction for cancer using DL-based data analysis in OC is currently unavailable. Furthermore, current research has not yet covered the subdomains of TM and prognosis as thoroughly as it has DD (66/96, Figure~\ref{fig2b}), leaving room for significant future advancements in these areas.

\subsection{Data Features}
\begin{itemize}
    
    \item Balance Data Set: Imbalanced data distribution and unequal quality in the majority and minority classes of data sets can result in incorrect classification. This imbalanced data set is a significant concern, as a wrong diagnosis can have fatal consequences for cancer patients. Therefore, while efforts to balance data sets are essential, examining the issue of class imbalance from a medical perspective is equally crucial.
    
    \item Sample Size and Diversity: Studies~\cite{BENKENDORF2020101137} showed that with smaller sample sizes (100-1000), there is no obvious advantage of using DNNs. However, most existing OC data analysis sample sizes are between 20-500. Furthermore, cancer is a multi-factorial disease~\cite{CancerRiskPrediction2018}. Therefore, DL algorithms must be trained on diverse data sets not limited to demography and~\cite{behravan2020predicting} an ethnic group. However, 75\% of existing DL-based OC studies (SCD or MCD1, Figure~\ref{fig4b}) are limited to a demography or population. Furthermore, only 6\% of the remaining 25\% studies~\cite{zhao2020cup, shin2021style,white2022deep, guo2020deep} considered data sets from two demographics. Therefore, research for robust and clinically acceptable models using large and diverse data sets is necessary.
    
    \item Data Quality: Training data quality can significantly improve cancer diagnostic performance~\cite{meng2021computationally}. Sometimes, data quality enhancement, including transformation, may decrease the need for diverse large data sets for generalisable DL models. Many (38/96) existing studies used DL techniques to improve data quality, especially image quality. However, with a minimal (2/96)~\cite{Multi-omics-2021, levine2020synthesis} use of generative models, further research is needed in this direction. 
    
\end{itemize}

\subsection{Integrated Analysis}

A significant number (32/96, Figure~\ref{fig4b} and Table~\ref{tab:IA}) of the existing studies did an integrated analysis of OC data. Most of these studies considered CDI, and only eight were on CODI. Furthermore, only two~\cite{guo2020deep, Multi-omics-2021} of these eight did integrated tri-omics analysis without proteomics or metabolomics. Considering the importance of CODI in cancer,  more research is needed in this direction, including tetra- and penta-omics analysis~\cite{Multi-omics-2021}.

\subsection{DL Techniques and Models}

\textbf{Insufficient Data and Transfer Learning:} Standard DL algorithms are trained on thousands, if not millions, of human-labelled samples outside of cancer research. Despite the decreasing cost of creating data, including molecular readings using various technologies, access to large amounts of high-quality data sets in precision oncology still needs to be improved. One solution to the challenge is transfer learning/TL. Some existing studies ~\cite{cancers12092373, christiansen2021ultrasound,du2018classification,alshibli2019shallow} used TL. More research must be done using TL to address the data insufficiency issue in oncology.

\textbf{Model for Small Data:} Model for small data is ongoing in various domains, including cancer~\cite{ko2021gves}. However, existing DL-based OC studies have yet to explore this. Hence, research is needed on models for small cancer data.

\textbf{Generative AI in Cancer Research:} Deep generative models, one of which type (Large Language Models) has recently revolutionised the world, can be used in health and medical care research~\cite{LLMs-23}, including cancer. Only two existing studies used generative models in their analyses. One used a variational autoencoder for data compression~\cite{Multi-omics-2021}, and another used GAN for data image generation~\cite{levine2020synthesis}. Further research is needed to explore the possibility of generating normal/benign samples for relevant cancer. 

\textbf{Hybrid Models and Reinforcement Learning (RL):} A combination of ML and DL algorithms or different DL algorithms can improve a DL system's performance by taking advantage of both algorithms. Inspired by this, some OC data analyses~\cite{wang2021ovarian, sengupta2022deep, Multi-omics-2021, liu2022pattern, kilicarslan2020diagnosis} have proposed hybrid (ML+ DL) solutions and improved their performance compared to their single algorithm-based (ML/DL) counterparts. On the contrary, RL is not widely studied in cancer research. RL integrated with DL algorithms~\cite{eckardt2021reinforcement} have already made their way to precision oncology but still need to be in OC. 

\subsection{AI Assurance:} Considering the associated risks of DL in health and medical care, the inclusion of assurance from all six aspects in DL systems is compulsory. However, only 6.25\% (6/96) (Figure~\ref{6d}) of the existing studies addressed one aspect (XIA) of AIA. The other five goals/aspects of AIA, including trustworthiness (critical for healthcare), are yet to be considered. Most studies used data sets from a population or from a country, which could make the DL model biased to that population. Also, publicly available databases could be a target of data poisoning (security). So, immediate research is needed in other aspects of AIA, including trustworthy, fair and secure DL/AI.

\subsection{Validation Method:} Their clinical validation is the biggest obstacle~\cite{DL-histology21} in developing DL systems in health and medical care. Using one data set to develop and validate a model can lead to overfitting, resulting in a system that only works well for that particular group of patients rather than others. Therefore, validating the DL system with external data sets, preferably from multiple centres, is crucial to ensure that it can be used in routine practice and receive regulatory approval. However, as summarised in Figure~\ref{fig:5c}, 8.3\% (8/96) of existing studies did EV with MCD2~\cite{zhao2020cup, shin2021style,white2022deep, guo2020deep} or data from different population. Therefore, research with multicentre and diverse demographic data sets is needed.

\section{Summary and Future Work}
\label{sec5}

ML/DL-based cancer data analysis is necessary for data-driven understanding and management of the disease. Many DL-based data analyses have focused on OC. The analyses are diverse in terms of (i) the cancer type they address, (ii) their goals (e.g., prediction, detection), (iii) types and origin/source of the data, (iv) data integration methods and (v) their DL technique utilised for the analyses. To provide a holistic view of these diverse analyses, we present a systematic review of the field, especially from the perspectives of the key features of cancer data analysis and AIA. 

The comprehensive analysis of 96 studies revealed a series of pivotal insights that highlight both achievements and gaps in this burgeoning field. 

\begin{itemize}
    \item A predominant emphasis on detection and diagnosis (71\% of the studies) subdomain, while regrettably, no endeavours were found to address the subdomains of prediction and prevention of OC. 
    \item noteworthy observation emerged from the geographic confines of the analysed studies, with 75\% solely restricted to specific regions or countries, limiting the generalisability of their findings.
    \item  The exploration of integrated analyses was limited (33\% of studies), primarily confined to homogenous data sets such as clinical or omics data.
    \item Furthermore, only 8.3\% of the studies validated their models using diverse and external data sets, emphasising the need for rigorous model validation.
    \item The incorporation of AIA within the sphere of cancer data analysis has emerged as nascent, with only a tiny 2.1\% explicitly addressing this crucial aspect through explainability. This highlights the pressing need to embrace AIA comprehensively, encompassing all six goals of AIA, including trustworthiness, fairness, and security.
\end{itemize}
 
Considering the above findings, the following research endeavours should be geared toward bridging these discernible gaps. These include exploring diverse and heterogeneous data-driven integrated analyses, robust validation of models via external data sets that span diverse demographic populations, and a coordinated emphasis on AI assurance encompassing all AI goals.

\bibliographystyle{model1-num-names}

\bibliography{ref}

\begin{thebibliography}{179}
\expandafter\ifx\csname natexlab\endcsname\relax\def\natexlab#1{#1}\fi
\providecommand{\url}[1]{\texttt{#1}}
\providecommand{\href}[2]{#2}
\providecommand{\path}[1]{#1}
\providecommand{\DOIprefix}{doi:}
\providecommand{\ArXivprefix}{arXiv:}
\providecommand{\URLprefix}{URL: }
\providecommand{\Pubmedprefix}{pmid:}
\providecommand{\doi}[1]{\href{http://dx.doi.org/#1}{\path{#1}}}
\providecommand{\Pubmed}[1]{\href{pmid:#1}{\path{#1}}}
\providecommand{\bibinfo}[2]{#2}
\ifx\xfnm\relax \def\xfnm[#1]{\unskip,\space#1}\fi
\bibitem[{UK(????)}]{ovarian-uk}
\bibinfo{author}{UK}, \bibinfo{title}{{Cancer Research, Ovarian cancer
  statistics}},
  \bibinfo{howpublished}{\url{https://www.cancerresearchuk.org/health-professional/cancer-statistics/statistics-by-cancer-type/ovarian-cancer/heading-One}},
  ???? \bibinfo{note}{[Online; accessed 8-August-2020]}.
\bibitem[{Torre et~al.(2018)Torre, Trabert, DeSantis, Miller, Samimi, Runowicz,
  Gaudet, Jemal, and Siegel}]{OvCAStats2018}
\bibinfo{author}{L.~A. Torre}, \bibinfo{author}{B.~Trabert},
  \bibinfo{author}{C.~E. DeSantis}, \bibinfo{author}{K.~D. Miller},
  \bibinfo{author}{G.~Samimi}, \bibinfo{author}{C.~D. Runowicz},
  \bibinfo{author}{M.~M. Gaudet}, \bibinfo{author}{A.~Jemal},
  \bibinfo{author}{R.~L. Siegel},
\newblock \bibinfo{title}{Ovarian cancer statistics, 2018},
\newblock \bibinfo{journal}{CA: A Cancer Journal for Clinicians}
  \bibinfo{volume}{68} (\bibinfo{year}{2018}) \bibinfo{pages}{284--296}.
\bibitem[{Doubeni et~al.(2016)Doubeni, Doubeni, and
  Myers}]{doubeni2016diagnosis}
\bibinfo{author}{C.~A. Doubeni}, \bibinfo{author}{A.~R. Doubeni},
  \bibinfo{author}{A.~E. Myers},
\newblock \bibinfo{title}{Diagnosis and management of ovarian cancer},
\newblock \bibinfo{journal}{American family physician} \bibinfo{volume}{93}
  (\bibinfo{year}{2016}) \bibinfo{pages}{937--944}.
\bibitem[{Lu and Zhan(2018)}]{Crucial_role_multiomic2018}
\bibinfo{author}{M.~Lu}, \bibinfo{author}{X.~Zhan},
\newblock \bibinfo{title}{The crucial role of multiomic approach in cancer
  research and clinically relevant outcomes},
\newblock \bibinfo{journal}{EPMA Journal} \bibinfo{volume}{9}
  (\bibinfo{year}{2018}) \bibinfo{pages}{77--102}.
\bibitem[{Zhang et~al.(2019)Zhang, Zhang, Sun, Yang, Dai, and
  Guo}]{OmiVAE2019a}
\bibinfo{author}{X.~Zhang}, \bibinfo{author}{J.~Zhang},
  \bibinfo{author}{K.~Sun}, \bibinfo{author}{X.~Yang},
  \bibinfo{author}{C.~Dai}, \bibinfo{author}{Y.~Guo},
\newblock \bibinfo{title}{Integrated multi-omics analysis using variational
  autoencoders: Application to pan-cancer classification},
\newblock \bibinfo{journal}{arXiv preprint arXiv:1908.06278}
  (\bibinfo{year}{2019}).
\bibitem[{Chen et~al.(2022)Chen, Yang, Qian, Meng, Bai, Hong, He, Jiang, Yuan,
  Du et~al.}]{chen2022deep}
\bibinfo{author}{H.~Chen}, \bibinfo{author}{B.-W. Yang},
  \bibinfo{author}{L.~Qian}, \bibinfo{author}{Y.-S. Meng},
  \bibinfo{author}{X.-H. Bai}, \bibinfo{author}{X.-W. Hong},
  \bibinfo{author}{X.~He}, \bibinfo{author}{M.-J. Jiang},
  \bibinfo{author}{F.~Yuan}, \bibinfo{author}{Q.-W. Du}, et~al.,
\newblock \bibinfo{title}{Deep learning prediction of ovarian malignancy at us
  compared with o-rads and expert assessment},
\newblock \bibinfo{journal}{Radiology}  (\bibinfo{year}{2022})
  \bibinfo{pages}{211367}.
\bibitem[{Saida et~al.(2022)Saida, Mori, Hoshiai, Sakai, Urushibara, Ishiguro,
  Minami, Satoh, and Nakajima}]{saida2022diagnosing}
\bibinfo{author}{T.~Saida}, \bibinfo{author}{K.~Mori},
  \bibinfo{author}{S.~Hoshiai}, \bibinfo{author}{M.~Sakai},
  \bibinfo{author}{A.~Urushibara}, \bibinfo{author}{T.~Ishiguro},
  \bibinfo{author}{M.~Minami}, \bibinfo{author}{T.~Satoh},
  \bibinfo{author}{T.~Nakajima},
\newblock \bibinfo{title}{{Diagnosing Ovarian Cancer on MRI: A Preliminary
  Study Comparing Deep Learning and Radiologist Assessments}},
\newblock \bibinfo{journal}{Cancers} \bibinfo{volume}{14}
  (\bibinfo{year}{2022}) \bibinfo{pages}{987}.
\bibitem[{Hira et~al.(2021)Hira, Razzaque, Angione, Scrivens, Sawan, and
  Sarkar}]{Multi-omics-2021}
\bibinfo{author}{M.~T. Hira}, \bibinfo{author}{M.~Razzaque},
  \bibinfo{author}{C.~Angione}, \bibinfo{author}{J.~Scrivens},
  \bibinfo{author}{S.~Sawan}, \bibinfo{author}{M.~Sarkar},
\newblock \bibinfo{title}{Integrated multi-omics analysis of ovarian cancer
  using variational autoencoders},
\newblock \bibinfo{journal}{Scientific reports} \bibinfo{volume}{11}
  (\bibinfo{year}{2021}) \bibinfo{pages}{1--16}.
\bibitem[{De~Silva et~al.(2018)De~Silva, Ranasinghe, Bandaragoda, Adikari,
  Mills, Iddamalgoda, Alahakoon, Lawrentschuk, Persad, Osipov, Gray, and
  Bolton}]{Empowerment18}
\bibinfo{author}{D.~De~Silva}, \bibinfo{author}{W.~Ranasinghe},
  \bibinfo{author}{T.~Bandaragoda}, \bibinfo{author}{A.~Adikari},
  \bibinfo{author}{N.~Mills}, \bibinfo{author}{L.~Iddamalgoda},
  \bibinfo{author}{D.~Alahakoon}, \bibinfo{author}{N.~Lawrentschuk},
  \bibinfo{author}{R.~Persad}, \bibinfo{author}{E.~Osipov},
  \bibinfo{author}{R.~Gray}, \bibinfo{author}{D.~Bolton},
\newblock \bibinfo{title}{Machine learning to support social media empowered
  patients in cancer care and cancer treatment decisions},
\newblock \bibinfo{journal}{PLOS ONE} \bibinfo{volume}{13}
  (\bibinfo{year}{2018}) \bibinfo{pages}{1--19}.
\bibitem[{Echle et~al.(2021)Echle, Rindtorff, Brinker, Luedde, Pearson, and
  Kather}]{DL-histology21}
\bibinfo{author}{A.~Echle}, \bibinfo{author}{N.~T. Rindtorff},
  \bibinfo{author}{T.~J. Brinker}, \bibinfo{author}{T.~Luedde},
  \bibinfo{author}{A.~T. Pearson}, \bibinfo{author}{J.~N. Kather},
\newblock \bibinfo{title}{Deep learning in cancer pathology: a new generation
  of clinical biomarkers},
\newblock \bibinfo{journal}{British journal of cancer} \bibinfo{volume}{124}
  (\bibinfo{year}{2021}) \bibinfo{pages}{686--696}.
\bibitem[{Zhang et~al.(2019)Zhang, Huang, and Liu}]{OV-detection-DL2019}
\bibinfo{author}{L.~Zhang}, \bibinfo{author}{J.~Huang},
  \bibinfo{author}{L.~Liu},
\newblock \bibinfo{title}{Improved deep learning network based in combination
  with cost-sensitive learning for early detection of ovarian cancer in color
  ultrasound detecting system},
\newblock \bibinfo{journal}{Journal of medical systems} \bibinfo{volume}{43}
  (\bibinfo{year}{2019}) \bibinfo{pages}{1--9}.
\bibitem[{Zhao et~al.(2020)Zhao, Pan, Namburi, Pattison, Posner, Balachander,
  Paisie, Reddi, Rueter, Gill et~al.}]{zhao2020cup}
\bibinfo{author}{Y.~Zhao}, \bibinfo{author}{Z.~Pan},
  \bibinfo{author}{S.~Namburi}, \bibinfo{author}{A.~Pattison},
  \bibinfo{author}{A.~Posner}, \bibinfo{author}{S.~Balachander},
  \bibinfo{author}{C.~A. Paisie}, \bibinfo{author}{H.~V. Reddi},
  \bibinfo{author}{J.~Rueter}, \bibinfo{author}{A.~J. Gill}, et~al.,
\newblock \bibinfo{title}{Cup-ai-dx: A tool for inferring cancer tissue of
  origin and molecular subtype using rna gene-expression data and artificial
  intelligence},
\newblock \bibinfo{journal}{EBioMedicine} \bibinfo{volume}{61}
  (\bibinfo{year}{2020}) \bibinfo{pages}{103030}.
\bibitem[{Wang et~al.(2022)Wang, Lee, Chang, Lin, Liou, Hsu, Chang, Sai, Wang,
  and Chao}]{wang2022weakly}
\bibinfo{author}{C.-W. Wang}, \bibinfo{author}{Y.-C. Lee},
  \bibinfo{author}{C.-C. Chang}, \bibinfo{author}{Y.-J. Lin},
  \bibinfo{author}{Y.-A. Liou}, \bibinfo{author}{P.-C. Hsu},
  \bibinfo{author}{C.-C. Chang}, \bibinfo{author}{A.-K.-O. Sai},
  \bibinfo{author}{C.-H. Wang}, \bibinfo{author}{T.-K. Chao},
\newblock \bibinfo{title}{A weakly supervised deep learning method for guiding
  ovarian cancer treatment and identifying an effective biomarker},
\newblock \bibinfo{journal}{Cancers} \bibinfo{volume}{14}
  (\bibinfo{year}{2022}) \bibinfo{pages}{1651}.
\bibitem[{Han et~al.(2022)Han, Cheung, Yaffe, and Martel}]{han2022cell}
\bibinfo{author}{W.~Han}, \bibinfo{author}{A.~M. Cheung},
  \bibinfo{author}{M.~J. Yaffe}, \bibinfo{author}{A.~L. Martel},
\newblock \bibinfo{title}{Cell segmentation for immunofluorescence multiplexed
  images using two-stage domain adaptation and weakly labeled data for
  pre-training},
\newblock \bibinfo{journal}{Scientific Reports} \bibinfo{volume}{12}
  (\bibinfo{year}{2022}) \bibinfo{pages}{1--14}.
\bibitem[{Kim et~al.(2021)Kim, Yoon, Kim, Cho, Chung, and
  Song}]{kim2021prognostic}
\bibinfo{author}{S.~I. Kim}, \bibinfo{author}{S.~Yoon}, \bibinfo{author}{T.~M.
  Kim}, \bibinfo{author}{J.~Y. Cho}, \bibinfo{author}{H.~H. Chung},
  \bibinfo{author}{Y.~S. Song},
\newblock \bibinfo{title}{Prognostic implications of body composition change
  during primary treatment in patients with ovarian cancer: a retrospective
  study using an artificial intelligence-based volumetric technique},
\newblock \bibinfo{journal}{Gynecologic Oncology} \bibinfo{volume}{162}
  (\bibinfo{year}{2021}) \bibinfo{pages}{72--79}.
\bibitem[{Zhu et~al.(2021)Zhu, Ferri-Borgogno, Sheng, Yeung, Burks, Cappello,
  Jazaeri, Kim, Han, Birrer et~al.}]{zhu2021sio}
\bibinfo{author}{Y.~Zhu}, \bibinfo{author}{S.~Ferri-Borgogno},
  \bibinfo{author}{J.~Sheng}, \bibinfo{author}{T.-L. Yeung},
  \bibinfo{author}{J.~K. Burks}, \bibinfo{author}{P.~Cappello},
  \bibinfo{author}{A.~A. Jazaeri}, \bibinfo{author}{J.-H. Kim},
  \bibinfo{author}{G.~H. Han}, \bibinfo{author}{M.~J. Birrer}, et~al.,
\newblock \bibinfo{title}{Sio: a spatioimageomics pipeline to identify
  prognostic biomarkers associated with the ovarian tumor microenvironment},
\newblock \bibinfo{journal}{Cancers} \bibinfo{volume}{13}
  (\bibinfo{year}{2021}) \bibinfo{pages}{1777}.
\bibitem[{LeCun et~al.(2015)LeCun, Bengio, and Hinton}]{lecun2015deep}
\bibinfo{author}{Y.~LeCun}, \bibinfo{author}{Y.~Bengio},
  \bibinfo{author}{G.~Hinton},
\newblock \bibinfo{title}{Deep learning},
\newblock \bibinfo{journal}{nature} \bibinfo{volume}{521}
  (\bibinfo{year}{2015}) \bibinfo{pages}{436--444}.
\bibitem[{M{\"o}kander et~al.(2021)M{\"o}kander, Morley, Taddeo, and
  Floridi}]{mokander2021ethics}
\bibinfo{author}{J.~M{\"o}kander}, \bibinfo{author}{J.~Morley},
  \bibinfo{author}{M.~Taddeo}, \bibinfo{author}{L.~Floridi},
\newblock \bibinfo{title}{Ethics-based auditing of automated decision-making
  systems: Nature, scope, and limitations},
\newblock \bibinfo{journal}{Science and Engineering Ethics}
  \bibinfo{volume}{27} (\bibinfo{year}{2021}) \bibinfo{pages}{44}.
\bibitem[{Hasani et~al.(2022)Hasani, Morris, Rahmim, Summers, Jones, Siegel,
  and Saboury}]{hasani2022trustworthy}
\bibinfo{author}{N.~Hasani}, \bibinfo{author}{M.~A. Morris},
  \bibinfo{author}{A.~Rahmim}, \bibinfo{author}{R.~M. Summers},
  \bibinfo{author}{E.~Jones}, \bibinfo{author}{E.~Siegel},
  \bibinfo{author}{B.~Saboury},
\newblock \bibinfo{title}{Trustworthy artificial intelligence in medical
  imaging},
\newblock \bibinfo{journal}{PET clinics} \bibinfo{volume}{17}
  (\bibinfo{year}{2022}) \bibinfo{pages}{1--12}.
\bibitem[{Batarseh et~al.(2021)Batarseh, Freeman, and
  Huang}]{batarseh2021survey}
\bibinfo{author}{F.~A. Batarseh}, \bibinfo{author}{L.~Freeman},
  \bibinfo{author}{C.-H. Huang},
\newblock \bibinfo{title}{A survey on artificial intelligence assurance},
\newblock \bibinfo{journal}{Journal of Big Data} \bibinfo{volume}{8}
  (\bibinfo{year}{2021}) \bibinfo{pages}{60}.
\bibitem[{Hwangbo et~al.(2021)Hwangbo, Kim, Kim, Eoh, Lee, Kim, Suh, Park, and
  Song}]{hwangbo2021development}
\bibinfo{author}{S.~Hwangbo}, \bibinfo{author}{S.~I. Kim},
  \bibinfo{author}{J.-H. Kim}, \bibinfo{author}{K.~J. Eoh},
  \bibinfo{author}{C.~Lee}, \bibinfo{author}{Y.~T. Kim}, \bibinfo{author}{D.-S.
  Suh}, \bibinfo{author}{T.~Park}, \bibinfo{author}{Y.~S. Song},
\newblock \bibinfo{title}{Development of machine learning models to predict
  platinum sensitivity of high-grade serous ovarian carcinoma},
\newblock \bibinfo{journal}{Cancers} \bibinfo{volume}{13}
  (\bibinfo{year}{2021}) \bibinfo{pages}{1875}.
\bibitem[{Page(2021)}]{Pagen71}
\bibinfo{author}{M.~J. e.~a. Page},
\newblock \bibinfo{title}{The prisma 2020 statement: an updated guideline for
  reporting systematic reviews},
\newblock \bibinfo{journal}{BMJ} \bibinfo{volume}{372} (\bibinfo{year}{2021}).
\bibitem[{Rajula et~al.(2020)Rajula, Verlato, Manchia, Antonucci, and
  Fanos}]{ML-vs-Stats-Methods20}
\bibinfo{author}{H.~S.~R. Rajula}, \bibinfo{author}{G.~Verlato},
  \bibinfo{author}{M.~Manchia}, \bibinfo{author}{N.~Antonucci},
  \bibinfo{author}{V.~Fanos},
\newblock \bibinfo{title}{Comparison of conventional statistical methods with
  machine learning in medicine: diagnosis, drug development, and treatment},
\newblock \bibinfo{journal}{Medicina} \bibinfo{volume}{56}
  (\bibinfo{year}{2020}) \bibinfo{pages}{455}.
\bibitem[{Cruz and Wishart(2006)}]{ML-in-Cancer2006}
\bibinfo{author}{J.~A. Cruz}, \bibinfo{author}{D.~S. Wishart},
\newblock \bibinfo{title}{Applications of machine learning in cancer prediction
  and prognosis},
\newblock \bibinfo{journal}{Cancer informatics} \bibinfo{volume}{2}
  (\bibinfo{year}{2006}) \bibinfo{pages}{117693510600200030}.
\bibitem[{Kourou et~al.(2015)Kourou, Exarchos, Exarchos, Karamouzis, and
  Fotiadis}]{ML-in-Cancer2015}
\bibinfo{author}{K.~Kourou}, \bibinfo{author}{T.~P. Exarchos},
  \bibinfo{author}{K.~P. Exarchos}, \bibinfo{author}{M.~V. Karamouzis},
  \bibinfo{author}{D.~I. Fotiadis},
\newblock \bibinfo{title}{Machine learning applications in cancer prognosis and
  prediction},
\newblock \bibinfo{journal}{Computational and structural biotechnology journal}
  \bibinfo{volume}{13} (\bibinfo{year}{2015}) \bibinfo{pages}{8--17}.
\bibitem[{WONG and YIP(2018)}]{ML-in-Cancer2018}
\bibinfo{author}{D.~WONG}, \bibinfo{author}{S.~YIP}, \bibinfo{title}{{Machine
  learning classifies cancer}},
  \bibinfo{howpublished}{\url{https://www.nature.com/articles/d41586-018-02881-7}},
  \bibinfo{year}{2018}. \bibinfo{note}{[Online; accessed 22-August-2021]}.
\bibitem[{Gulum et~al.(2021)Gulum, Trombley, and
  Kantardzic}]{Explainable-ML-in-Cancer2021}
\bibinfo{author}{M.~A. Gulum}, \bibinfo{author}{C.~M. Trombley},
  \bibinfo{author}{M.~Kantardzic},
\newblock \bibinfo{title}{A review of explainable deep learning cancer
  detection models in medical imaging},
\newblock \bibinfo{journal}{Applied Sciences} \bibinfo{volume}{11}
  (\bibinfo{year}{2021}) \bibinfo{pages}{4573}.
\bibitem[{Kim and Kim(2018)}]{CancerRiskPrediction2018}
\bibinfo{author}{B.-J. Kim}, \bibinfo{author}{S.-H. Kim},
\newblock \bibinfo{title}{Prediction of inherited genomic susceptibility to 20
  common cancer types by a supervised machine-learning method},
\newblock \bibinfo{journal}{Proceedings of the National Academy of Sciences}
  \bibinfo{volume}{115} (\bibinfo{year}{2018}) \bibinfo{pages}{1322--1327}.
\bibitem[{Lu et~al.(2020)Lu, Fan, Xu, Chen, Zheng, Li, Znati, Mi, and
  Jiang}]{CancerRiskPrediction2020a}
\bibinfo{author}{M.~Lu}, \bibinfo{author}{Z.~Fan}, \bibinfo{author}{B.~Xu},
  \bibinfo{author}{L.~Chen}, \bibinfo{author}{X.~Zheng},
  \bibinfo{author}{J.~Li}, \bibinfo{author}{T.~Znati}, \bibinfo{author}{Q.~Mi},
  \bibinfo{author}{J.~Jiang},
\newblock \bibinfo{title}{Using machine learning to predict ovarian cancer},
\newblock \bibinfo{journal}{International Journal of Medical Informatics}
  \bibinfo{volume}{141} (\bibinfo{year}{2020}) \bibinfo{pages}{104195}.
\bibitem[{Abdullah~Alfayez et~al.(2021)Abdullah~Alfayez, Kunz, and
  Grace~Lai}]{CancerRiskPrediction2021}
\bibinfo{author}{A.~Abdullah~Alfayez}, \bibinfo{author}{H.~Kunz},
  \bibinfo{author}{A.~Grace~Lai},
\newblock \bibinfo{title}{Predicting the risk of cancer in adults using
  supervised machine learning: a scoping review},
\newblock \bibinfo{journal}{BMJ Open} \bibinfo{volume}{11}
  (\bibinfo{year}{2021}).
\bibitem[{Patel et~al.(2020)Patel, George, and Rai}]{ML-in-Cancer2020}
\bibinfo{author}{S.~K. Patel}, \bibinfo{author}{B.~George},
  \bibinfo{author}{V.~Rai},
\newblock \bibinfo{title}{Artificial intelligence to decode cancer mechanism:
  beyond patient stratification for precision oncology},
\newblock \bibinfo{journal}{Frontiers in Pharmacology} \bibinfo{volume}{11}
  (\bibinfo{year}{2020}) \bibinfo{pages}{1177}.
\bibitem[{Islam et~al.(2021)Islam, Poly, Yang, and Li}]{Labtest-Based-2021}
\bibinfo{author}{M.~M. Islam}, \bibinfo{author}{T.~N. Poly},
  \bibinfo{author}{H.-C. Yang}, \bibinfo{author}{Y.-C.~J. Li},
\newblock \bibinfo{title}{Deep into laboratory: An artificial intelligence
  approach to recommend laboratory tests},
\newblock \bibinfo{journal}{Diagnostics} \bibinfo{volume}{11}
  (\bibinfo{year}{2021}).
\bibitem[{Giger(2018)}]{Imagingtest-Based-2020a}
\bibinfo{author}{M.~L. Giger},
\newblock \bibinfo{title}{Machine learning in medical imaging},
\newblock \bibinfo{journal}{Journal of the American College of Radiology}
  \bibinfo{volume}{15} (\bibinfo{year}{2018}) \bibinfo{pages}{512--520}.
  \bibinfo{note}{Data Science: Big Data Machine Learning and Artificial
  Intelligence}.
\bibitem[{AKAZAWA and HASHIMOTO(2020)}]{Imagingtest-Based-2020b}
\bibinfo{author}{M.~AKAZAWA}, \bibinfo{author}{K.~HASHIMOTO},
\newblock \bibinfo{title}{Artificial intelligence in ovarian cancer diagnosis},
\newblock \bibinfo{journal}{Anticancer Research} \bibinfo{volume}{40}
  (\bibinfo{year}{2020}) \bibinfo{pages}{4795--4800}.
\bibitem[{Komura and Ishikawa(2018)}]{ML-histology18}
\bibinfo{author}{D.~Komura}, \bibinfo{author}{S.~Ishikawa},
\newblock \bibinfo{title}{Machine learning methods for histopathological image
  analysis},
\newblock \bibinfo{journal}{Computational and Structural Biotechnology Journal}
  \bibinfo{volume}{16} (\bibinfo{year}{2018}) \bibinfo{pages}{34--42}.
\bibitem[{Sun et~al.(2020)Sun, Huang, Jiang, and Hu}]{sun2020dtf}
\bibinfo{author}{Z.~Sun}, \bibinfo{author}{S.~Huang},
  \bibinfo{author}{P.~Jiang}, \bibinfo{author}{P.~Hu},
\newblock \bibinfo{title}{Dtf: deep tensor factorization for predicting
  anticancer drug synergy},
\newblock \bibinfo{journal}{Bioinformatics} \bibinfo{volume}{36}
  (\bibinfo{year}{2020}) \bibinfo{pages}{4483--4489}.
\bibitem[{Boniolo et~al.(2021)Boniolo, Dorigatti, Ohnmacht, Saur, Schubert, and
  Menden}]{Drug-design-2021}
\bibinfo{author}{F.~Boniolo}, \bibinfo{author}{E.~Dorigatti},
  \bibinfo{author}{A.~Ohnmacht}, \bibinfo{author}{D.~Saur},
  \bibinfo{author}{B.~Schubert}, \bibinfo{author}{M.~P. Menden},
\newblock \bibinfo{title}{Artificial intelligence in early drug discovery
  enabling precision medicine},
\newblock \bibinfo{journal}{Expert Opinion on Drug Discovery}
  (\bibinfo{year}{2021}) \bibinfo{pages}{1--17}.
\bibitem[{De~Silva et~al.(2018)De~Silva, Ranasinghe, Bandaragoda, Adikari,
  Mills, Iddamalgoda, Alahakoon, Lawrentschuk, Persad, Osipov
  et~al.}]{Patient-Empowerment2018}
\bibinfo{author}{D.~De~Silva}, \bibinfo{author}{W.~Ranasinghe},
  \bibinfo{author}{T.~Bandaragoda}, \bibinfo{author}{A.~Adikari},
  \bibinfo{author}{N.~Mills}, \bibinfo{author}{L.~Iddamalgoda},
  \bibinfo{author}{D.~Alahakoon}, \bibinfo{author}{N.~Lawrentschuk},
  \bibinfo{author}{R.~Persad}, \bibinfo{author}{E.~Osipov}, et~al.,
\newblock \bibinfo{title}{Machine learning to support social media empowered
  patients in cancer care and cancer treatment decisions},
\newblock \bibinfo{journal}{PloS one} \bibinfo{volume}{13}
  (\bibinfo{year}{2018}) \bibinfo{pages}{e0205855}.
\bibitem[{Avesani et~al.(2022)Avesani, Tran, Cammarata, Botta, Raimondi, Russo,
  Persiani, Bonatti, Tagliaferri, Dolciami et~al.}]{avesani2022ct}
\bibinfo{author}{G.~Avesani}, \bibinfo{author}{H.~E. Tran},
  \bibinfo{author}{G.~Cammarata}, \bibinfo{author}{F.~Botta},
  \bibinfo{author}{S.~Raimondi}, \bibinfo{author}{L.~Russo},
  \bibinfo{author}{S.~Persiani}, \bibinfo{author}{M.~Bonatti},
  \bibinfo{author}{T.~Tagliaferri}, \bibinfo{author}{M.~Dolciami}, et~al.,
\newblock \bibinfo{title}{Ct-based radiomics and deep learning for brca
  mutation and progression-free survival prediction in ovarian cancer using a
  multicentric dataset},
\newblock \bibinfo{journal}{Cancers} \bibinfo{volume}{14}
  (\bibinfo{year}{2022}) \bibinfo{pages}{2739}.
\bibitem[{Xu et~al.(2020)Xu, Ju, Tong, Zhou, and
  Yang}]{Recurrence-prediction-2020}
\bibinfo{author}{Y.~Xu}, \bibinfo{author}{L.~Ju}, \bibinfo{author}{J.~Tong},
  \bibinfo{author}{C.-M. Zhou}, \bibinfo{author}{J.-J. Yang},
\newblock \bibinfo{title}{Machine learning algorithms for predicting the
  recurrence of stage iv colorectal cancer after tumor resection},
\newblock \bibinfo{journal}{Scientific reports} \bibinfo{volume}{10}
  (\bibinfo{year}{2020}) \bibinfo{pages}{1--9}.
\bibitem[{Meti et~al.(2021)Meti, Saednia, Lagree, Tabbarah, Mohebpour, Kiss,
  Lu, Slodkowska, Gandhi, Jerzak et~al.}]{Residual-Disease2021}
\bibinfo{author}{N.~Meti}, \bibinfo{author}{K.~Saednia},
  \bibinfo{author}{A.~Lagree}, \bibinfo{author}{S.~Tabbarah},
  \bibinfo{author}{M.~Mohebpour}, \bibinfo{author}{A.~Kiss},
  \bibinfo{author}{F.-I. Lu}, \bibinfo{author}{E.~Slodkowska},
  \bibinfo{author}{S.~Gandhi}, \bibinfo{author}{K.~J. Jerzak}, et~al.,
\newblock \bibinfo{title}{Machine learning frameworks to predict neoadjuvant
  chemotherapy response in breast cancer using clinical and pathological
  features},
\newblock \bibinfo{journal}{JCO Clinical Cancer Informatics}
  \bibinfo{volume}{5} (\bibinfo{year}{2021}) \bibinfo{pages}{66--80}.
\bibitem[{Tseng et~al.(2020)Tseng, Wang, Lin, Lu, Hsieh, and
  Liao}]{Risk-startification2020}
\bibinfo{author}{Y.-J. Tseng}, \bibinfo{author}{H.-Y. Wang},
  \bibinfo{author}{T.-W. Lin}, \bibinfo{author}{J.-J. Lu},
  \bibinfo{author}{C.-H. Hsieh}, \bibinfo{author}{C.-T. Liao},
\newblock \bibinfo{title}{Development of a machine learning model for survival
  risk stratification of patients with advanced oral cancer},
\newblock \bibinfo{journal}{JAMA network open} \bibinfo{volume}{3}
  (\bibinfo{year}{2020}) \bibinfo{pages}{e2011768--e2011768}.
\bibitem[{Kim and Kim(2014)}]{kim2014empirical}
\bibinfo{author}{M.~Kim}, \bibinfo{author}{S.-H. Kim},
\newblock \bibinfo{title}{Empirical prediction of genomic susceptibilities for
  multiple cancer classes},
\newblock \bibinfo{journal}{Proceedings of the National Academy of Sciences}
  \bibinfo{volume}{111} (\bibinfo{year}{2014}) \bibinfo{pages}{1921--1926}.
\bibitem[{Hu et~al.(2023)Hu, Jian, Li, and Gao}]{hu2023deep}
\bibinfo{author}{D.~Hu}, \bibinfo{author}{J.~Jian}, \bibinfo{author}{Y.~Li},
  \bibinfo{author}{X.~Gao},
\newblock \bibinfo{title}{Deep learning-based segmentation of epithelial
  ovarian cancer on t2-weighted magnetic resonance images},
\newblock \bibinfo{journal}{Quantitative Imaging in Medicine and Surgery}
  \bibinfo{volume}{13} (\bibinfo{year}{2023}) \bibinfo{pages}{1464}.
\bibitem[{Wang et~al.(2023)Wang, Zhang, Wang, Yao, Zhang, Liu, Yang, and
  Yuan}]{wang2023deep}
\bibinfo{author}{Y.~Wang}, \bibinfo{author}{H.~Zhang},
  \bibinfo{author}{T.~Wang}, \bibinfo{author}{L.~Yao},
  \bibinfo{author}{G.~Zhang}, \bibinfo{author}{X.~Liu},
  \bibinfo{author}{G.~Yang}, \bibinfo{author}{L.~Yuan},
\newblock \bibinfo{title}{Deep learning for the ovarian lesion localization and
  discrimination between borderline and malignant ovarian tumors based on
  routine mr imaging},
\newblock \bibinfo{journal}{Scientific Reports} \bibinfo{volume}{13}
  (\bibinfo{year}{2023}) \bibinfo{pages}{2770}.
\bibitem[{Gangadhar et~al.(2023)Gangadhar, Sari-Sarraf, and
  Vanapalli}]{gangadhar2023deep}
\bibinfo{author}{A.~Gangadhar}, \bibinfo{author}{H.~Sari-Sarraf},
  \bibinfo{author}{S.~A. Vanapalli},
\newblock \bibinfo{title}{Deep learning assisted holography microscopy for
  in-flow enumeration of tumor cells in blood},
\newblock \bibinfo{journal}{RSC advances} \bibinfo{volume}{13}
  (\bibinfo{year}{2023}) \bibinfo{pages}{4222--4235}.
\bibitem[{Wang et~al.(2023)Wang, Xu, Du, Chen, and Mei}]{wang2023attention}
\bibinfo{author}{S.~Wang}, \bibinfo{author}{X.~Xu}, \bibinfo{author}{H.~Du},
  \bibinfo{author}{Y.~Chen}, \bibinfo{author}{W.~Mei},
\newblock \bibinfo{title}{Attention feature fusion methodology with additional
  constraint for ovarian lesion diagnosis on magnetic resonance images},
\newblock \bibinfo{journal}{Medical Physics} \bibinfo{volume}{50}
  (\bibinfo{year}{2023}) \bibinfo{pages}{297--310}.
\bibitem[{Gajjela et~al.(2023)Gajjela, Brun, Mankar, Corvigno, Kennedy, Zhong,
  Liu, Sood, Mayerich, Berisha et~al.}]{gajjela2023leveraging}
\bibinfo{author}{C.~C. Gajjela}, \bibinfo{author}{M.~Brun},
  \bibinfo{author}{R.~Mankar}, \bibinfo{author}{S.~Corvigno},
  \bibinfo{author}{N.~Kennedy}, \bibinfo{author}{Y.~Zhong},
  \bibinfo{author}{J.~Liu}, \bibinfo{author}{A.~K. Sood},
  \bibinfo{author}{D.~Mayerich}, \bibinfo{author}{S.~Berisha}, et~al.,
\newblock \bibinfo{title}{Leveraging mid-infrared spectroscopic imaging and
  deep learning for tissue subtype classification in ovarian cancer},
\newblock \bibinfo{journal}{Analyst} \bibinfo{volume}{148}
  (\bibinfo{year}{2023}) \bibinfo{pages}{2699--2708}.
\bibitem[{Wei et~al.(2023)Wei, Zhang, Ding, Jia, Xu, Dai, Feng, Qin, Bai, Chen
  et~al.}]{wei2023associating}
\bibinfo{author}{M.~Wei}, \bibinfo{author}{Y.~Zhang},
  \bibinfo{author}{C.~Ding}, \bibinfo{author}{J.~Jia}, \bibinfo{author}{H.~Xu},
  \bibinfo{author}{Y.~Dai}, \bibinfo{author}{G.~Feng},
  \bibinfo{author}{C.~Qin}, \bibinfo{author}{G.~Bai},
  \bibinfo{author}{S.~Chen}, et~al.,
\newblock \bibinfo{title}{Associating peritoneal metastasis with t2-weighted
  mri images in epithelial ovarian cancer using deep learning and radiomics: A
  multicenter study},
\newblock \bibinfo{journal}{Journal of Magnetic Resonance Imaging}
  (\bibinfo{year}{2023}).
\bibitem[{Boyanapalli and Shanthini(2023)}]{boyanapalli2023ovarian}
\bibinfo{author}{A.~Boyanapalli}, \bibinfo{author}{A.~Shanthini},
\newblock \bibinfo{title}{Ovarian cancer detection in computed tomography
  images using ensembled deep optimized learning classifier},
\newblock \bibinfo{journal}{Concurrency and Computation: Practice and
  Experience}  (\bibinfo{year}{2023}) \bibinfo{pages}{e7716}.
\bibitem[{Wang et~al.(2022)Wang, Zhan, Luo, Kang, Li, Xi, Liu, and
  Zhuo}]{wang2022automated}
\bibinfo{author}{G.~Wang}, \bibinfo{author}{H.~Zhan}, \bibinfo{author}{T.~Luo},
  \bibinfo{author}{B.~Kang}, \bibinfo{author}{X.~Li}, \bibinfo{author}{G.~Xi},
  \bibinfo{author}{Z.~Liu}, \bibinfo{author}{S.~Zhuo},
\newblock \bibinfo{title}{Automated ovarian cancer identification using
  end-to-end deep learning and second harmonic generation imaging},
\newblock \bibinfo{journal}{IEEE Journal of Selected Topics in Quantum
  Electronics} \bibinfo{volume}{29} (\bibinfo{year}{2022})
  \bibinfo{pages}{1--9}.
\bibitem[{Kodipalli et~al.(2023)Kodipalli, Fernandes, Dasar, and
  Ismail}]{kodipalli2023computational}
\bibinfo{author}{A.~Kodipalli}, \bibinfo{author}{S.~L. Fernandes},
  \bibinfo{author}{S.~K. Dasar}, \bibinfo{author}{T.~Ismail},
\newblock \bibinfo{title}{Computational framework of inverted fuzzy c-means and
  quantum convolutional neural network towards accurate detection of ovarian
  tumors},
\newblock \bibinfo{journal}{International Journal of E-Health and Medical
  Communications (IJEHMC)} \bibinfo{volume}{14} (\bibinfo{year}{2023})
  \bibinfo{pages}{1--16}.
\bibitem[{Wang et~al.(2023)Wang, Zhou, Wang, Wang, and Wu}]{wang2023dmff}
\bibinfo{author}{M.~Wang}, \bibinfo{author}{G.~Zhou},
  \bibinfo{author}{X.~Wang}, \bibinfo{author}{L.~Wang},
  \bibinfo{author}{Z.~Wu},
\newblock \bibinfo{title}{Dmff-net: A dual encoding multiscale feature fusion
  network for ovarian tumor segmentation},
\newblock \bibinfo{journal}{Frontiers in Public Health} \bibinfo{volume}{10}
  (\bibinfo{year}{2023}) \bibinfo{pages}{1054177}.
\bibitem[{Ho et~al.(2023)Ho, Chui, Vanderbilt, Jung, Robson, Park, Roh, and
  Fuchs}]{ho2023deep}
\bibinfo{author}{D.~J. Ho}, \bibinfo{author}{M.~H. Chui},
  \bibinfo{author}{C.~M. Vanderbilt}, \bibinfo{author}{J.~Jung},
  \bibinfo{author}{M.~E. Robson}, \bibinfo{author}{C.-S. Park},
  \bibinfo{author}{J.~Roh}, \bibinfo{author}{T.~J. Fuchs},
\newblock \bibinfo{title}{Deep interactive learning-based ovarian cancer
  segmentation of h\&e-stained whole slide images to study morphological
  patterns of brca mutation},
\newblock \bibinfo{journal}{Journal of Pathology Informatics}
  \bibinfo{volume}{14} (\bibinfo{year}{2023}) \bibinfo{pages}{100160}.
\bibitem[{Hema et~al.(2022)Hema, Manikandan, Alhomrani, Pradeep, Alamri,
  Sharma, Alhassan et~al.}]{hema2022region}
\bibinfo{author}{L.~Hema}, \bibinfo{author}{R.~Manikandan},
  \bibinfo{author}{M.~Alhomrani}, \bibinfo{author}{N.~Pradeep},
  \bibinfo{author}{A.~S. Alamri}, \bibinfo{author}{S.~Sharma},
  \bibinfo{author}{M.~Alhassan}, et~al.,
\newblock \bibinfo{title}{Region-based segmentation and classification for
  ovarian cancer detection using convolution neural network},
\newblock \bibinfo{journal}{Contrast Media \& Molecular Imaging}
  \bibinfo{volume}{2022} (\bibinfo{year}{2022}).
\bibitem[{Hsu et~al.(2022)Hsu, Su, Hung, Chen, Lu, and Kuo}]{hsu2022automatic}
\bibinfo{author}{S.-T. Hsu}, \bibinfo{author}{Y.-J. Su}, \bibinfo{author}{C.-H.
  Hung}, \bibinfo{author}{M.-J. Chen}, \bibinfo{author}{C.-H. Lu},
  \bibinfo{author}{C.-E. Kuo},
\newblock \bibinfo{title}{Automatic ovarian tumors recognition system based on
  ensemble convolutional neural network with ultrasound imaging},
\newblock \bibinfo{journal}{BMC Medical Informatics and Decision Making}
  \bibinfo{volume}{22} (\bibinfo{year}{2022}) \bibinfo{pages}{298}.
\bibitem[{Bahado-Singh et~al.(2022)Bahado-Singh, Ibrahim, Al-Wahab, Aydas,
  Radhakrishna, Yilmaz, and Vishweswaraiah}]{bahado2022precision}
\bibinfo{author}{R.~O. Bahado-Singh}, \bibinfo{author}{A.~Ibrahim},
  \bibinfo{author}{Z.~Al-Wahab}, \bibinfo{author}{B.~Aydas},
  \bibinfo{author}{U.~Radhakrishna}, \bibinfo{author}{A.~Yilmaz},
  \bibinfo{author}{S.~Vishweswaraiah},
\newblock \bibinfo{title}{Precision gynecologic oncology: circulating cell free
  dna epigenomic analysis, artificial intelligence and the accurate detection
  of ovarian cancer},
\newblock \bibinfo{journal}{Scientific Reports} \bibinfo{volume}{12}
  (\bibinfo{year}{2022}) \bibinfo{pages}{18625}.
\bibitem[{Wang et~al.(2022)Wang, Li, Zheng et~al.}]{wang2022automatic}
\bibinfo{author}{X.~Wang}, \bibinfo{author}{H.~Li}, \bibinfo{author}{P.~Zheng},
  et~al.,
\newblock \bibinfo{title}{Automatic detection and segmentation of ovarian
  cancer using a multitask model in pelvic ct images},
\newblock \bibinfo{journal}{Oxidative Medicine And Cellular Longevity}
  \bibinfo{volume}{2022} (\bibinfo{year}{2022}).
\bibitem[{Nero et~al.(2022)Nero, Boldrini, Lenkowicz, Giudice, Piermattei,
  Inzani, Pasciuto, Minucci, Fagotti, Zannoni et~al.}]{nero2022deep}
\bibinfo{author}{C.~Nero}, \bibinfo{author}{L.~Boldrini},
  \bibinfo{author}{J.~Lenkowicz}, \bibinfo{author}{M.~T. Giudice},
  \bibinfo{author}{A.~Piermattei}, \bibinfo{author}{F.~Inzani},
  \bibinfo{author}{T.~Pasciuto}, \bibinfo{author}{A.~Minucci},
  \bibinfo{author}{A.~Fagotti}, \bibinfo{author}{G.~Zannoni}, et~al.,
\newblock \bibinfo{title}{Deep-learning to predict brca mutation and survival
  from digital h\&e slides of epithelial ovarian cancer},
\newblock \bibinfo{journal}{International Journal of Molecular Sciences}
  \bibinfo{volume}{23} (\bibinfo{year}{2022}) \bibinfo{pages}{11326}.
\bibitem[{Jung et~al.(2022)Jung, Kim, Han, Kim, Kim, Lee, and
  Choi}]{jung2022ovarian}
\bibinfo{author}{Y.~Jung}, \bibinfo{author}{T.~Kim}, \bibinfo{author}{M.-R.
  Han}, \bibinfo{author}{S.~Kim}, \bibinfo{author}{G.~Kim},
  \bibinfo{author}{S.~Lee}, \bibinfo{author}{Y.~J. Choi},
\newblock \bibinfo{title}{Ovarian tumor diagnosis using deep convolutional
  neural networks and a denoising convolutional autoencoder},
\newblock \bibinfo{journal}{Scientific Reports} \bibinfo{volume}{12}
  (\bibinfo{year}{2022}) \bibinfo{pages}{17024}.
\bibitem[{Mayer et~al.(2022)Mayer, Gretser, Heckmann, Ziegler, Walter, Reis,
  Bankov, Becker, Triesch, Wild et~al.}]{mayer2022learn}
\bibinfo{author}{R.~S. Mayer}, \bibinfo{author}{S.~Gretser},
  \bibinfo{author}{L.~E. Heckmann}, \bibinfo{author}{P.~K. Ziegler},
  \bibinfo{author}{B.~Walter}, \bibinfo{author}{H.~Reis},
  \bibinfo{author}{K.~Bankov}, \bibinfo{author}{S.~Becker},
  \bibinfo{author}{J.~Triesch}, \bibinfo{author}{P.~J. Wild}, et~al.,
\newblock \bibinfo{title}{How to learn with intentional mistakes:
  Noisyensembles to overcome poor tissue quality for deep learning in
  computational pathology},
\newblock \bibinfo{journal}{Frontiers in Medicine} \bibinfo{volume}{9}
  (\bibinfo{year}{2022}) \bibinfo{pages}{959068}.
\bibitem[{Farahani et~al.(2022)Farahani, Boschman, Farnell, Darbandsari, Zhang,
  Ahmadvand, Jones, Huntsman, K{\"o}bel, Gilks et~al.}]{farahani2022deep}
\bibinfo{author}{H.~Farahani}, \bibinfo{author}{J.~Boschman},
  \bibinfo{author}{D.~Farnell}, \bibinfo{author}{A.~Darbandsari},
  \bibinfo{author}{A.~Zhang}, \bibinfo{author}{P.~Ahmadvand},
  \bibinfo{author}{S.~J. Jones}, \bibinfo{author}{D.~Huntsman},
  \bibinfo{author}{M.~K{\"o}bel}, \bibinfo{author}{C.~B. Gilks}, et~al.,
\newblock \bibinfo{title}{Deep learning-based histotype diagnosis of ovarian
  carcinoma whole-slide pathology images},
\newblock \bibinfo{journal}{Modern Pathology} \bibinfo{volume}{35}
  (\bibinfo{year}{2022}) \bibinfo{pages}{1983--1990}.
\bibitem[{Wu et~al.(2022)Wu, Xing, Wang, Feng, Wietek, Chong, El-Sahhar, Ahmed,
  Zang, and Zheng}]{wu2022investigation}
\bibinfo{author}{W.~Wu}, \bibinfo{author}{X.~Xing}, \bibinfo{author}{M.~Wang},
  \bibinfo{author}{Y.~Feng}, \bibinfo{author}{N.~Wietek},
  \bibinfo{author}{K.~Chong}, \bibinfo{author}{S.~El-Sahhar},
  \bibinfo{author}{A.~A. Ahmed}, \bibinfo{author}{R.~Zang},
  \bibinfo{author}{Y.~Zheng},
\newblock \bibinfo{title}{Investigation of the potential mechanisms underlying
  nuclear f-actin organization in ovarian cancer cells by high-throughput
  screening in combination with deep learning},
\newblock \bibinfo{journal}{Frontiers in Cell and Developmental Biology}
  \bibinfo{volume}{10} (\bibinfo{year}{2022}).
\bibitem[{White et~al.(2022)White, Park, Sheridan, Chuang
  et~al.}]{white2022deep}
\bibinfo{author}{B.~S. White}, \bibinfo{author}{J.~Park},
  \bibinfo{author}{T.~B. Sheridan}, \bibinfo{author}{J.~H. Chuang}, et~al.,
\newblock \bibinfo{title}{Deep learning features encode interpretable
  morphologies within histological images},
\newblock \bibinfo{journal}{Scientific Reports} \bibinfo{volume}{12}
  (\bibinfo{year}{2022}) \bibinfo{pages}{1--12}.
\bibitem[{Reilly et~al.(2022)Reilly, Bullock, Greenwood, Ure, Stewart,
  Davidoff, DeGrazia, Fritsche, Dunton, Bhardwaj et~al.}]{reilly2022analytical}
\bibinfo{author}{G.~Reilly}, \bibinfo{author}{R.~G. Bullock},
  \bibinfo{author}{J.~Greenwood}, \bibinfo{author}{D.~R. Ure},
  \bibinfo{author}{E.~Stewart}, \bibinfo{author}{P.~Davidoff},
  \bibinfo{author}{J.~DeGrazia}, \bibinfo{author}{H.~Fritsche},
  \bibinfo{author}{C.~J. Dunton}, \bibinfo{author}{N.~Bhardwaj}, et~al.,
\newblock \bibinfo{title}{Analytical validation of a deep neural network
  algorithm for the detection of ovarian cancer},
\newblock \bibinfo{journal}{JCO Clinical Cancer Informatics}
  \bibinfo{volume}{6} (\bibinfo{year}{2022}) \bibinfo{pages}{e2100192}.
\bibitem[{Sun et~al.(2022)Sun, Sun, Zhou, Chen, Hao, Liu, Liu, and
  Chen}]{sun2022xgbg}
\bibinfo{author}{K.~F. Sun}, \bibinfo{author}{L.~M. Sun},
  \bibinfo{author}{D.~Zhou}, \bibinfo{author}{Y.~Y. Chen},
  \bibinfo{author}{X.~W. Hao}, \bibinfo{author}{H.~R. Liu},
  \bibinfo{author}{X.~Liu}, \bibinfo{author}{J.~J. Chen},
\newblock \bibinfo{title}{Xgbg: A novel method for identifying ovarian
  carcinoma susceptible genes based on deep learning},
\newblock \bibinfo{journal}{Frontiers in Oncology} \bibinfo{volume}{12}
  (\bibinfo{year}{2022}).
\bibitem[{Gao et~al.(2022)Gao, Zeng, Xu, Li, Yao, Song, Li, Chen, Tang, Xing
  et~al.}]{gao2022deep}
\bibinfo{author}{Y.~Gao}, \bibinfo{author}{S.~Zeng}, \bibinfo{author}{X.~Xu},
  \bibinfo{author}{H.~Li}, \bibinfo{author}{S.~Yao}, \bibinfo{author}{K.~Song},
  \bibinfo{author}{X.~Li}, \bibinfo{author}{L.~Chen},
  \bibinfo{author}{J.~Tang}, \bibinfo{author}{H.~Xing}, et~al.,
\newblock \bibinfo{title}{Deep learning-enabled pelvic ultrasound images for
  accurate diagnosis of ovarian cancer in china: a retrospective, multicentre,
  diagnostic study},
\newblock \bibinfo{journal}{The Lancet Digital Health} \bibinfo{volume}{4}
  (\bibinfo{year}{2022}) \bibinfo{pages}{e179--e187}.
\bibitem[{Sengupta et~al.(2022)Sengupta, Ali, Bhattacharya, Mustafi,
  Mukhopadhyay, and Sengupta}]{sengupta2022deep}
\bibinfo{author}{D.~Sengupta}, \bibinfo{author}{S.~N. Ali},
  \bibinfo{author}{A.~Bhattacharya}, \bibinfo{author}{J.~Mustafi},
  \bibinfo{author}{A.~Mukhopadhyay}, \bibinfo{author}{K.~Sengupta},
\newblock \bibinfo{title}{A deep hybrid learning pipeline for accurate
  diagnosis of ovarian cancer based on nuclear morphology},
\newblock \bibinfo{journal}{PloS one} \bibinfo{volume}{17}
  (\bibinfo{year}{2022}) \bibinfo{pages}{e0261181}.
\bibitem[{Jian et~al.(2022)Jian, Li, Xia, He, Zhang, Li, Zhao, Zhao, Zhang, Cai
  et~al.}]{jian2022mri}
\bibinfo{author}{J.~Jian}, \bibinfo{author}{Y.~Li}, \bibinfo{author}{W.~Xia},
  \bibinfo{author}{Z.~He}, \bibinfo{author}{R.~Zhang}, \bibinfo{author}{H.~Li},
  \bibinfo{author}{X.~Zhao}, \bibinfo{author}{S.~Zhao},
  \bibinfo{author}{J.~Zhang}, \bibinfo{author}{S.~Cai}, et~al.,
\newblock \bibinfo{title}{Mri-based multiple instance convolutional neural
  network for increased accuracy in the differentiation of borderline and
  malignant epithelial ovarian tumors},
\newblock \bibinfo{journal}{Journal of Magnetic Resonance Imaging}
  \bibinfo{volume}{56} (\bibinfo{year}{2022}) \bibinfo{pages}{173--181}.
\bibitem[{Jeong et~al.(2022)Jeong, Lee, Park, Kang, Yu, Hwang, and
  Kim}]{jeong2022deepcia}
\bibinfo{author}{S.~Jeong}, \bibinfo{author}{D.~Lee}, \bibinfo{author}{H.~R.
  Park}, \bibinfo{author}{J.~Kang}, \bibinfo{author}{Y.~Yu},
  \bibinfo{author}{J.~J. Hwang}, \bibinfo{author}{Y.~H. Kim},
\newblock \bibinfo{title}{Deepcia: a novel deep-learning model for cancer type
  identification using class activation map via transcription factor
  expression},
\newblock \bibinfo{journal}{American Journal of Cancer Research}
  \bibinfo{volume}{12} (\bibinfo{year}{2022}) \bibinfo{pages}{5631}.
\bibitem[{Jeya~Sundari and Brintha(2023)}]{jeya2023factorization}
\bibinfo{author}{M.~Jeya~Sundari}, \bibinfo{author}{N.~Brintha},
\newblock \bibinfo{title}{Factorization-based active contour segmentation and
  pelican optimization-based modified bidirectional long short-term memory for
  ovarian tumor detection},
\newblock \bibinfo{journal}{International Journal of Imaging Systems and
  Technology} \bibinfo{volume}{33} (\bibinfo{year}{2023})
  \bibinfo{pages}{230--245}.
\bibitem[{Ramasamy and Kaliyaperumal(2023)}]{ramasamy2023hybridized}
\bibinfo{author}{S.~Ramasamy}, \bibinfo{author}{V.~Kaliyaperumal},
\newblock \bibinfo{title}{A hybridized channel selection approach with deep
  convolutional neural network for effective ovarian cancer prediction in
  periodic acid-schiff-stained images},
\newblock \bibinfo{journal}{Concurrency and Computation: Practice and
  Experience} \bibinfo{volume}{35} (\bibinfo{year}{2023})
  \bibinfo{pages}{e7568}.
\bibitem[{Kodipalli et~al.(2022)Kodipalli, Guha, Dasar, and
  Ismail}]{kodipalli2022inception}
\bibinfo{author}{A.~Kodipalli}, \bibinfo{author}{S.~Guha},
  \bibinfo{author}{S.~Dasar}, \bibinfo{author}{T.~Ismail},
\newblock \bibinfo{title}{An inception-resnet deep learning approach to
  classify tumours in the ovary as benign and malignant},
\newblock \bibinfo{journal}{Expert Systems}  (\bibinfo{year}{2022})
  \bibinfo{pages}{e13215}.
\bibitem[{Jiang et~al.(2022)Jiang, Tekin, Yuan, Armasu, Winham, Goode, Liu,
  Huang, Guo, and Wang}]{jiang2022computational}
\bibinfo{author}{J.~Jiang}, \bibinfo{author}{B.~Tekin},
  \bibinfo{author}{L.~Yuan}, \bibinfo{author}{S.~Armasu},
  \bibinfo{author}{S.~J. Winham}, \bibinfo{author}{E.~L. Goode},
  \bibinfo{author}{H.~Liu}, \bibinfo{author}{Y.~Huang},
  \bibinfo{author}{R.~Guo}, \bibinfo{author}{C.~Wang},
\newblock \bibinfo{title}{Computational tumor stroma reaction evaluation led to
  novel prognosis-associated fibrosis and molecular signature discoveries in
  high-grade serous ovarian carcinoma},
\newblock \bibinfo{journal}{Frontiers in Medicine} \bibinfo{volume}{9}
  (\bibinfo{year}{2022}) \bibinfo{pages}{994467}.
\bibitem[{Ahn et~al.(2022)Ahn, Kim, Kim, Kim, Park, and
  Lee}]{ahn2022transcriptome}
\bibinfo{author}{T.~Ahn}, \bibinfo{author}{K.~Kim}, \bibinfo{author}{H.~Kim},
  \bibinfo{author}{S.~Kim}, \bibinfo{author}{S.~Park},
  \bibinfo{author}{K.~Lee},
\newblock \bibinfo{title}{A transcriptome-based deep neural network classifier
  for identifying the site of origin in mucinous cancer},
\newblock \bibinfo{journal}{Cancer Informatics} \bibinfo{volume}{21}
  (\bibinfo{year}{2022}) \bibinfo{pages}{11769351221135141}.
\bibitem[{Petrovsky et~al.(2021)Petrovsky, Kopylov, Rudnev, Stepanov, Kulikova,
  Malsagova, and Kaysheva}]{petrovsky2021managing}
\bibinfo{author}{D.~V. Petrovsky}, \bibinfo{author}{A.~T. Kopylov},
  \bibinfo{author}{V.~R. Rudnev}, \bibinfo{author}{A.~A. Stepanov},
  \bibinfo{author}{L.~I. Kulikova}, \bibinfo{author}{K.~A. Malsagova},
  \bibinfo{author}{A.~L. Kaysheva},
\newblock \bibinfo{title}{Managing of unassigned mass spectrometric data by
  neural network for cancer phenotypes classification},
\newblock \bibinfo{journal}{Journal of Personalized Medicine}
  \bibinfo{volume}{11} (\bibinfo{year}{2021}) \bibinfo{pages}{1288}.
\bibitem[{Qazi and Raza(2021)}]{qazi2021silico}
\bibinfo{author}{S.~Qazi}, \bibinfo{author}{K.~Raza},
\newblock \bibinfo{title}{In silico approach to understand epigenetics of potee
  in ovarian cancer},
\newblock \bibinfo{journal}{Journal of integrative bioinformatics}
  \bibinfo{volume}{18} (\bibinfo{year}{2021}).
\bibitem[{Jian et~al.(2021)Jian, Xia, Zhang, Zhao, Zhang, Wu, Qiang, Gao
  et~al.}]{jian2021multiple}
\bibinfo{author}{J.~Jian}, \bibinfo{author}{W.~Xia},
  \bibinfo{author}{R.~Zhang}, \bibinfo{author}{X.~Zhao},
  \bibinfo{author}{J.~Zhang}, \bibinfo{author}{X.~Wu},
  \bibinfo{author}{J.~Qiang}, \bibinfo{author}{X.~Gao}, et~al.,
\newblock \bibinfo{title}{Multiple instance convolutional neural network with
  modality-based attention and contextual multi-instance learning pooling layer
  for effective differentiation between borderline and malignant epithelial
  ovarian tumors},
\newblock \bibinfo{journal}{Artificial Intelligence in Medicine}
  \bibinfo{volume}{121} (\bibinfo{year}{2021}) \bibinfo{pages}{102194}.
\bibitem[{Ye et~al.(2021)Ye, Zhang, Yang, Shen, and Xu}]{ye2021ovarian}
\bibinfo{author}{L.~Ye}, \bibinfo{author}{Y.~Zhang}, \bibinfo{author}{X.~Yang},
  \bibinfo{author}{F.~Shen}, \bibinfo{author}{B.~Xu},
\newblock \bibinfo{title}{An ovarian cancer susceptible gene prediction method
  based on deep learning methods},
\newblock \bibinfo{journal}{Frontiers in Cell and Developmental Biology}
  \bibinfo{volume}{9} (\bibinfo{year}{2021}).
\bibitem[{Meng et~al.(2021)Meng, Li, and Wang}]{meng2021computationally}
\bibinfo{author}{X.~Meng}, \bibinfo{author}{X.~Li}, \bibinfo{author}{X.~Wang},
\newblock \bibinfo{title}{A computationally virtual histological staining
  method to ovarian cancer tissue by deep generative adversarial networks},
\newblock \bibinfo{journal}{Computational and Mathematical Methods in Medicine}
  \bibinfo{volume}{2021} (\bibinfo{year}{2021}).
\bibitem[{Pastuszak et~al.(2021)Pastuszak, Supernat, Best, 't~Veld, Łapińska
  Szumczyk, Łojkowska, Różański, Żaczek, Jassem, Würdinger, Wurdinger,
  and Stokowy}]{Pastuszak_2021}
\bibinfo{author}{K.~Pastuszak}, \bibinfo{author}{A.~Supernat},
  \bibinfo{author}{M.~G. Best}, \bibinfo{author}{S.~G. J. G.~I. 't~Veld},
  \bibinfo{author}{S.~Łapińska Szumczyk}, \bibinfo{author}{A.~Łojkowska},
  \bibinfo{author}{R.~Różański}, \bibinfo{author}{A.~J. Żaczek},
  \bibinfo{author}{J.~Jassem}, \bibinfo{author}{T.~Würdinger},
  \bibinfo{author}{T.~Wurdinger}, \bibinfo{author}{T.~Stokowy},
\newblock \bibinfo{title}{Implatelet classifier: image-converted rna biomarker
  profiles enable blood-based cancer diagnostics},
\newblock \bibinfo{journal}{Molecular Oncology}  (\bibinfo{year}{2021}).
\bibitem[{Xie et~al.(2021)Xie, Erickson, Sheedy, Yin, and
  Hubbard}]{xie2021diagnosis}
\bibinfo{author}{H.~Xie}, \bibinfo{author}{B.~J. Erickson},
  \bibinfo{author}{S.~P. Sheedy}, \bibinfo{author}{J.~Yin},
  \bibinfo{author}{J.~M. Hubbard},
\newblock \bibinfo{title}{The diagnosis and outcome of krukenberg tumors},
\newblock \bibinfo{journal}{Journal of Gastrointestinal Oncology}
  \bibinfo{volume}{12} (\bibinfo{year}{2021}) \bibinfo{pages}{226}.
\bibitem[{Gonz{\'a}lez et~al.(2021)Gonz{\'a}lez, Lakatos, Hoballah,
  Fritz-Klaus, Al-Johani, Brooker, Jeong, Evans, Krauledat, Cramer
  et~al.}]{gonzalez2021characterization}
\bibinfo{author}{G.~Gonz{\'a}lez}, \bibinfo{author}{K.~Lakatos},
  \bibinfo{author}{J.~Hoballah}, \bibinfo{author}{R.~Fritz-Klaus},
  \bibinfo{author}{L.~Al-Johani}, \bibinfo{author}{J.~Brooker},
  \bibinfo{author}{S.~Jeong}, \bibinfo{author}{C.~L. Evans},
  \bibinfo{author}{P.~Krauledat}, \bibinfo{author}{D.~W. Cramer}, et~al.,
\newblock \bibinfo{title}{Characterization of cell-bound ca125 on immune cell
  subtypes of ovarian cancer patients using a novel imaging platform},
\newblock \bibinfo{journal}{Cancers} \bibinfo{volume}{13}
  (\bibinfo{year}{2021}) \bibinfo{pages}{2072}.
\bibitem[{Shin et~al.(2021)Shin, You, Jeon, Jung, An, Park, and
  Roh}]{shin2021style}
\bibinfo{author}{S.~J. Shin}, \bibinfo{author}{S.~C. You},
  \bibinfo{author}{H.~Jeon}, \bibinfo{author}{J.~W. Jung},
  \bibinfo{author}{M.~H. An}, \bibinfo{author}{R.~W. Park},
  \bibinfo{author}{J.~Roh},
\newblock \bibinfo{title}{Style transfer strategy for developing a
  generalizable deep learning application in digital pathology},
\newblock \bibinfo{journal}{Computer Methods and Programs in Biomedicine}
  \bibinfo{volume}{198} (\bibinfo{year}{2021}) \bibinfo{pages}{105815}.
\bibitem[{Christiansen et~al.(2021)Christiansen, Epstein, Smedberg,
  {\AA}kerlund, Smith, and Epstein}]{christiansen2021ultrasound}
\bibinfo{author}{F.~Christiansen}, \bibinfo{author}{E.~Epstein},
  \bibinfo{author}{E.~Smedberg}, \bibinfo{author}{M.~{\AA}kerlund},
  \bibinfo{author}{K.~Smith}, \bibinfo{author}{E.~Epstein},
\newblock \bibinfo{title}{Ultrasound image analysis using deep neural networks
  for discriminating between benign and malignant ovarian tumors: comparison
  with expert subjective assessment},
\newblock \bibinfo{journal}{Ultrasound in Obstetrics \& Gynecology}
  \bibinfo{volume}{57} (\bibinfo{year}{2021}) \bibinfo{pages}{155--163}.
\bibitem[{Wang et~al.(2021)Wang, Cai, Lee, Hu, Purkayastha, Pan, Yi, Tran, Lu,
  Liu et~al.}]{wang2021evaluation}
\bibinfo{author}{R.~Wang}, \bibinfo{author}{Y.~Cai}, \bibinfo{author}{I.~K.
  Lee}, \bibinfo{author}{R.~Hu}, \bibinfo{author}{S.~Purkayastha},
  \bibinfo{author}{I.~Pan}, \bibinfo{author}{T.~Yi}, \bibinfo{author}{T.~M.~L.
  Tran}, \bibinfo{author}{S.~Lu}, \bibinfo{author}{T.~Liu}, et~al.,
\newblock \bibinfo{title}{Evaluation of a convolutional neural network for
  ovarian tumor differentiation based on magnetic resonance imaging},
\newblock \bibinfo{journal}{European radiology} \bibinfo{volume}{31}
  (\bibinfo{year}{2021}) \bibinfo{pages}{4960--4971}.
\bibitem[{Ghoniem et~al.(2021)Ghoniem, Algarni, Refky, and
  Ewees}]{ghoniem2021multi}
\bibinfo{author}{R.~M. Ghoniem}, \bibinfo{author}{A.~D. Algarni},
  \bibinfo{author}{B.~Refky}, \bibinfo{author}{A.~A. Ewees},
\newblock \bibinfo{title}{Multi-modal evolutionary deep learning model for
  ovarian cancer diagnosis},
\newblock \bibinfo{journal}{Symmetry} \bibinfo{volume}{13}
  (\bibinfo{year}{2021}) \bibinfo{pages}{643}.
\bibitem[{Gupta et~al.(2021)Gupta, Gupta, and Kumar}]{Gupta_2021}
\bibinfo{author}{S.~Gupta}, \bibinfo{author}{M.~K. Gupta},
  \bibinfo{author}{R.~Kumar},
\newblock \bibinfo{title}{A novel multi-neural ensemble approach for cancer
  diagnosis},
\newblock \bibinfo{journal}{Applied Artificial Intelligence}
  (\bibinfo{year}{2021}).
\bibitem[{Kopylov et~al.(2021)Kopylov, Petrovsky, Stepanov, Rudnev, Malsagova,
  Butkova, Zakharova, Kostyuk, Kulikova, Enikeev, Potoldykova, Kulikov,
  Zulkarnaev, and Kaysheva}]{Kopylov_2021}
\bibinfo{author}{A.~T. Kopylov}, \bibinfo{author}{D.~V. Petrovsky},
  \bibinfo{author}{A.~A. Stepanov}, \bibinfo{author}{V.~R. Rudnev},
  \bibinfo{author}{K.~A. Malsagova}, \bibinfo{author}{T.~V. Butkova},
  \bibinfo{author}{N.~V. Zakharova}, \bibinfo{author}{G.~P. Kostyuk},
  \bibinfo{author}{L.~Kulikova}, \bibinfo{author}{D.~Enikeev},
  \bibinfo{author}{N.~V. Potoldykova}, \bibinfo{author}{D.~A. Kulikov},
  \bibinfo{author}{A.~Zulkarnaev}, \bibinfo{author}{A.~L. Kaysheva},
\newblock \bibinfo{title}{Convolutional neural network in proteomics and
  metabolomics for determination of comorbidity between cancer and
  schizophrenia.},
\newblock \bibinfo{journal}{Journal of Biomedical Informatics}
  (\bibinfo{year}{2021}).
\bibitem[{Chen et~al.(2021)Chen, Chen, Sun, Wang, He, Sun, Ha, Li, Ou, Zhang
  et~al.}]{chen2021prediction}
\bibinfo{author}{J.~Chen}, \bibinfo{author}{Y.~Chen}, \bibinfo{author}{K.~Sun},
  \bibinfo{author}{Y.~Wang}, \bibinfo{author}{H.~He}, \bibinfo{author}{L.~Sun},
  \bibinfo{author}{S.~Ha}, \bibinfo{author}{X.~Li}, \bibinfo{author}{Y.~Ou},
  \bibinfo{author}{X.~Zhang}, et~al.,
\newblock \bibinfo{title}{Prediction of ovarian cancer-related metabolites
  based on graph neural network},
\newblock \bibinfo{journal}{Frontiers in Cell and Developmental Biology}
  (\bibinfo{year}{2021}) \bibinfo{pages}{2585}.
\bibitem[{Mohammed et~al.(2021)Mohammed, Mwambi, Mboya, Elbashir, and
  Omolo}]{mohammed2021stacking}
\bibinfo{author}{M.~Mohammed}, \bibinfo{author}{H.~Mwambi},
  \bibinfo{author}{I.~B. Mboya}, \bibinfo{author}{M.~K. Elbashir},
  \bibinfo{author}{B.~Omolo},
\newblock \bibinfo{title}{A stacking ensemble deep learning approach to cancer
  type classification based on tcga data},
\newblock \bibinfo{journal}{Scientific reports} \bibinfo{volume}{11}
  (\bibinfo{year}{2021}) \bibinfo{pages}{1--22}.
\bibitem[{Guo et~al.(2020)Guo, Wu, Wang, Zhang, Chai, and Liang}]{guo2020deep}
\bibinfo{author}{L.-Y. Guo}, \bibinfo{author}{A.-H. Wu}, \bibinfo{author}{Y.-x.
  Wang}, \bibinfo{author}{L.-p. Zhang}, \bibinfo{author}{H.~Chai},
  \bibinfo{author}{X.-F. Liang},
\newblock \bibinfo{title}{Deep learning-based ovarian cancer subtypes
  identification using multi-omics data},
\newblock \bibinfo{journal}{BioData Mining} \bibinfo{volume}{13}
  (\bibinfo{year}{2020}) \bibinfo{pages}{1--12}.
\bibitem[{Tanabe(2020)}]{cancers12092373}
\bibinfo{author}{K.~e.~a. Tanabe},
\newblock \bibinfo{title}{Comprehensive serum glycopeptide spectra analysis
  combined with artificial intelligence (csgsa-ai) to diagnose early-stage
  ovarian cancer},
\newblock \bibinfo{journal}{Cancers} \bibinfo{volume}{12}
  (\bibinfo{year}{2020}).
\bibitem[{Basharat et~al.(2020)Basharat, Basharat, Ning, Ning, Ning, and
  Liu}]{Basharat_2020}
\bibinfo{author}{A.~R. Basharat}, \bibinfo{author}{A.~R. Basharat},
  \bibinfo{author}{X.~Ning}, \bibinfo{author}{X.~Ning},
  \bibinfo{author}{X.~Ning}, \bibinfo{author}{X.~Liu},
\newblock \bibinfo{title}{Envcnn: A convolutional neural network model for
  evaluating isotopic envelopes in top-down mass-spectral deconvolution.},
\newblock \bibinfo{journal}{Analytical Chemistry}  (\bibinfo{year}{2020}).
\bibitem[{Urase et~al.(2020)Urase, Nishio, Ueno, Kono, Sofue, Kanda, Maeda,
  Nogami, Hori, and Murakami}]{urase2020simulation}
\bibinfo{author}{Y.~Urase}, \bibinfo{author}{M.~Nishio},
  \bibinfo{author}{Y.~Ueno}, \bibinfo{author}{A.~K. Kono},
  \bibinfo{author}{K.~Sofue}, \bibinfo{author}{T.~Kanda},
  \bibinfo{author}{T.~Maeda}, \bibinfo{author}{M.~Nogami},
  \bibinfo{author}{M.~Hori}, \bibinfo{author}{T.~Murakami},
\newblock \bibinfo{title}{Simulation study of low-dose sparse-sampling ct with
  deep learning-based reconstruction: usefulness for evaluation of ovarian
  cancer metastasis},
\newblock \bibinfo{journal}{Applied Sciences} \bibinfo{volume}{10}
  (\bibinfo{year}{2020}) \bibinfo{pages}{4446}.
\bibitem[{Kilicarslan et~al.(2020)Kilicarslan, Adem, and
  Celik}]{kilicarslan2020diagnosis}
\bibinfo{author}{S.~Kilicarslan}, \bibinfo{author}{K.~Adem},
  \bibinfo{author}{M.~Celik},
\newblock \bibinfo{title}{Diagnosis and classification of cancer using hybrid
  model based on relieff and convolutional neural network},
\newblock \bibinfo{journal}{Medical hypotheses} \bibinfo{volume}{137}
  (\bibinfo{year}{2020}) \bibinfo{pages}{109577}.
\bibitem[{Mallavarapu et~al.(2020)Mallavarapu, Hao, Kim, Oh, and
  Kang}]{mallavarapu2020pathway}
\bibinfo{author}{T.~Mallavarapu}, \bibinfo{author}{J.~Hao},
  \bibinfo{author}{Y.~Kim}, \bibinfo{author}{J.~H. Oh},
  \bibinfo{author}{M.~Kang},
\newblock \bibinfo{title}{Pathway-based deep clustering for molecular subtyping
  of cancer},
\newblock \bibinfo{journal}{Methods} \bibinfo{volume}{173}
  (\bibinfo{year}{2020}) \bibinfo{pages}{24--31}.
\bibitem[{Levine et~al.(2020)Levine, Peng, Farnell, Nursey, Wang, Naso, Ren,
  Farahani, Chen, Chiu et~al.}]{levine2020synthesis}
\bibinfo{author}{A.~B. Levine}, \bibinfo{author}{J.~Peng},
  \bibinfo{author}{D.~Farnell}, \bibinfo{author}{M.~Nursey},
  \bibinfo{author}{Y.~Wang}, \bibinfo{author}{J.~R. Naso},
  \bibinfo{author}{H.~Ren}, \bibinfo{author}{H.~Farahani},
  \bibinfo{author}{C.~Chen}, \bibinfo{author}{D.~Chiu}, et~al.,
\newblock \bibinfo{title}{Synthesis of diagnostic quality cancer pathology
  images by generative adversarial networks},
\newblock \bibinfo{journal}{The Journal of pathology} \bibinfo{volume}{252}
  (\bibinfo{year}{2020}) \bibinfo{pages}{178--188}.
\bibitem[{Klein et~al.(2019)Klein, Kanter, Kulbe, Jank, Denkert, Nebrich,
  Schmitt, Wu, Kunze, Sehouli et~al.}]{klein2019maldi}
\bibinfo{author}{O.~Klein}, \bibinfo{author}{F.~Kanter},
  \bibinfo{author}{H.~Kulbe}, \bibinfo{author}{P.~Jank},
  \bibinfo{author}{C.~Denkert}, \bibinfo{author}{G.~Nebrich},
  \bibinfo{author}{W.~D. Schmitt}, \bibinfo{author}{Z.~Wu},
  \bibinfo{author}{C.~A. Kunze}, \bibinfo{author}{J.~Sehouli}, et~al.,
\newblock \bibinfo{title}{Maldi-imaging for classification of epithelial
  ovarian cancer histotypes from a tissue microarray using machine learning
  methods},
\newblock \bibinfo{journal}{PROTEOMICS--Clinical Applications}
  \bibinfo{volume}{13} (\bibinfo{year}{2019}) \bibinfo{pages}{1700181}.
\bibitem[{AlShibli and Mathkour(2019)}]{alshibli2019shallow}
\bibinfo{author}{A.~AlShibli}, \bibinfo{author}{H.~Mathkour},
\newblock \bibinfo{title}{A shallow convolutional learning network for
  classification of cancers based on copy number variations},
\newblock \bibinfo{journal}{Sensors} \bibinfo{volume}{19}
  (\bibinfo{year}{2019}) \bibinfo{pages}{4207}.
\bibitem[{V{\'a}zquez et~al.(2018)V{\'a}zquez, Mari{\~n}o, Blyuss, Ryan,
  Gentry-Maharaj, Kalsi, Manchanda, Jacobs, Menon, and
  Zaikin}]{vazquez2018quantitative}
\bibinfo{author}{M.~A. V{\'a}zquez}, \bibinfo{author}{I.~P. Mari{\~n}o},
  \bibinfo{author}{O.~Blyuss}, \bibinfo{author}{A.~Ryan},
  \bibinfo{author}{A.~Gentry-Maharaj}, \bibinfo{author}{J.~Kalsi},
  \bibinfo{author}{R.~Manchanda}, \bibinfo{author}{I.~Jacobs},
  \bibinfo{author}{U.~Menon}, \bibinfo{author}{A.~Zaikin},
\newblock \bibinfo{title}{A quantitative performance study of two automatic
  methods for the diagnosis of ovarian cancer},
\newblock \bibinfo{journal}{Biomedical Signal Processing and Control}
  \bibinfo{volume}{46} (\bibinfo{year}{2018}) \bibinfo{pages}{86--93}.
\bibitem[{Du et~al.(2018)Du, Zhang, Zargari, Thai, Gunderson, Moxley, Liu,
  Zheng, and Qiu}]{du2018classification}
\bibinfo{author}{Y.~Du}, \bibinfo{author}{R.~Zhang},
  \bibinfo{author}{A.~Zargari}, \bibinfo{author}{T.~Thai},
  \bibinfo{author}{C.~Gunderson}, \bibinfo{author}{K.~M. Moxley},
  \bibinfo{author}{H.~Liu}, \bibinfo{author}{B.~Zheng},
  \bibinfo{author}{Y.~Qiu},
\newblock \bibinfo{title}{Classification of tumor epithelium and stroma by
  exploiting image features learned by deep convolutional neural networks},
\newblock \bibinfo{journal}{Annals of biomedical Eng.} \bibinfo{volume}{46}
  (\bibinfo{year}{2018}) \bibinfo{pages}{1988--1999}.
\bibitem[{Wu et~al.(2018)Wu, Yan, Liu, and Liu}]{wu2018automatic}
\bibinfo{author}{M.~Wu}, \bibinfo{author}{C.~Yan}, \bibinfo{author}{H.~Liu},
  \bibinfo{author}{Q.~Liu},
\newblock \bibinfo{title}{Automatic classification of ovarian cancer types from
  cytological images using deep convolutional neural networks},
\newblock \bibinfo{journal}{Bioscience reports} \bibinfo{volume}{38}
  (\bibinfo{year}{2018}).
\bibitem[{Liang et~al.(2015)Liang, Li, Chen, and Zeng}]{Liang2015}
\bibinfo{author}{M.~Liang}, \bibinfo{author}{Z.~Li}, \bibinfo{author}{T.~Chen},
  \bibinfo{author}{J.~Zeng},
\newblock \bibinfo{title}{Integrative data analysis of multi-platform cancer
  data with a multimodal deep learning approach},
\newblock \bibinfo{journal}{IEEE/ACM Transactions on Computational Biology and
  Bioinformatics} \bibinfo{volume}{12} (\bibinfo{year}{2015})
  \bibinfo{pages}{928--937}.
\bibitem[{Irajizad et~al.(2022)Irajizad, Han, Celestino, Wu, Murage, Spencer,
  Dennison, Vykoukal, Long, Do et~al.}]{irajizad2022blood}
\bibinfo{author}{E.~Irajizad}, \bibinfo{author}{C.~Y. Han},
  \bibinfo{author}{J.~Celestino}, \bibinfo{author}{R.~Wu},
  \bibinfo{author}{E.~Murage}, \bibinfo{author}{R.~Spencer},
  \bibinfo{author}{J.~B. Dennison}, \bibinfo{author}{J.~Vykoukal},
  \bibinfo{author}{J.~P. Long}, \bibinfo{author}{K.~A. Do}, et~al.,
\newblock \bibinfo{title}{A blood-based metabolite panel for distinguishing
  ovarian cancer from benign pelvic masses},
\newblock \bibinfo{journal}{Clinical Cancer Research} \bibinfo{volume}{28}
  (\bibinfo{year}{2022}) \bibinfo{pages}{4669--4676}.
\bibitem[{Li et~al.(2022)Li, Chen, Zhang, Zhang, He, Yan, Li, Xu, Burkhoff, Luo
  et~al.}]{li2022deep}
\bibinfo{author}{J.~Li}, \bibinfo{author}{Y.~Chen}, \bibinfo{author}{M.~Zhang},
  \bibinfo{author}{P.~Zhang}, \bibinfo{author}{K.~He},
  \bibinfo{author}{F.~Yan}, \bibinfo{author}{J.~Li}, \bibinfo{author}{H.~Xu},
  \bibinfo{author}{D.~Burkhoff}, \bibinfo{author}{Y.~Luo}, et~al.,
\newblock \bibinfo{title}{A deep learning model system for diagnosis and
  management of adnexal masses},
\newblock \bibinfo{journal}{Cancers} \bibinfo{volume}{14}
  (\bibinfo{year}{2022}) \bibinfo{pages}{5291}.
\bibitem[{Reilly et~al.(2023)Reilly, Dunton, Bullock, Ure, Fritsche, Ghosh,
  Pappas, and Phan}]{reilly2023validation}
\bibinfo{author}{G.~P. Reilly}, \bibinfo{author}{C.~J. Dunton},
  \bibinfo{author}{R.~G. Bullock}, \bibinfo{author}{D.~R. Ure},
  \bibinfo{author}{H.~Fritsche}, \bibinfo{author}{S.~Ghosh},
  \bibinfo{author}{T.~C. Pappas}, \bibinfo{author}{R.~T. Phan},
\newblock \bibinfo{title}{Validation of a deep neural network-based algorithm
  supporting clinical management of adnexal mass},
\newblock \bibinfo{journal}{Frontiers in Medicine} \bibinfo{volume}{10}
  (\bibinfo{year}{2023}) \bibinfo{pages}{1102437}.
\bibitem[{Wang et~al.(2023{\natexlab{a}})Wang, Lee, Lin, Chang, Wang, Chao
  et~al.}]{wang2023ensemble}
\bibinfo{author}{C.-W. Wang}, \bibinfo{author}{Y.-C. Lee},
  \bibinfo{author}{Y.-J. Lin}, \bibinfo{author}{C.-C. Chang},
  \bibinfo{author}{C.-H. Wang}, \bibinfo{author}{T.-K. Chao}, et~al.,
\newblock \bibinfo{title}{Ensemble biomarkers for guiding anti-angiogenesis
  therapy for ovarian cancer using deep learning},
\newblock \bibinfo{journal}{Clinical and Translational Medicine}
  \bibinfo{volume}{13} (\bibinfo{year}{2023}{\natexlab{a}}).
\bibitem[{Wang et~al.(2023{\natexlab{b}})Wang, Lee, Lin, Chang, Wang, Chao
  et~al.}]{wang2023interpretable}
\bibinfo{author}{C.-W. Wang}, \bibinfo{author}{Y.-C. Lee},
  \bibinfo{author}{Y.-J. Lin}, \bibinfo{author}{C.-C. Chang},
  \bibinfo{author}{C.-H. Wang}, \bibinfo{author}{T.-K. Chao}, et~al.,
\newblock \bibinfo{title}{Interpretable attention-based deep learning ensemble
  for personalized ovarian cancer treatment without manual annotations},
\newblock \bibinfo{journal}{Computerized Medical Imaging and Graphics}
  \bibinfo{volume}{107} (\bibinfo{year}{2023}{\natexlab{b}})
  \bibinfo{pages}{102233}.
\bibitem[{Nasimian et~al.(2023)Nasimian, Ahmed, Hedenfalk, and
  Kazi}]{nasimian2023deep}
\bibinfo{author}{A.~Nasimian}, \bibinfo{author}{M.~Ahmed},
  \bibinfo{author}{I.~Hedenfalk}, \bibinfo{author}{J.~U. Kazi},
\newblock \bibinfo{title}{A deep tabular data learning model predicting
  cisplatin sensitivity identifies bcl2l1 dependency in cancer},
\newblock \bibinfo{journal}{Computational and Structural Biotechnology Journal}
  \bibinfo{volume}{21} (\bibinfo{year}{2023}) \bibinfo{pages}{956--964}.
\bibitem[{Laios et~al.(2022{\natexlab{a}})Laios, De~Freitas, Saalmink, Tan,
  Johnson, Zubayraeva, Munot, Hutson, Thangavelu, Broadhead
  et~al.}]{laios2022stratification}
\bibinfo{author}{A.~Laios}, \bibinfo{author}{D.~L.~D. De~Freitas},
  \bibinfo{author}{G.~Saalmink}, \bibinfo{author}{Y.~S. Tan},
  \bibinfo{author}{R.~Johnson}, \bibinfo{author}{A.~Zubayraeva},
  \bibinfo{author}{S.~Munot}, \bibinfo{author}{R.~Hutson},
  \bibinfo{author}{A.~Thangavelu}, \bibinfo{author}{T.~Broadhead}, et~al.,
\newblock \bibinfo{title}{Stratification of length of stay prediction following
  surgical cytoreduction in advanced high-grade serous ovarian cancer patients
  using artificial intelligence; the leeds l-ai-os score},
\newblock \bibinfo{journal}{Current Oncology} \bibinfo{volume}{29}
  (\bibinfo{year}{2022}{\natexlab{a}}) \bibinfo{pages}{9088--9104}.
\bibitem[{Laios et~al.(2022{\natexlab{b}})Laios, Kalampokis, Johnson, Munot,
  Thangavelu, Hutson, Broadhead, Theophilou, Leach, Nugent
  et~al.}]{laios2022factors}
\bibinfo{author}{A.~Laios}, \bibinfo{author}{E.~Kalampokis},
  \bibinfo{author}{R.~Johnson}, \bibinfo{author}{S.~Munot},
  \bibinfo{author}{A.~Thangavelu}, \bibinfo{author}{R.~Hutson},
  \bibinfo{author}{T.~Broadhead}, \bibinfo{author}{G.~Theophilou},
  \bibinfo{author}{C.~Leach}, \bibinfo{author}{D.~Nugent}, et~al.,
\newblock \bibinfo{title}{Factors predicting surgical effort using explainable
  artificial intelligence in advanced stage epithelial ovarian cancer},
\newblock \bibinfo{journal}{Cancers} \bibinfo{volume}{14}
  (\bibinfo{year}{2022}{\natexlab{b}}) \bibinfo{pages}{3447}.
\bibitem[{Wang et~al.(2022)Wang, Chang, Lee, Lin, Lo, Hsu, Liou, Wang, and
  Chao}]{wang2022a-weakly}
\bibinfo{author}{C.-W. Wang}, \bibinfo{author}{C.-C. Chang},
  \bibinfo{author}{Y.-C. Lee}, \bibinfo{author}{Y.-J. Lin},
  \bibinfo{author}{S.-C. Lo}, \bibinfo{author}{P.-C. Hsu},
  \bibinfo{author}{Y.-A. Liou}, \bibinfo{author}{C.-H. Wang},
  \bibinfo{author}{T.-K. Chao},
\newblock \bibinfo{title}{Weakly supervised deep learning for prediction of
  treatment effectiveness on ovarian cancer from histopathology images},
\newblock \bibinfo{journal}{Computerized Medical Imaging and Graphics}
  \bibinfo{volume}{99} (\bibinfo{year}{2022}) \bibinfo{pages}{102093}.
\bibitem[{Laury et~al.(2021)Laury, Blom, Ropponen, Virtanen, and
  Carp{\'e}n}]{laury2021artificial}
\bibinfo{author}{A.~R. Laury}, \bibinfo{author}{S.~Blom},
  \bibinfo{author}{T.~Ropponen}, \bibinfo{author}{A.~Virtanen},
  \bibinfo{author}{O.~M. Carp{\'e}n},
\newblock \bibinfo{title}{Artificial intelligence-based image analysis can
  predict outcome in high-grade serous carcinoma via histology alone},
\newblock \bibinfo{journal}{Scientific reports} \bibinfo{volume}{11}
  (\bibinfo{year}{2021}) \bibinfo{pages}{1--9}.
\bibitem[{Liu and Xie(2021)}]{liu2021transynergy}
\bibinfo{author}{Q.~Liu}, \bibinfo{author}{L.~Xie},
\newblock \bibinfo{title}{Transynergy: Mechanism-driven interpretable deep
  neural network for the synergistic prediction and pathway deconvolution of
  drug combinations},
\newblock \bibinfo{journal}{PLoS computational biology} \bibinfo{volume}{17}
  (\bibinfo{year}{2021}) \bibinfo{pages}{e1008653}.
\bibitem[{Yu et~al.(2020)Yu, Hu, Wang, Matulonis, Mutter, Golden, and
  Kohane}]{yu2020deciphering}
\bibinfo{author}{K.-H. Yu}, \bibinfo{author}{V.~Hu}, \bibinfo{author}{F.~Wang},
  \bibinfo{author}{U.~A. Matulonis}, \bibinfo{author}{G.~L. Mutter},
  \bibinfo{author}{J.~A. Golden}, \bibinfo{author}{I.~S. Kohane},
\newblock \bibinfo{title}{Deciphering serous ovarian carcinoma histopathology
  and platinum response by convolutional neural networks},
\newblock \bibinfo{journal}{BMC medicine} \bibinfo{volume}{18}
  (\bibinfo{year}{2020}) \bibinfo{pages}{1--14}.
\bibitem[{Lei et~al.(2022)Lei, Yu, Li, Yao, Wang, Gao, Wu, Ren, Tan, Zhang
  et~al.}]{lei2022deep}
\bibinfo{author}{R.~Lei}, \bibinfo{author}{Y.~Yu}, \bibinfo{author}{Q.~Li},
  \bibinfo{author}{Q.~Yao}, \bibinfo{author}{J.~Wang},
  \bibinfo{author}{M.~Gao}, \bibinfo{author}{Z.~Wu}, \bibinfo{author}{W.~Ren},
  \bibinfo{author}{Y.~Tan}, \bibinfo{author}{B.~Zhang}, et~al.,
\newblock \bibinfo{title}{Deep learning magnetic resonance imaging predicts
  platinum sensitivity in patients with epithelial ovarian cancer},
\newblock \bibinfo{journal}{Frontiers in Oncology} \bibinfo{volume}{12}
  (\bibinfo{year}{2022}) \bibinfo{pages}{895177}.
\bibitem[{Bote-Curiel et~al.(2022)Bote-Curiel, Ruiz-Llorente, Mu{\~n}oz-Romero,
  Yag{\"u}e-Fern{\'a}ndez, Barqu{\'\i}n, Garc{\'\i}a-Donas, and
  Rojo-{\'A}lvarez}]{bote2022multivariate}
\bibinfo{author}{L.~Bote-Curiel}, \bibinfo{author}{S.~Ruiz-Llorente},
  \bibinfo{author}{S.~Mu{\~n}oz-Romero},
  \bibinfo{author}{M.~Yag{\"u}e-Fern{\'a}ndez},
  \bibinfo{author}{A.~Barqu{\'\i}n}, \bibinfo{author}{J.~Garc{\'\i}a-Donas},
  \bibinfo{author}{J.~L. Rojo-{\'A}lvarez},
\newblock \bibinfo{title}{Multivariate feature selection and autoencoder
  embeddings of ovarian cancer clinical and genetic data},
\newblock \bibinfo{journal}{Expert Systems with Applications}
  \bibinfo{volume}{206} (\bibinfo{year}{2022}) \bibinfo{pages}{117865}.
\bibitem[{Liu et~al.(2023)Liu, Wan, Liu, Wang, Tang, Cui, and Li}]{liu2023deep}
\bibinfo{author}{L.~Liu}, \bibinfo{author}{H.~Wan}, \bibinfo{author}{L.~Liu},
  \bibinfo{author}{J.~Wang}, \bibinfo{author}{Y.~Tang},
  \bibinfo{author}{S.~Cui}, \bibinfo{author}{Y.~Li},
\newblock \bibinfo{title}{Deep learning provides a new magnetic resonance
  imaging-based prognostic biomarker for recurrence prediction in high-grade
  serous ovarian cancer},
\newblock \bibinfo{journal}{Diagnostics} \bibinfo{volume}{13}
  (\bibinfo{year}{2023}) \bibinfo{pages}{748}.
\bibitem[{Zhang et~al.(2023)Zhang, Wei, Zhao, Gu, and
  Meng}]{zhang2023assessing}
\bibinfo{author}{Z.~Zhang}, \bibinfo{author}{Z.~Wei},
  \bibinfo{author}{L.~Zhao}, \bibinfo{author}{C.~Gu},
  \bibinfo{author}{Y.~Meng},
\newblock \bibinfo{title}{Assessing the clinical utility of multi-omics data
  for predicting serous ovarian cancer prognosis},
\newblock \bibinfo{journal}{Journal of Obstetrics and Gynaecology}
  \bibinfo{volume}{43} (\bibinfo{year}{2023}) \bibinfo{pages}{2171778}.
\bibitem[{Zheng et~al.(2022)Zheng, Wang, Zhang, Li, Yang, Yang, and
  Dong}]{zheng2022preoperative}
\bibinfo{author}{Y.~Zheng}, \bibinfo{author}{F.~Wang},
  \bibinfo{author}{W.~Zhang}, \bibinfo{author}{Y.~Li},
  \bibinfo{author}{B.~Yang}, \bibinfo{author}{X.~Yang},
  \bibinfo{author}{T.~Dong},
\newblock \bibinfo{title}{Preoperative ct-based deep learning model for
  predicting overall survival in patients with high-grade serous ovarian
  cancer},
\newblock \bibinfo{journal}{Frontiers in Oncology} \bibinfo{volume}{12}
  (\bibinfo{year}{2022}) \bibinfo{pages}{986089}.
\bibitem[{LIU(2022)}]{LIU2022117643}
\bibinfo{title}{Eocsa: Predicting prognosis of epithelial ovarian cancer with
  whole slide histopathological images},
\newblock \bibinfo{journal}{Expert Systems with Applications}
  \bibinfo{volume}{206} (\bibinfo{year}{2022}) \bibinfo{pages}{117643}.
\bibitem[{Liu et~al.(2022)Liu, Li, Gao, Xie, Chi, Li, Zeng, Xiong, Liu, Shi
  et~al.}]{liu2022platelet}
\bibinfo{author}{C.-J. Liu}, \bibinfo{author}{H.-Y. Li},
  \bibinfo{author}{Y.~Gao}, \bibinfo{author}{G.-Y. Xie}, \bibinfo{author}{J.-H.
  Chi}, \bibinfo{author}{G.-L. Li}, \bibinfo{author}{S.-Q. Zeng},
  \bibinfo{author}{X.-M. Xiong}, \bibinfo{author}{J.-H. Liu},
  \bibinfo{author}{L.-L. Shi}, et~al.,
\newblock \bibinfo{title}{Platelet rna signature independently predicts ovarian
  cancer prognosis by deep learning neural network model},
\newblock \bibinfo{journal}{Protein \& Cell}  (\bibinfo{year}{2022})
  \bibinfo{pages}{pwac053}.
\bibitem[{Yokomizo et~al.(2022)Yokomizo, Lopes, Takashima, Hirose, Kawabata,
  Takenaka, Iida, Yanaihara, Yura, Sago et~al.}]{yokomizo2022o3c}
\bibinfo{author}{R.~Yokomizo}, \bibinfo{author}{T.~J. Lopes},
  \bibinfo{author}{N.~Takashima}, \bibinfo{author}{S.~Hirose},
  \bibinfo{author}{A.~Kawabata}, \bibinfo{author}{M.~Takenaka},
  \bibinfo{author}{Y.~Iida}, \bibinfo{author}{N.~Yanaihara},
  \bibinfo{author}{K.~Yura}, \bibinfo{author}{H.~Sago}, et~al.,
\newblock \bibinfo{title}{O3c glass-class: A machine-learning framework for
  prognostic prediction of ovarian clear-cell carcinoma},
\newblock \bibinfo{journal}{Bioinformatics and Biology Insights}
  \bibinfo{volume}{16} (\bibinfo{year}{2022})
  \bibinfo{pages}{11779322221134312}.
\bibitem[{Tong et~al.(2021)Tong, Wu, and Wang}]{tong2021integrating}
\bibinfo{author}{L.~Tong}, \bibinfo{author}{H.~Wu}, \bibinfo{author}{M.~D.
  Wang},
\newblock \bibinfo{title}{Integrating multi-omics data by learning modality
  invariant representations for improved prediction of overall survival of
  cancer},
\newblock \bibinfo{journal}{Methods} \bibinfo{volume}{189}
  (\bibinfo{year}{2021}) \bibinfo{pages}{74--85}.
\bibitem[{Hao et~al.(2019)Hao, Kim, Mallavarapu, Oh, and
  Kang}]{hao2019interpretable}
\bibinfo{author}{J.~Hao}, \bibinfo{author}{Y.~Kim},
  \bibinfo{author}{T.~Mallavarapu}, \bibinfo{author}{J.~H. Oh},
  \bibinfo{author}{M.~Kang},
\newblock \bibinfo{title}{Interpretable deep neural network for cancer survival
  analysis by integrating genomic and clinical data},
\newblock \bibinfo{journal}{BMC medical genomics} \bibinfo{volume}{12}
  (\bibinfo{year}{2019}) \bibinfo{pages}{1--13}.
\bibitem[{Wang et~al.(2019)Wang, Liu, Rong, Zhou, Bai, Wei, Wei, Wang, Guo, and
  Tian}]{WANG2019171}
\bibinfo{author}{S.~Wang}, \bibinfo{author}{Z.~Liu}, \bibinfo{author}{Y.~Rong},
  \bibinfo{author}{B.~Zhou}, \bibinfo{author}{Y.~Bai},
  \bibinfo{author}{W.~Wei}, \bibinfo{author}{W.~Wei},
  \bibinfo{author}{M.~Wang}, \bibinfo{author}{Y.~Guo},
  \bibinfo{author}{J.~Tian},
\newblock \bibinfo{title}{Deep learning provides a new computed
  tomography-based prognostic biomarker for recurrence prediction in high-grade
  serous ovarian cancer},
\newblock \bibinfo{journal}{Radiotherapy and Oncology} \bibinfo{volume}{132}
  (\bibinfo{year}{2019}) \bibinfo{pages}{171--177}.
\bibitem[{Marusyk and Polyak(2010)}]{Tumour-heterogeneity10}
\bibinfo{author}{A.~Marusyk}, \bibinfo{author}{K.~Polyak},
\newblock \bibinfo{title}{Tumor heterogeneity: causes and consequences},
\newblock \bibinfo{journal}{Biochimica et Biophysica Acta (BBA)-Reviews on
  Cancer} \bibinfo{volume}{1805} (\bibinfo{year}{2010})
  \bibinfo{pages}{105--117}.
\bibitem[{Marusyk et~al.(2012)Marusyk, Almendro, and
  Polyak}]{Tumour-heterogeneity12}
\bibinfo{author}{A.~Marusyk}, \bibinfo{author}{V.~Almendro},
  \bibinfo{author}{K.~Polyak},
\newblock \bibinfo{title}{Intra-tumour heterogeneity: a looking glass for
  cancer?},
\newblock \bibinfo{journal}{Nature Reviews Cancer} \bibinfo{volume}{12}
  (\bibinfo{year}{2012}) \bibinfo{pages}{323--334}.
\bibitem[{Bedard et~al.(2013)Bedard, Hansen, Ratain, and
  Siu}]{Tumour-heterogeneity13}
\bibinfo{author}{P.~L. Bedard}, \bibinfo{author}{A.~R. Hansen},
  \bibinfo{author}{M.~J. Ratain}, \bibinfo{author}{L.~L. Siu},
\newblock \bibinfo{title}{Tumour heterogeneity in the clinic},
\newblock \bibinfo{journal}{Nature} \bibinfo{volume}{501}
  (\bibinfo{year}{2013}) \bibinfo{pages}{355--364}.
\bibitem[{Dagogo-Jack and Shaw(2018)}]{Tumour-heterogeneity18}
\bibinfo{author}{I.~Dagogo-Jack}, \bibinfo{author}{A.~T. Shaw},
\newblock \bibinfo{title}{Tumour heterogeneity and resistance to cancer
  therapies},
\newblock \bibinfo{journal}{Nature reviews Clinical oncology}
  \bibinfo{volume}{15} (\bibinfo{year}{2018}) \bibinfo{pages}{81--94}.
\bibitem[{De~Leo et~al.(2021)De~Leo, Santini, Ceccarelli, Santandrea,
  Palicelli, Acquaviva, Chiarucci, Rosini, Ravegnini, Pession, Turchetti,
  Zamagni, Perrone, De~Iaco, Tallini, and de~Biase}]{OV-types2021}
\bibinfo{author}{A.~De~Leo}, \bibinfo{author}{D.~Santini},
  \bibinfo{author}{C.~Ceccarelli}, \bibinfo{author}{G.~Santandrea},
  \bibinfo{author}{A.~Palicelli}, \bibinfo{author}{G.~Acquaviva},
  \bibinfo{author}{F.~Chiarucci}, \bibinfo{author}{F.~Rosini},
  \bibinfo{author}{G.~Ravegnini}, \bibinfo{author}{A.~Pession},
  \bibinfo{author}{D.~Turchetti}, \bibinfo{author}{C.~Zamagni},
  \bibinfo{author}{A.~M. Perrone}, \bibinfo{author}{P.~De~Iaco},
  \bibinfo{author}{G.~Tallini}, \bibinfo{author}{D.~de~Biase},
\newblock \bibinfo{title}{What is new on ovarian carcinoma: Integrated
  morphologic and molecular analysis following the new 2020 world health
  organization classification of female genital tumors},
\newblock \bibinfo{journal}{Diagnostics} \bibinfo{volume}{11}
  (\bibinfo{year}{2021}).
\bibitem[{Rafique et~al.(2021)Rafique, Islam, and
  Kazi}]{Therapy-prediction2021}
\bibinfo{author}{R.~Rafique}, \bibinfo{author}{S.~R. Islam},
  \bibinfo{author}{J.~U. Kazi},
\newblock \bibinfo{title}{Machine learning in the prediction of cancer
  therapy},
\newblock \bibinfo{journal}{Computational and Structural Biotechnology Journal}
  \bibinfo{volume}{19} (\bibinfo{year}{2021}) \bibinfo{pages}{4003--4017}.
\bibitem[{Kim et~al.(2016)Kim, Huang, and Emery}]{GiGo-HealthResearch2016}
\bibinfo{author}{Y.~Kim}, \bibinfo{author}{J.~Huang},
  \bibinfo{author}{S.~Emery},
\newblock \bibinfo{title}{Garbage in, garbage out: Data collection, quality
  assessment and reporting standards for social media data use in health
  research, infodemiology and digital disease detection},
\newblock \bibinfo{journal}{J Med Internet Res} \bibinfo{volume}{18}
  (\bibinfo{year}{2016}) \bibinfo{pages}{e41}.
\bibitem[{Zhang et~al.(2020)Zhang, Yin, Zeng, Yuan, and
  Zhang}]{zhang2020combining}
\bibinfo{author}{D.~Zhang}, \bibinfo{author}{C.~Yin},
  \bibinfo{author}{J.~Zeng}, \bibinfo{author}{X.~Yuan},
  \bibinfo{author}{P.~Zhang},
\newblock \bibinfo{title}{Combining structured and unstructured data for
  predictive models: a deep learning approach},
\newblock \bibinfo{journal}{BMC medical informatics and decision making}
  \bibinfo{volume}{20} (\bibinfo{year}{2020}) \bibinfo{pages}{1--11}.
\bibitem[{Institute(2019)}]{Data-features2019}
\bibinfo{author}{N.~C. Institute}, \bibinfo{title}{{How Cancer Is Diagnosed}},
  \bibinfo{howpublished}{\url{https://www.cancer.gov/about-cancer/diagnosis-staging/diagnosis}},
  \bibinfo{year}{2019}. \bibinfo{note}{[Online; accessed 26-Dec-2021]}.
\bibitem[{Hu et~al.(2013)Hu, Wang, and Zhan}]{hu2013multi}
\bibinfo{author}{R.~Hu}, \bibinfo{author}{X.~Wang}, \bibinfo{author}{X.~Zhan},
\newblock \bibinfo{title}{Multi-parameter systematic strategies for predictive,
  preventive and personalised medicine in cancer},
\newblock \bibinfo{journal}{EPMA Journal} \bibinfo{volume}{4}
  (\bibinfo{year}{2013}) \bibinfo{pages}{2}.
\bibitem[{Cheng and Zhan(2017)}]{cheng2017pattern}
\bibinfo{author}{T.~Cheng}, \bibinfo{author}{X.~Zhan},
\newblock \bibinfo{title}{Pattern recognition for predictive, preventive, and
  personalized medicine in cancer},
\newblock \bibinfo{journal}{EPMA Journal} \bibinfo{volume}{8}
  (\bibinfo{year}{2017}) \bibinfo{pages}{51--60}.
\bibitem[{Huang et~al.(2017)Huang, Chaudhary, and Garmire}]{huang2017more}
\bibinfo{author}{S.~Huang}, \bibinfo{author}{K.~Chaudhary},
  \bibinfo{author}{L.~X. Garmire},
\newblock \bibinfo{title}{More is better: recent progress in multi-omics data
  integration methods},
\newblock \bibinfo{journal}{Frontiers in genetics} \bibinfo{volume}{8}
  (\bibinfo{year}{2017}) \bibinfo{pages}{84}.
\bibitem[{Zhu et~al.(2017)Zhu, Song, Shen, Arora, Machiela, Song, Landi, Ghosh,
  Chatterjee, Baladandayuthapani et~al.}]{Data-integration-2017}
\bibinfo{author}{B.~Zhu}, \bibinfo{author}{N.~Song}, \bibinfo{author}{R.~Shen},
  \bibinfo{author}{A.~Arora}, \bibinfo{author}{M.~J. Machiela},
  \bibinfo{author}{L.~Song}, \bibinfo{author}{M.~T. Landi},
  \bibinfo{author}{D.~Ghosh}, \bibinfo{author}{N.~Chatterjee},
  \bibinfo{author}{V.~Baladandayuthapani}, et~al.,
\newblock \bibinfo{title}{Integrating clinical and multiple omics data for
  prognostic assessment across human cancers},
\newblock \bibinfo{journal}{Scientific reports} \bibinfo{volume}{7}
  (\bibinfo{year}{2017}) \bibinfo{pages}{1--13}.
\bibitem[{Chakraborty et~al.(2018)Chakraborty, Hosen, Ahmed, and
  Shekhar}]{Chakraborty2018}
\bibinfo{author}{S.~Chakraborty}, \bibinfo{author}{M.~I. Hosen},
  \bibinfo{author}{M.~Ahmed}, \bibinfo{author}{H.~U. Shekhar},
\newblock \bibinfo{title}{Onco-multi-omics approach: A new frontier in cancer
  research},
\newblock \bibinfo{journal}{BioMed research international}
  \bibinfo{volume}{2018} (\bibinfo{year}{2018})
  \bibinfo{pages}{9836256--9836256}.
\bibitem[{Wu et~al.(2019)Wu, Zhou, Ren, Li, Jiang, and
  Ma}]{Data-Integration-methods2019}
\bibinfo{author}{C.~Wu}, \bibinfo{author}{F.~Zhou}, \bibinfo{author}{J.~Ren},
  \bibinfo{author}{X.~Li}, \bibinfo{author}{Y.~Jiang}, \bibinfo{author}{S.~Ma},
\newblock \bibinfo{title}{A selective review of multi-level omics data
  integration using variable selection},
\newblock \bibinfo{journal}{High-throughput} \bibinfo{volume}{8}
  (\bibinfo{year}{2019}) \bibinfo{pages}{4}.
\bibitem[{Subramanian et~al.(2020)Subramanian, Verma, Kumar, Jere, and
  Anamika}]{Data-Integration-methods2020}
\bibinfo{author}{I.~Subramanian}, \bibinfo{author}{S.~Verma},
  \bibinfo{author}{S.~Kumar}, \bibinfo{author}{A.~Jere},
  \bibinfo{author}{K.~Anamika},
\newblock \bibinfo{title}{Multi-omics data integration, interpretation, and its
  application},
\newblock \bibinfo{journal}{Bioinformatics and biology insights}
  \bibinfo{volume}{14} (\bibinfo{year}{2020})
  \bibinfo{pages}{1177932219899051}.
\bibitem[{Zanfardino et~al.(2021)Zanfardino, Castaldo, Pane, Affinito, Aiello,
  Salvatore, and Franzese}]{Musa-2021}
\bibinfo{author}{M.~Zanfardino}, \bibinfo{author}{R.~Castaldo},
  \bibinfo{author}{K.~Pane}, \bibinfo{author}{O.~Affinito},
  \bibinfo{author}{M.~Aiello}, \bibinfo{author}{M.~Salvatore},
  \bibinfo{author}{M.~Franzese},
\newblock \bibinfo{title}{Musa: a graphical user interface for multi-omics data
  integration in radiogenomic studies},
\newblock \bibinfo{journal}{Scientific Reports} \bibinfo{volume}{11}
  (\bibinfo{year}{2021}) \bibinfo{pages}{1--13}.
\bibitem[{Jendoubi(2021)}]{jendoubi2021approaches}
\bibinfo{author}{T.~Jendoubi},
\newblock \bibinfo{title}{Approaches to integrating metabolomics and
  multi-omics data: a primer},
\newblock \bibinfo{journal}{Metabolites} \bibinfo{volume}{11}
  (\bibinfo{year}{2021}) \bibinfo{pages}{184}.
\bibitem[{Tuncbag et~al.(2012)Tuncbag, McCallum, Huang, and
  Fraenkel}]{tuncbag2012steinernet}
\bibinfo{author}{N.~Tuncbag}, \bibinfo{author}{S.~McCallum},
  \bibinfo{author}{S.-s.~C. Huang}, \bibinfo{author}{E.~Fraenkel},
\newblock \bibinfo{title}{Steinernet: a web server for integrating
  ‘omic’data to discover hidden components of response pathways},
\newblock \bibinfo{journal}{Nucleic acids research} \bibinfo{volume}{40}
  (\bibinfo{year}{2012}) \bibinfo{pages}{W505--W509}.
\bibitem[{Cun and Fr{\"o}hlich(2014)}]{cun2014stSVM}
\bibinfo{author}{Y.~Cun}, \bibinfo{author}{H.~Fr{\"o}hlich},
\newblock \bibinfo{title}{Netclass: an r-package for network based, integrative
  biomarker signature discovery},
\newblock \bibinfo{journal}{Bioinformatics} \bibinfo{volume}{30}
  (\bibinfo{year}{2014}) \bibinfo{pages}{1325--1326}.
\bibitem[{Vaske et~al.(2010)Vaske, Benz, Sanborn, Earl, Szeto, Zhu, Haussler,
  and Stuart}]{vaske2010paradigm}
\bibinfo{author}{C.~J. Vaske}, \bibinfo{author}{S.~C. Benz},
  \bibinfo{author}{J.~Z. Sanborn}, \bibinfo{author}{D.~Earl},
  \bibinfo{author}{C.~Szeto}, \bibinfo{author}{J.~Zhu},
  \bibinfo{author}{D.~Haussler}, \bibinfo{author}{J.~M. Stuart},
\newblock \bibinfo{title}{Inference of patient-specific pathway activities from
  multi-dimensional cancer genomics data using paradigm},
\newblock \bibinfo{journal}{Bioinformatics} \bibinfo{volume}{26}
  (\bibinfo{year}{2010}) \bibinfo{pages}{i237--i245}.
\bibitem[{Akavia et~al.(2010)Akavia, Litvin, Kim, Sanchez-Garcia, Kotliar,
  Causton, Pochanard, Mozes, Garraway, and Pe'er}]{akavia2010integrated}
\bibinfo{author}{U.~D. Akavia}, \bibinfo{author}{O.~Litvin},
  \bibinfo{author}{J.~Kim}, \bibinfo{author}{F.~Sanchez-Garcia},
  \bibinfo{author}{D.~Kotliar}, \bibinfo{author}{H.~C. Causton},
  \bibinfo{author}{P.~Pochanard}, \bibinfo{author}{E.~Mozes},
  \bibinfo{author}{L.~A. Garraway}, \bibinfo{author}{D.~Pe'er},
\newblock \bibinfo{title}{An integrated approach to uncover drivers of cancer},
\newblock \bibinfo{journal}{Cell} \bibinfo{volume}{143} (\bibinfo{year}{2010})
  \bibinfo{pages}{1005--1017}.
\bibitem[{Shen et~al.(2009)Shen, Olshen, and Ladanyi}]{shen2009iCluster}
\bibinfo{author}{R.~Shen}, \bibinfo{author}{A.~B. Olshen},
  \bibinfo{author}{M.~Ladanyi},
\newblock \bibinfo{title}{Integrative clustering of multiple genomic data types
  using a joint latent variable model with application to breast and lung
  cancer subtype analysis},
\newblock \bibinfo{journal}{Bioinformatics} \bibinfo{volume}{25}
  (\bibinfo{year}{2009}) \bibinfo{pages}{2906--2912}.
\bibitem[{Kirk et~al.(2012)Kirk, Griffin, Savage, Ghahramani, and
  Wild}]{kirk2012MDI}
\bibinfo{author}{P.~Kirk}, \bibinfo{author}{J.~E. Griffin},
  \bibinfo{author}{R.~S. Savage}, \bibinfo{author}{Z.~Ghahramani},
  \bibinfo{author}{D.~L. Wild},
\newblock \bibinfo{title}{Bayesian correlated clustering to integrate multiple
  datasets},
\newblock \bibinfo{journal}{Bioinformatics} \bibinfo{volume}{28}
  (\bibinfo{year}{2012}) \bibinfo{pages}{3290--3297}.
\bibitem[{Aure et~al.(2013)Aure, Steinfeld, Baumbusch, Liest{\o}l, Lipson,
  Nyberg, Naume, Sahlberg, Kristensen, B{\o}rresen-Dale et~al.}]{aure2013iPAC}
\bibinfo{author}{M.~R. Aure}, \bibinfo{author}{I.~Steinfeld},
  \bibinfo{author}{L.~O. Baumbusch}, \bibinfo{author}{K.~Liest{\o}l},
  \bibinfo{author}{D.~Lipson}, \bibinfo{author}{S.~Nyberg},
  \bibinfo{author}{B.~Naume}, \bibinfo{author}{K.~K. Sahlberg},
  \bibinfo{author}{V.~N. Kristensen}, \bibinfo{author}{A.-L. B{\o}rresen-Dale},
  et~al.,
\newblock \bibinfo{title}{Identifying in-trans process associated genes in
  breast cancer by integrated analysis of copy number and expression data},
\newblock \bibinfo{journal}{PloS one} \bibinfo{volume}{8}
  (\bibinfo{year}{2013}) \bibinfo{pages}{e53014}.
\bibitem[{Louhimo and Hautaniemi(2011)}]{louhimo2011cnamet}
\bibinfo{author}{R.~Louhimo}, \bibinfo{author}{S.~Hautaniemi},
\newblock \bibinfo{title}{Cnamet: an r package for integrating copy number,
  methylation and expression data},
\newblock \bibinfo{journal}{Bioinformatics} \bibinfo{volume}{27}
  (\bibinfo{year}{2011}) \bibinfo{pages}{887--888}.
\bibitem[{Rajula et~al.(2020)Rajula, Verlato, Manchia, Antonucci, and
  Fanos}]{rajula2020Stats-ML}
\bibinfo{author}{H.~S.~R. Rajula}, \bibinfo{author}{G.~Verlato},
  \bibinfo{author}{M.~Manchia}, \bibinfo{author}{N.~Antonucci},
  \bibinfo{author}{V.~Fanos},
\newblock \bibinfo{title}{Comparison of conventional statistical methods with
  machine learning in medicine: diagnosis, drug development, and treatment},
\newblock \bibinfo{journal}{Medicina} \bibinfo{volume}{56}
  (\bibinfo{year}{2020}) \bibinfo{pages}{455}.
\bibitem[{Huang et~al.(2019)Huang, Zhan, Xiang, Johnson, Helm, Yu, Zhang,
  Salama, Rizkalla, Han, and Huang}]{SALMON-ANN-survival2019}
\bibinfo{author}{Z.~Huang}, \bibinfo{author}{X.~Zhan},
  \bibinfo{author}{S.~Xiang}, \bibinfo{author}{T.~S. Johnson},
  \bibinfo{author}{B.~Helm}, \bibinfo{author}{C.~Y. Yu},
  \bibinfo{author}{J.~Zhang}, \bibinfo{author}{P.~Salama},
  \bibinfo{author}{M.~Rizkalla}, \bibinfo{author}{Z.~Han},
  \bibinfo{author}{K.~Huang},
\newblock \bibinfo{title}{Salmon: Survival analysis learning with multi-omics
  neural networks on breast cancer},
\newblock \bibinfo{journal}{Frontiers in Genetics} \bibinfo{volume}{10}
  (\bibinfo{year}{2019}) \bibinfo{pages}{166}.
\bibitem[{Eckardt et~al.(2021)Eckardt, Wendt, Bornh{\"a}user, and
  Middeke}]{eckardt2021reinforcement}
\bibinfo{author}{J.-N. Eckardt}, \bibinfo{author}{K.~Wendt},
  \bibinfo{author}{M.~Bornh{\"a}user}, \bibinfo{author}{J.~M. Middeke},
\newblock \bibinfo{title}{Reinforcement learning for precision oncology},
\newblock \bibinfo{journal}{Cancers} \bibinfo{volume}{13}
  (\bibinfo{year}{2021}) \bibinfo{pages}{4624}.
\bibitem[{Pocevičiūtė et~al.(2021)Pocevičiūtė, Eilertsen, and
  Lundström}]{2021unsupervised}
\bibinfo{author}{M.~Pocevičiūtė}, \bibinfo{author}{G.~Eilertsen},
  \bibinfo{author}{C.~Lundström}, \bibinfo{title}{Unsupervised anomaly
  detection in digital pathology using gans}, \bibinfo{year}{2021}.
  \href{http://arxiv.org/abs/2103.08945}{\tt arXiv:2103.08945}.
\bibitem[{Khalifa et~al.(2019)Khalifa, Loey, Taha, and
  Mohamed}]{khalifa2019deep}
\bibinfo{author}{N.~E.~M. Khalifa}, \bibinfo{author}{M.~Loey},
  \bibinfo{author}{M.~H.~N. Taha}, \bibinfo{author}{H.~N. E.~T. Mohamed},
\newblock \bibinfo{title}{Deep transfer learning models for medical diabetic
  retinopathy detection},
\newblock \bibinfo{journal}{Acta Informatica Medica} \bibinfo{volume}{27}
  (\bibinfo{year}{2019}) \bibinfo{pages}{327}.
\bibitem[{Zemmal et~al.(2016)Zemmal, Azizi, Dey, and Sellami}]{SLL2016}
\bibinfo{author}{N.~Zemmal}, \bibinfo{author}{N.~Azizi},
  \bibinfo{author}{N.~Dey}, \bibinfo{author}{M.~Sellami},
\newblock \bibinfo{title}{Adaptive semi supervised support vector machine semi
  supervised learning with features cooperation for breast cancer
  classification},
\newblock \bibinfo{journal}{Journal of Medical Imaging and Health Informatics}
  \bibinfo{volume}{6} (\bibinfo{year}{2016}) \bibinfo{pages}{53--62}.
\bibitem[{Al-Azzam and Shatnawi(2021)}]{SSL2021}
\bibinfo{author}{N.~Al-Azzam}, \bibinfo{author}{I.~Shatnawi},
\newblock \bibinfo{title}{Comparing supervised and semi-supervised machine
  learning models on diagnosing breast cancer},
\newblock \bibinfo{journal}{Annals of Medicine and Surgery}
  \bibinfo{volume}{62} (\bibinfo{year}{2021}) \bibinfo{pages}{53--64}.
\bibitem[{Peikari et~al.(2018)Peikari, Salama, Nofech-Mozes, and
  Martel}]{SSL2018}
\bibinfo{author}{M.~Peikari}, \bibinfo{author}{S.~Salama},
  \bibinfo{author}{S.~Nofech-Mozes}, \bibinfo{author}{A.~L. Martel},
\newblock \bibinfo{title}{A cluster-then-label semi-supervised learning
  approach for pathology image classification},
\newblock \bibinfo{journal}{Scientific reports} \bibinfo{volume}{8}
  (\bibinfo{year}{2018}) \bibinfo{pages}{1--13}.
\bibitem[{Ma and Zhang(2018)}]{SSL2018a}
\bibinfo{author}{T.~Ma}, \bibinfo{author}{A.~Zhang},
\newblock \bibinfo{title}{Affinity network fusion and semi-supervised learning
  for cancer patient clustering},
\newblock \bibinfo{journal}{Methods} \bibinfo{volume}{145}
  (\bibinfo{year}{2018}) \bibinfo{pages}{16--24}.
\bibitem[{Balaprakash et~al.(2019)Balaprakash, Egele, Salim, Wild, Vishwanath,
  Xia, Brettin, and Stevens}]{balaprakash2019scalable}
\bibinfo{author}{P.~Balaprakash}, \bibinfo{author}{R.~Egele},
  \bibinfo{author}{M.~Salim}, \bibinfo{author}{S.~Wild},
  \bibinfo{author}{V.~Vishwanath}, \bibinfo{author}{F.~Xia},
  \bibinfo{author}{T.~Brettin}, \bibinfo{author}{R.~Stevens},
\newblock \bibinfo{title}{Scalable reinforcement-learning-based neural
  architecture search for cancer deep learning research},
\newblock in: \bibinfo{booktitle}{Proceedings of the International Conference
  for High Performance Computing, Networking, Storage and Analysis},
  \bibinfo{year}{2019}, pp. \bibinfo{pages}{1--33}.
\bibitem[{Daoud et~al.(2020)Daoud, Mdhaffar, Jmaiel, and
  Freisleben}]{daoud2020q}
\bibinfo{author}{S.~Daoud}, \bibinfo{author}{A.~Mdhaffar},
  \bibinfo{author}{M.~Jmaiel}, \bibinfo{author}{B.~Freisleben},
\newblock \bibinfo{title}{Q-rank: Reinforcement learning for recommending
  algorithms to predict drug sensitivity to cancer therapy},
\newblock \bibinfo{journal}{IEEE Journal of Biomedical and Health Informatics}
  \bibinfo{volume}{24} (\bibinfo{year}{2020}) \bibinfo{pages}{3154--3161}.
\bibitem[{Ramspek et~al.(2020)Ramspek, Jager, Dekker, Zoccali, and van
  Diepen}]{EV2020}
\bibinfo{author}{C.~L. Ramspek}, \bibinfo{author}{K.~J. Jager},
  \bibinfo{author}{F.~W. Dekker}, \bibinfo{author}{C.~Zoccali},
  \bibinfo{author}{M.~van Diepen},
\newblock \bibinfo{title}{{External validation of prognostic models:
  what, why, how, when and where?}},
\newblock \bibinfo{journal}{Clinical Kidney Journal} \bibinfo{volume}{14}
  (\bibinfo{year}{2020}) \bibinfo{pages}{49--58}.
\bibitem[{Siontis et~al.(2015)Siontis, Tzoulaki, Castaldi, and
  Ioannidis}]{siontis2015EV}
\bibinfo{author}{G.~C. Siontis}, \bibinfo{author}{I.~Tzoulaki},
  \bibinfo{author}{P.~J. Castaldi}, \bibinfo{author}{J.~P. Ioannidis},
\newblock \bibinfo{title}{External validation of new risk prediction models is
  infrequent and reveals worse prognostic discrimination},
\newblock \bibinfo{journal}{Journal of clinical epidemiology}
  \bibinfo{volume}{68} (\bibinfo{year}{2015}) \bibinfo{pages}{25--34}.
\bibitem[{Ho et~al.(2020)Ho, Phua, Wong, and Goh}]{EV2020a}
\bibinfo{author}{S.~Y. Ho}, \bibinfo{author}{K.~Phua},
  \bibinfo{author}{L.~Wong}, \bibinfo{author}{W.~W.~B. Goh},
\newblock \bibinfo{title}{Extensions of the external validation for checking
  learned model interpretability and generalizability},
\newblock \bibinfo{journal}{Patterns} \bibinfo{volume}{1}
  (\bibinfo{year}{2020}) \bibinfo{pages}{100129}.
\bibitem[{EU(2021)}]{AIA-EC2021}
\bibinfo{author}{EU}, \bibinfo{title}{{LAYING DOWN HARMONISED RULES ON
  ARTIFICIAL INTELLIGENCE (ARTIFICIAL INTELLIGENCE ACT) AND AMENDING CERTAIN
  UNION LEGISLATIVE ACTS}},
  \bibinfo{howpublished}{\url{https://eur-lex.europa.eu/legal-content/EN/TXT/HTML/?uri=CELEX:52021PC0206&from=EN}},
  \bibinfo{year}{2021}. \bibinfo{note}{[Online; accessed 16-May-2023]}.
\bibitem[{for Data~Ethics and Innovation(2021)}]{AIA-UK2021}
\bibinfo{author}{C.~for Data~Ethics}, \bibinfo{author}{Innovation},
  \bibinfo{title}{{The roadmap to an effective AI assurance ecosystem -
  extended version}},
  \bibinfo{howpublished}{\url{https://www.gov.uk/government/publications/the-roadmap-to-an-effective-ai-assurance-ecosystem}},
  \bibinfo{year}{2021}. \bibinfo{note}{[Online; accessed 16-May-2023]}.
\bibitem[{BBC(2021)}]{AIA-US2023}
\bibinfo{author}{BBC}, \bibinfo{title}{{Sam Altman: CEO of OpenAI calls for US
  to regulate artificial intelligence}},
  \bibinfo{howpublished}{\url{https://https://www.bbc.co.uk/news/world-us-canada-65616866}},
  \bibinfo{year}{2021}. \bibinfo{note}{[Online; accessed 16-May-2023]}.
\bibitem[{Freeman et~al.(2022)Freeman, Batarseh, Kuhn, Raunak, and
  Kacker}]{AIA-Consesus2022}
\bibinfo{author}{L.~Freeman}, \bibinfo{author}{F.~A. Batarseh},
  \bibinfo{author}{D.~R. Kuhn}, \bibinfo{author}{M.~S. Raunak},
  \bibinfo{author}{R.~N. Kacker},
\newblock \bibinfo{title}{The path to a consensus on artificial intelligence
  assurance},
\newblock \bibinfo{journal}{Computer} \bibinfo{volume}{55}
  (\bibinfo{year}{2022}) \bibinfo{pages}{82--86}.
\bibitem[{{De Silva} and Alahakoon(2022)}]{AI-lifecycle22}
\bibinfo{author}{D.~{De Silva}}, \bibinfo{author}{D.~Alahakoon},
\newblock \bibinfo{title}{An artificial intelligence life cycle: From
  conception to production},
\newblock \bibinfo{journal}{Patterns} \bibinfo{volume}{3}
  (\bibinfo{year}{2022}) \bibinfo{pages}{100489}.
\bibitem[{Halliwell and Lecue(2020)}]{halliwell2020trustworthy}
\bibinfo{author}{N.~Halliwell}, \bibinfo{author}{F.~Lecue},
\newblock \bibinfo{title}{Trustworthy convolutional neural networks: A gradient
  penalized-based approach},
\newblock \bibinfo{journal}{arXiv preprint arXiv:2009.14260}
  (\bibinfo{year}{2020}).
\bibitem[{Benkendorf and Hawkins(2020)}]{BENKENDORF2020101137}
\bibinfo{author}{D.~J. Benkendorf}, \bibinfo{author}{C.~P. Hawkins},
\newblock \bibinfo{title}{Effects of sample size and network depth on a deep
  learning approach to species distribution modeling},
\newblock \bibinfo{journal}{Ecological Informatics} \bibinfo{volume}{60}
  (\bibinfo{year}{2020}) \bibinfo{pages}{101137}.
\bibitem[{Behravan et~al.(2020)Behravan, Hartikainen, Tengstr{\"o}m, Kosma, and
  Mannermaa}]{behravan2020predicting}
\bibinfo{author}{H.~Behravan}, \bibinfo{author}{J.~M. Hartikainen},
  \bibinfo{author}{M.~Tengstr{\"o}m}, \bibinfo{author}{V.-M. Kosma},
  \bibinfo{author}{A.~Mannermaa},
\newblock \bibinfo{title}{Predicting breast cancer risk using interacting
  genetic and demographic factors and machine learning},
\newblock \bibinfo{journal}{Scientific Reports} \bibinfo{volume}{10}
  (\bibinfo{year}{2020}) \bibinfo{pages}{1--16}.
\bibitem[{Ko et~al.(2021)Ko, Choi, and Ahn}]{ko2021gves}
\bibinfo{author}{S.~Ko}, \bibinfo{author}{J.~Choi}, \bibinfo{author}{J.~Ahn},
\newblock \bibinfo{title}{Gves: machine learning model for identification of
  prognostic genes with a small dataset},
\newblock \bibinfo{journal}{Scientific Reports} \bibinfo{volume}{11}
  (\bibinfo{year}{2021}) \bibinfo{pages}{1--8}.
\bibitem[{Egli(2023)}]{LLMs-23}
\bibinfo{author}{A.~Egli},
\newblock \bibinfo{title}{{ChatGPT, GPT-4, and Other Large Language Models: The
  Next Revolution for Clinical Microbiology?}},
\newblock \bibinfo{journal}{Clinical Infectious Diseases}
  (\bibinfo{year}{2023}) \bibinfo{pages}{ciad407}.
\bibitem[{Wang and Zeng(2021)}]{wang2021ovarian}
\bibinfo{author}{Y.~Wang}, \bibinfo{author}{Q.~Zeng},
\newblock \bibinfo{title}{Ovarian tumor texture classification based on sparse
  auto-encoder network combined with multi-feature fusion and random forest in
  ultrasound image},
\newblock \bibinfo{journal}{Journal of Medical Imaging and Health Informatics}
  \bibinfo{volume}{11} (\bibinfo{year}{2021}) \bibinfo{pages}{424--431}.
\bibitem[{Liu et~al.(2022)Liu, Liang, Liao, and Lu}]{liu2022pattern}
\bibinfo{author}{P.~Liu}, \bibinfo{author}{X.~Liang},
  \bibinfo{author}{S.~Liao}, \bibinfo{author}{Z.~Lu},
\newblock \bibinfo{title}{Pattern classification for ovarian tumors by
  integration of radiomics and deep learning features.},
\newblock \bibinfo{journal}{Current Medical Imaging}  (\bibinfo{year}{2022}).

\end{thebibliography}


%



\end{document}